%% file: main.tex
\documentclass[runningheads]{llncs}

\usepackage{eccv}

\usepackage{eccvabbrv}

\usepackage{graphicx}
\usepackage{multirow}
\usepackage{booktabs}
\usepackage{makecell}
\usepackage{array}
\usepackage[table]{xcolor}
\usepackage{pdflscape}
\usepackage{pifont} %

\input{custom_commands}
\usepackage[accsupp]{axessibility}  %

\usepackage{hyperref}

\usepackage{orcidlink}

\begin{document}

\title{Unified Multi-Layer Subspace Modeling for Cross-Domain OOD Detection}

\titlerunning{PRISM: Multi-Layer OOD Detection}

\author{Gerhard Krumpl\inst{1,2}\orcidlink{0009-0009-6993-1894} \and
Henning Avenhaus\inst{2}\orcidlink{0000-0002-1302-4613} \and
Horst Possegger\inst{1}\orcidlink{0000-0002-5427-9938}}

\authorrunning{G.~Krumpl et al.}

\institute{Institute of Visual Computing, Graz University of Technology, Austria \email{gerhard.krumpl@icg.tugraz.at, possegger@tugraz.at} \and
KESTRELEYE GmbH, Austria}

\maketitle

\input{sec/00_abstract}
\input{sec/01_introduction}
\input{sec/02_related_work}

\input{sec/03_methodology}

\input{sec/04_experiments}

\input{sec/05_conclusion}

\section*{Acknowledgements}

We gratefully acknowledge support from KESTRELEYE GmbH, which provided the core infrastructure and resources throughout the project.
HP was partially funded by the AI Ecosystems programme of the Austrian Federal Ministry of Innovation, Mobility and Infrastructure (BMIMI), managed by the Austrian Research Promotion Agency (FFG), under project DOMINO (923575).

\bibliographystyle{splncs04}
\bibliography{main}

\input{Supplementary_Material/sec/supplementary_sections}

\end{document}

%% file: custom_commands.tex
\usepackage{dsfont}
\usepackage{etoolbox}

\newif\ifshowedits

\newcommand{\addeditor}[3]{%
  \definecolor{#1color}{rgb}{#3}
  \expandafter\newcommand\csname #1\endcsname[1]{%
  \ifshowedits
    {\color{#1color} ##1}%
  \else
    {##1}%
  \fi
  }%
  \expandafter\newcommand\csname #1rmk\endcsname[1]{%
  \ifshowedits
    {\color{#1color} {\bf [#2: ##1]}}
  \fi
  }%
  \expandafter\newcommand\csname #1rpl\endcsname[2]{%
  \ifshowedits
    {\color{#1color} ##1 \sout{##2}}
  \else
    {##1}
  \fi
  }%
}

\newcommand{\createtextvar}[1]{
  \expandafter\newcommand\csname #1\endcsname{%
  {\text{#1}}
}%
}

\newcommand{\mycomment}[1]{}

\newcommand{\calD}{{\cal D}}

\newcommand{\calO}{{\cal O}}
\newcommand{\calP}{{\cal P}}

\newcommand{\calX}{{\cal X}}
\newcommand{\calY}{{\cal Y}}

\newcommand{\br}{{\bf r}}

\newcommand{\bu}{{\bf u}}

\newcommand{\bx}{{\bf x}}

\newcommand{\bz}{{\bf z}}

\newcommand{\bI}{{\bf I}}

\newcommand{\bW}{{\bf W}}

\newcommand{\myparagraph}[1]{\medskip\noindent\textbf{#1}\hspace{0.2em plus 0.3em minus 0.2em}}

\definecolor{ours_orange}{HTML}{ff7f0e}
\definecolor{oodclassblue}{HTML}{0000ff}
\definecolor{idclassred}{HTML}{ff0000}

\definecolor{markred}{HTML}{CD0f01} %
\definecolor{markgreen}{HTML}{008836} %
\newcommand{\cmark}{{\ding{51}}}
\newcommand{\cmarkred}{\textcolor{markred}{\ding{51}}}

%% file: sec/00_abstract.tex
\begin{abstract}

Out-of-Distribution (OOD) detection remains a fundamental challenge for neural networks, whose predictions can be overconfident on inputs that deviate from the training distribution. 
Most post-hoc OOD detection methods derive scores from a single representation level (\eg, logits or penultimate features) or combine multiple layers via depth selection or OOD-calibrated weighting.
However, because OOD shifts are diverse, the most informative representation level can vary strongly across OOD types and domains, making fixed-layer choices and OOD-calibrated aggregation brittle.
In this paper, we propose PRISM (Projected Representation with Intermediate-layer Subspace Modeling), a model-agnostic post-hoc OOD detection method that models a unified multi-layer feature representation rather than aggregating independently scored layers. 
PRISM fuses intermediate and deep features into a single hierarchical embedding, estimates an in-distribution (ID) principal subspace, and then combines two complementary signals: (i) a class-conditional Mahalanobis distance in the projected subspace and (ii) the residual energy orthogonal to the learned manifold.
This simple design avoids OOD-tuned layer weighting while capturing both in-subspace semantic deviations and off-subspace anomalies. 
Across diverse benchmarks spanning natural images, medical imaging, and industrial visual inspection, PRISM achieves consistent state-of-the-art cross-domain OOD detection performance with a single default configuration across all evaluated domains and architectures. 
We further show that PRISM incurs minimal inference overhead, making it practical for real-world deployment.

\keywords{Out-of-Distribution Detection \and Intermediate Layer Feature Fusion}
\end{abstract}

%% file: sec/01_introduction.tex
\section{Introduction}
\label{sec:intro}

\input{figures/teaser}

Machine learning models have demonstrated strong performance across diverse domains~\cite{ramesh21a_zero_shot_text_to_image, jumper2021_highaccurate, Suhendar2022_fruit_quality_classification} and are now widely deployed in real-world systems.
Despite their strong in-distribution (ID) accuracy, such models can produce highly confident yet incorrect predictions for inputs that deviate substantially from the training distribution, creating a mismatch between confidence and correctness that may lead to silent and potentially safety-critical failures~\cite{Nguyen2014, moosavidezfooli17, hein2019relu}.
This is particularly concerning in high-consequence applications, where erroneous predictions may directly affect safety or operational reliability~\cite{hendrycks2021_ml_safety, hendrycks2022_ml_safety, mohseni2021_ml_safety}.
For example, in medical imaging~\cite{gutbrod2025_openmibood}, models may fail to deliver accurate diagnoses due to scanner variability or unfamiliar instruments, while in industrial inspection \cite{Bergmann_mvtecad_1, krumpl2026_iconic444} systems may misclassify foreign materials as food products or fail to detect damage to machinery or infrastructure during quality control.
These limitations motivate out-of-distribution (OOD) detection: identifying inputs outside the training distribution to improve deployment reliability.

Post-hoc OOD detection is particularly attractive in practice because it can be applied to pretrained models without retraining, thereby preserving ID accuracy and avoiding costly re-optimization. 
Accordingly, a wide range of post-hoc approaches have been proposed, leveraging predicted probabilities~\cite{hendrycks2016_ood_msp, Liu2023_GEN}, output logits~\cite{hendrycks2019_ood_mls, liu2020_ood_ebo}, or feature embeddings from the penultimate layer~\cite{lee2018_mahala, sun2022knnood, Ren2021_rmds}.

Our motivation arises from the observation that post-hoc OOD performance varies substantially across OOD type and application domains (see \cref{fig:teaser}) \cite{gutbrod2025_openmibood, krumpl2026_iconic444}. %
Recent benchmarks such as OpenMIBOOD~\cite{gutbrod2025_openmibood} and ICONIC-444~\cite{krumpl2026_iconic444} provide large-scale OOD evaluations in medical and industrial domains, respectively, and show that methods performing strongly on natural image benchmarks~\cite{zhang2024openood} often degrade markedly in these specialized settings.

A key driver of this variability is that post-hoc OOD detection methods typically rely on a single network layer for discrimination, yet the most informative layer depends strongly on the ID dataset and OOD type \cite{gutbrod2025_openmibood, Krumpl2024_ats}.
Consequently, most existing methods largely neglect information in intermediate layers of deep neural networks, implicitly assuming that the final representation alone is sufficient for OOD detection.
While deep layers emphasize representations highly specialized for the ID task, intermediate and shallow layers capture complementary information that can be critical for distinguishing diverse OOD characteristics, which may manifest differently as network depth increases.
Other methods exist, but prior multi-layer methods either select a single layer \cite{Yu2023_featurenorm}, aggregate layer-wise scores \cite{haroush2022_statframework, sastry20a_gram, Krumpl2024_ats}, or calibrate aggregation/hyperparameters using synthetic or held-out OOD data \cite{lee2018_mahala, Dong_2022_CVPR_nmd, Tang2024_cores}, introducing additional complexity and OOD dependence.
This raises a key question: how can intermediate representations be effectively integrated into a unified OOD detection framework that remains reliable across domains and OOD types without OOD-specific calibration or tuning?

To address this question, we propose \textbf{PRISM}, an architecture-agnostic post-hoc OOD detection method designed for cross-domain robustness. 
PRISM follows a simple fuse-then-model principle: instead of computing and aggregating layer-wise OOD scores, it first integrates intermediate representations from shallow to deep layers into a unified feature space and then performs statistical modeling once in that fused representation. 
The fused features are projected into a compact subspace learned solely from ID training data, and PRISM combines complementary distance- and residual-based signals into a single OOD score.
This design avoids layer selection, layer-wise aggregation heuristics, and OOD-dependent calibration while retaining the robustness benefits of statistical feature modeling.
Our main contributions are as follows:
\begin{itemize}
    \item 
    We show that layer-wise aggregation and OOD-based calibration is highly sensitive to the chosen validation OOD set, leading to unstable cross-domain behavior.
    
    \item 
    We introduce a unified multi-layer subspace modeling principle that eliminates layer selection, score aggregation, and OOD-dependent calibration, while retaining the robustness benefits of statistical feature modeling.

    \item 
    We evaluate PRISM across natural, medical, and industrial benchmarks under a shared configuration, achieving the best overall mean AUROC and the best average rank across datasets without OOD-based calibration.
\end{itemize}

%% file: figures/teaser.tex
\begin{figure}[t]
  \centering
  \begin{subfigure}[t]{0.56\textwidth}
    \centering
    \includegraphics[height=4.3cm]{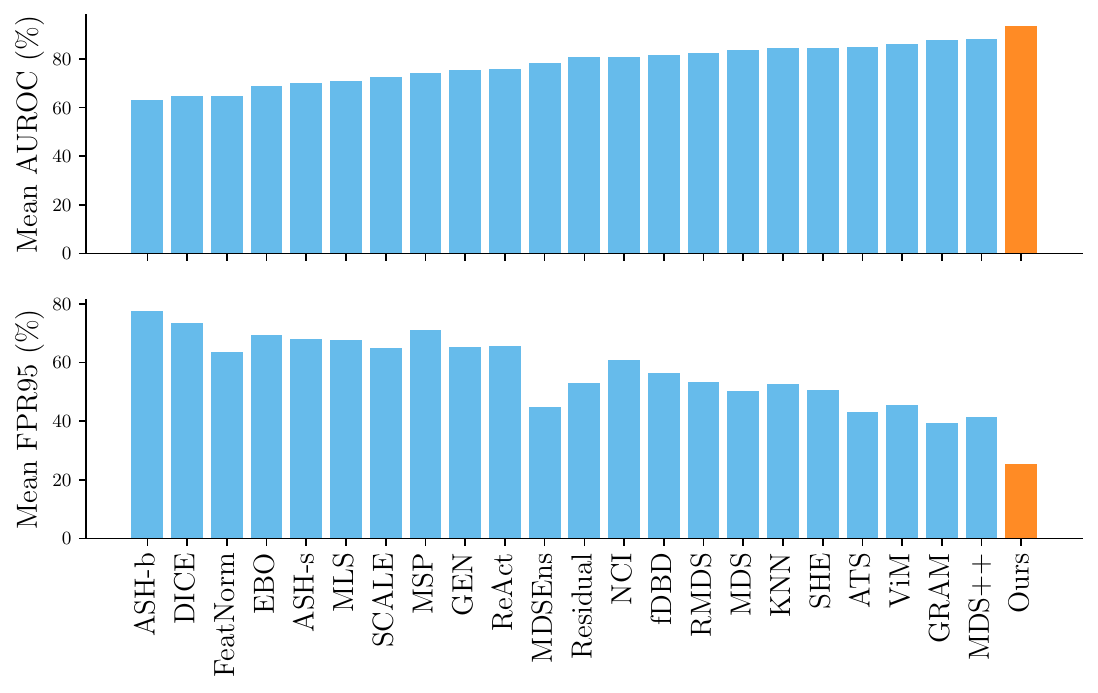}
    \caption{
    Aggregate performance across all benchmarks (top: AUROC$\uparrow$, bottom: FPR95$\downarrow$).
    }
    \label{fig:bars}
  \end{subfigure}\hfill
  \begin{subfigure}[t]{0.41\textwidth}
    \centering
    \includegraphics[height=4.3cm]{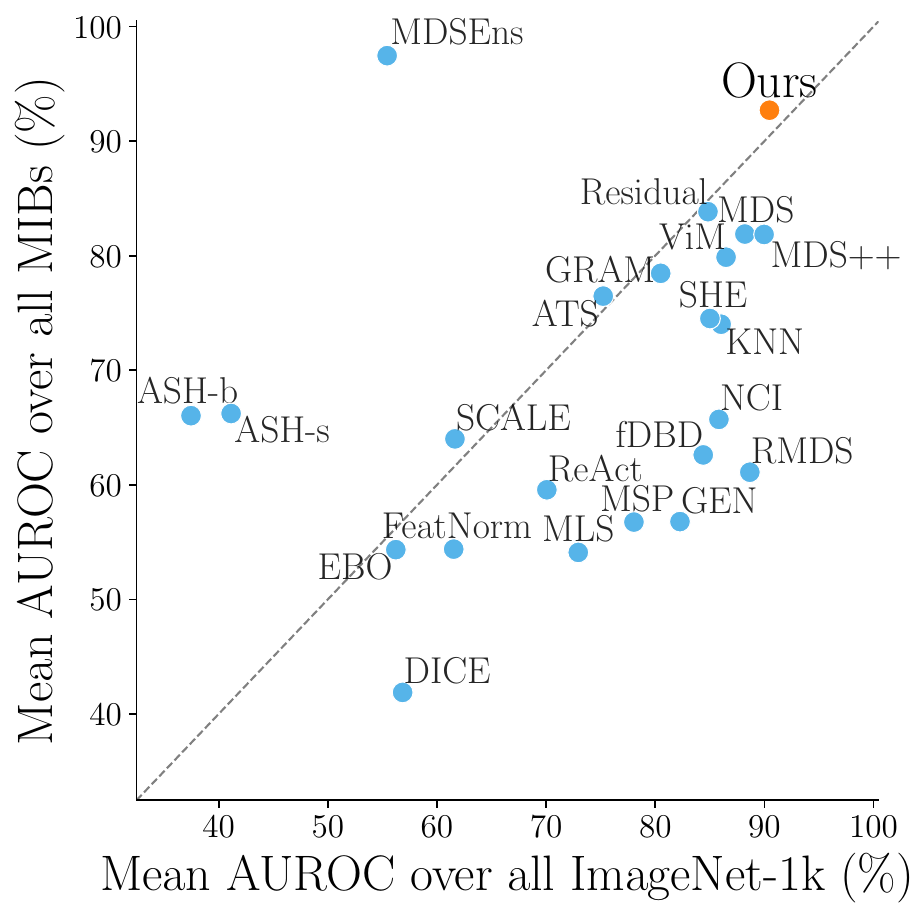}
    \caption{
    Cross-domain AUROC correlation across methods.
    }
    \label{fig:scatter}
  \end{subfigure}
  \caption{
  Comprehensive evaluation of post-hoc OOD detection across natural, medical, and industrial benchmarks.
  (a) Mean AUROC ($\uparrow$) and FPR95 ($\downarrow$) averaged over all datasets, OOD splits, and architectures.
  (b) Cross-domain correlation between ImageNet-1k \cite{deng2009_imagenet} (OpenOOD \cite{zhang2024openood}) and OpenMIBOOD \cite{gutbrod2025_openmibood} performance (AUROC).
  Our PRISM (\textcolor{ours_orange}{\textbf{orange}}) consistently exhibits strong performance across domains.
  }
  \label{fig:teaser}
\end{figure}

%% file: sec/02_related_work.tex
\section{Related Work}
\label{sec:related_work}

Out-of-distribution (OOD) detection methods are commonly categorized into training-based and post-hoc approaches \cite{yang2021_ood_survey, zhang2024openood}.
In this work, we focus on post-hoc OOD detection, which is particularly attractive in practice because it operates on pretrained models without retraining, thereby preserving ID accuracy and enabling scalable deployment~\cite{zhang2024openood, zhang2023_she, shiyu17}.

Early post-hoc methods derive OOD scores directly from the model output.
Probability-based approaches, such as the maximum softmax probability (MSP) \cite{hendrycks2016_ood_msp}, identify OOD samples by low prediction confidence.
Subsequent works leverage logits directly, including the maximum logit score (MLS) \cite{hendrycks2019_ood_mls} and energy-based OOD detection (EBO) \cite{liu2020_ood_ebo}.
Several model enhancement approaches further refine output-based detection by modifying activations or weights at inference time, for example, through activation clipping \cite{sun2021_ood_react} or sparsification \cite{sun2022dice, djurisic2023ash}.
Feature-based post-hoc methods instead model the distribution of deep representations and measure deviation using distance or density-based criteria.
Mahalanobis distance-based detection \cite{lee2018_mahala} fits class-conditional Gaussian distributions to penultimate-layer features and uses the minimum distance to the class means as an OOD score.
Several extensions refine this idea by normalizing or using relative scoring \cite{Ren2021_rmds, zhang2023_she}, while nearest-neighbor approaches \cite{sun2022knnood} exploit local structure in the embedding space.
ViM \cite{wang2022_ood_vim} complements feature distances with residual information derived from a principal subspace.
Kernel PCA \cite{fang2024_kernel} replaces linear PCA with non-linear kernel mappings in the penultimate feature space and uses reconstruction error or distance as an OOD score.
Overall, these approaches use the model output, the penultimate layer, or both, thereby ignoring information from intermediate layers.
PRISM instead explicitly leverages shallow-to-deep representations by modeling a fused feature hierarchy.

Existing multi-layer strategies typically leverage intermediate feature activations and compute layer-specific OOD scores, which are then aggregated.
For example, ATS \cite{Krumpl2024_ats} models activation statistics across layers, while GRAM \cite{sastry20a_gram} derives OOD scores from deviations in layer-wise Gram matrices capturing second-order feature correlations.
CORES \cite{Tang2024_cores} estimates consistency of representations across layers using statistical measures, whereas NAP \cite{Olber2023_binaryneuronactivationpattern} binarizes intermediate activations and computes Hamming distances to characterize deviations from ID patterns.
Similarly, multi-layer Mahalanobis-based approaches \cite{Anthony2023_maha} extend distance-based detection by estimating class-conditional statistics at multiple depths.
While intermediate representations capture complementary information, these strategies introduce additional design choices, aggregation heuristics, or calibration procedures that often rely on OOD or synthetic data and are sensitive to domain shifts.

Recent large-scale OOD detection benchmarks across medical \cite{gutbrod2025_openmibood} and industrial \cite{krumpl2026_iconic444} domains show that many post-hoc methods exhibit substantial performance variation outside the settings in which they were originally developed.
In particular, a large fraction of OOD detectors are designed and validated on CIFAR \cite{krizhevsky09cifar} or ImageNet-based \cite{deng2009_imagenet} benchmarks, where they often perform strongly, but show markedly reduced reliability when transferred to domain-specific scenarios.
This suggests that many methods implicitly adapt to the statistical characteristics of the benchmarks on which they are optimized \cite{zhang2024openood, krumpl2026_onemodel}.
In contrast, statistical and distance-based approaches that operate directly in feature space have been observed to be more stable across diverse settings, such as architectures and training methodologies, motivating methods that exploit feature geometry in a principled, domain-agnostic manner \cite{mueller2025_mahalanobispp, krumpl2026_onemodel}.

Overall, prior post-hoc methods demonstrate the effectiveness of feature-space geometry and multi-layer representations, but they differ substantially in how they incorporate hierarchical information.
In this work, we adopt a simple design choice that is underexplored in post-hoc OOD detection.
Rather than selecting individual layers or aggregating per-layer scores, we jointly leverage representations from shallow to deep layers and model the dominant ID structure in a low-dimensional subspace.
By deriving a single OOD score from this unified representation, our approach avoids aggregation heuristics, OOD-based calibration, and per-benchmark tuning while retaining the robustness benefits of statistical feature modeling.

%% file: sec/03_methodology.tex
\section{Method}
\label{sec:methodology}

We first formalize the OOD detection problem (\cref{sec:problem_statement}) and review methods based on Mahalanobis distance as a widely used post-hoc baseline (\cref{sec:mahalanobis}).
We then introduce PRISM (\cref{sec:prism}), which follows a \emph{fuse-then-model} design: it constructs a unified multi-layer representation, models the in-distribution (ID) structure in a principal subspace, and derives a single OOD score from projected Mahalanobis distance and residual energy.

\subsection{Problem Statement}
\label{sec:problem_statement}
We consider an image input space $\calX \subset \mathbb{R}^{C \times H \times W}$ and a set of class labels $\calY = \{1, \dots, K\}$.  
Let $f: \calX \rightarrow \mathbb{R}^{K}$ denote a classifier pretrained on an ID dataset $\calD_{\text{ID}} = \{(\bx_i, y_i)\}_{i=1}^{N}$, where samples are drawn \iid from the joint distribution $P_{\text{ID}}(X, Y)$ over $\calX_{\text{ID}} \times \calY$.  
We refer to $\calX_{\text{ID}} \subset \calX$ as the in-distribution (ID) input space and denote its associated label space by $\calY_{\text{ID}} = \calY$.
At deployment, the classifier may encounter inputs $\bx \in \calX$ that do not belong to $\calX_{\text{ID}}$, defining the out-of-distribution (OOD) input space $\calX_{\text{OOD}} = \calX \setminus \calX_{\text{ID}}$.
We target the \emph{semantic} OOD regime, where OOD samples belong to classes unseen during training, such that the corresponding label space $\calY_{\text{OOD}}$ satisfies $\calY_{\text{OOD}} \cap \calY_{\text{ID}} = \varnothing$.

OOD detection is formulated as a binary decision problem based on a scoring function $s: \calX \rightarrow \mathbb{R}$.  
An OOD detector $H: \calX \rightarrow \{\text{ID}, \text{OOD}\}$ is defined as:
\begin{equation}
\label{eqn:ood_detector_definition}
H(\bx) =
\begin{cases}
\text{ID}, & \text{if } s(\bx) \geq \tau \\
\text{OOD}, & \text{otherwise}
\end{cases},
\end{equation}
where $\tau$ is a threshold typically chosen to retain a high fraction of ID samples (\eg, $95\%$ true positive rate).

\subsection{Mahalanobis Distance for OOD Detection}
\label{sec:mahalanobis}

We briefly review Mahalanobis-based OOD detection \cite{lee2018_mahala} to establish notation and motivate PRISM's reformulation in a unified fused feature space.
Mahalanobis distance is a widely used post-hoc baseline that models ID features using class-conditional Gaussians. 
Let $\phi(\bx) \in \mathbb{R}^{M}$ denote a feature representation extracted from a pretrained classifier, typically from the penultimate layer. Given the ID training set $\{(\bx_i, y_i)\}_{i=1}^{N}$, the class-wise means $\boldsymbol{\widehat{\mu}}_c$ and shared covariance matrix $\boldsymbol{\widehat{\Sigma}}$ are estimated as
\begin{equation}
\boldsymbol{\widehat{\mu}}_c = \frac{1}{N_c} \sum_{i:y_i=c} \phi(\bx_i), \quad
\boldsymbol{\widehat{\Sigma}} = \frac{1}{N} \sum_{c=1}^{K} \sum_{i:y_i=c} (\phi(\bx_i)-\boldsymbol{\widehat{\mu}}_c)(\phi(\bx_i)-\boldsymbol{\widehat{\mu}}_c)^\top,
\end{equation}
where $N_c$ denotes the number of ID samples in class $c$ and $N$ the total number of ID samples.
The Mahalanobis distance of a test sample $\bx$ to class $c$ is 
\begin{equation}
d_{\text{Maha}}(\bx,c) = \left(\phi(\bx)-\boldsymbol{\widehat{\mu}}_c\right)^\top \boldsymbol{\widehat{\Sigma}}^{-1} \left(\phi(\bx)-\boldsymbol{\widehat{\mu}}_c\right),
\end{equation}
and the final OOD score is computed as the negative of the minimum distance to all class centroids.
Although effective, Mahalanobis-based detection is sensitive to the choice of feature representation, as also shown in \cref{fig:score_dist_different_layers}: applying it to shallow \vs deep layers yields markedly different ID-OOD separation across datasets.
A common remedy is to compute Mahalanobis scores at multiple layers and aggregate them, either by averaging or by learning weights (\eg, via logistic regression) using held-out or synthetic OOD data \cite{lee2018_mahala, Anthony2023_maha}.
However, while such aggregation can improve performance on specific benchmarks, it is often sensitive to the chosen OOD validation distribution and may not generalize beyond it.
As shown in \cref{fig:calib_ablation_main}, the layer weights learned by an OOD-calibrated aggregator (MDSEns \cite{lee2018_mahala}) vary substantially with the chosen OOD validation set: shallow layers are favored for some medical shifts (\eg, MIDOG \cite{aubreville2023_midog}), intermediate layers for others (\eg, PhaKIR \cite{RUECKERT2026_phakir}), and deeper layers for ImageNet-1k \cite{deng2009_imagenet}; even within the same domain, different near-OOD splits (\eg, SSB-Hard \cite{vaze2022openset} \vs NINCO \cite{bitterwolf2023_ninco}) lead to distinct optimal layer weights.
Together, \cref{fig:calib_ablation_main,fig:score_dist_different_layers} indicate that the preferred representation depth varies substantially across domains or OOD definitions.
This highlights a practical deployment trade-off: OOD-calibrated ensembles such as MDSEns can benefit from representative held-out OOD data for layer-weight calibration, but the resulting aggregation may be tied to the calibration distribution. 
To overcome the need for such calibration, we require an alternative that preserves information across depth without OOD-calibrated aggregation, which we realize in PRISM by fusing representations prior to statistical modeling.
With PRISM, we thus prioritize transferable ID-only deployment by estimating all statistics from ID data and using a fixed configuration when representative OOD data are unavailable.

\input{figures/score_dist_differnt_layers}
\input{figures/mdsens_calib_ablation}

\subsection{PRISM: Projected Representation with Intermediate-layer Subspace Modeling}
\label{sec:prism}
PRISM is based on a simple principle: instead of computing and aggregating layer-wise OOD scores, we first integrate intermediate representations across network depth into a single feature space and then apply statistical modeling once.
The method consists of three steps: (i) construction of a unified multi-layer representation, (ii) approximating its ID structure using principal component analysis (PCA), and (iii) deriving an OOD score from projected Mahalanobis distance and residual energy.
All feature statistics, PCA parameters, and class-conditional distributions are estimated using ID training data only. 
This formulation avoids selecting a single best layer, layer-wise aggregation heuristics, and OOD-dependent calibration. %

\myparagraph{Multi-Layer Feature Construction.}
Let $\{\phi^{(l)}(\bx)\}_{l=1}^{L}$ denote the feature maps extracted from $L$ selected intermediate layers of the classifier $f$.
The structure of these activations depends on the underlying architecture.
For convolutional neural networks (CNNs), $\phi^{(l)}(\bx) \in \mathbb{R}^{C_l \times H_l \times W_l}$ represents spatial feature maps with $C_l$ channels, whereas for Vision Transformers (ViTs), $\phi^{(l)}(\bx) \in \mathbb{R}^{T_l \times C_l}$ corresponds to a sequence of $T_l$ tokens of dimension $C_l$.
To unify these representations while preserving global context, we apply an architecture-agnostic pooling operator $\calP(\cdot)$ that aggregates over the non-channel dimensions:
\begin{equation}
\label{eqn:gap}
\bz^{(l)}(\bx) = \calP(\phi^{(l)}(\bx)).
\end{equation}
In practice, we use global average pooling for CNN feature maps and token averaging for transformer layers, resulting in a single vector per layer.
To prevent layers with larger activation magnitudes from dominating the joint representation, each pooled vector is $\ell_2$-normalized:
\begin{equation}
\label{eqn:l2_normalization}
\widetilde{\bz}^{(l)}(\bx) = \frac{\bz^{(l)}(\bx)}{\lVert \bz^{(l)}(\bx) \rVert_2}.
\end{equation}
Finally, the normalized representations are concatenated to form a single multi-layer feature vector:
\begin{equation}
\label{eqn:multi_layer_representation}
\bz(\bx) = \bigl[\widetilde{\bz}^{(1)}(\bx), \dots, \widetilde{\bz}^{(L)}(\bx)\bigr] \in \mathbb{R}^{D},
\end{equation}
where $D=\sum_{l=1}^{L} C_l$ is the dimensionality of the joint feature space.
This construction yields a single feature vector that jointly captures low-level structural and high-level semantic information.
This per-layer normalization also reduces architecture- and dataset-dependent scale differences before PCA and subsequent scoring, improving the stability of the downstream projected-distance and residual-energy terms.

\myparagraph{Subspace Modeling of ID Data.}
It is commonly assumed that neural network representations of ID data lie on a low-dimensional manifold and can be well-approximated by a low-dimensional subspace \cite{Zaeemzadeh2021_unionofsubspaces, wang2022_ood_vim}.
In PRISM, we apply this assumption to our fused multi-layer representation $\bz(\bx)$ and model its dominant ID variability with a linear subspace estimated by PCA.
Given fused representations $\{\bz(\bx_i)\}_{i=1}^{N}$ from $\calD_{\text{ID}}$, we compute the global mean $\boldsymbol{\mu} = \frac{1}{N}\sum_{i=1}^{N}\bz(\bx_i)$ and derive the principal subspace $\bW \in \mathbb{R}^{D \times d}$ from the top $d$ eigenvectors of the empirical covariance matrix.
This subspace captures the dominant modes of variability of the ID data while discarding low-variance directions that are weakly supported by the training data.

At inference, a feature vector $\mathbf{z}(\bx)$ is decomposed into its projection $\bu(\bx) = \bW^{\top}(\bz(\bx)-\boldsymbol{\mu})$ and its orthogonal residual:
\begin{equation}
\label{eqn:residual}
\mathbf{r}(\bx) = (\bz(\bx)-\boldsymbol{\mu}) - \bW\bu(\bx).
\end{equation}
Here, $\bu(\bx)$ captures the alignment with the dominant ID semantic patterns, while $\br(\bx)$ captures feature components orthogonal to the ID manifold.

\myparagraph{Projected Mahalanobis Distance.}
PRISM performs class-conditional Mahalanobis modeling directly in the unified projected space of the fused representation, yielding a single score without layer-wise scoring or aggregation.
Using the projected coordinates $\bu(\bx)$ of the ID training data, we estimate class-wise means $\boldsymbol{\widehat{\mu}}^{u}_c$ and a shared covariance matrix $\boldsymbol{\widehat{\Sigma}}^{u}$ in the PCA space. 
The projected Mahalanobis score for a sample $\bx$ is then defined as
\begin{equation}
\label{eqn:prj_maha_score}
s_{\text{proj}}(\bx) = \min_{c \in \calY} \left(\bu(\bx)-\boldsymbol{\widehat{\mu}}^{u}_c\right)^\top (\boldsymbol{\widehat{\Sigma}}^{u})^{-1} \left(\bu(\bx)-\boldsymbol{\widehat{\mu}}^{u}_c\right).
\end{equation}

\myparagraph{Residual Energy.}
Even if a sample is close to a class-conditional mean within the principal subspace, it may contain components that are not supported by the learned ID manifold.
We capture this via the residual energy
\begin{equation}
\label{eqn:res_energy}
s_{\text{res}}(\bx) = \lVert \br(\bx) \rVert_2^2 = \lVert \bz(\bx)-\boldsymbol{\mu} \rVert_2^2 - \lVert \bu(\bx) \rVert_2^2,
\end{equation}
where the equality follows from $\bW^\top\bW=\bI$.
The residual energy measures how much of the representation cannot be explained by the ID subspace, serving as a complementary signal that captures novel or unsupported feature directions.

\myparagraph{Out-of-Distribution Detection with PRISM.}
Both $s_{\text{proj}}(\bx)$ and $s_{\text{res}}(\bx)$ are nonnegative anomaly magnitudes.
PRISM combines them into a single score:
\begin{equation} 
s_{\text{PRISM}}(\bx) = -\, s_{\text{proj}}(\bx) \cdot s_{\text{res}}(\bx)
\end{equation}
This emphasizes samples that deviate from the ID structure both \emph{within} and \emph{outside} the principal subspace.
We negate the product so that larger $s_{\text{PRISM}}(\bx)$ indicates more ID-like samples, consistent with \cref{eqn:ood_detector_definition}.
In contrast to weighted sums or learned ensembles, this combination introduces no balancing coefficients and requires no OOD-dependent calibration.
We evaluate alternative score combinations in the supplementary material: additive fusion remains competitive but performs slightly worse than the multiplicative rule used in PRISM.

%% file: figures/score_dist_differnt_layers.tex
\begin{figure}[tb]
    \centering
    \begin{subfigure}[b]{0.3\linewidth}
        \centering
        \includegraphics[width=\linewidth]{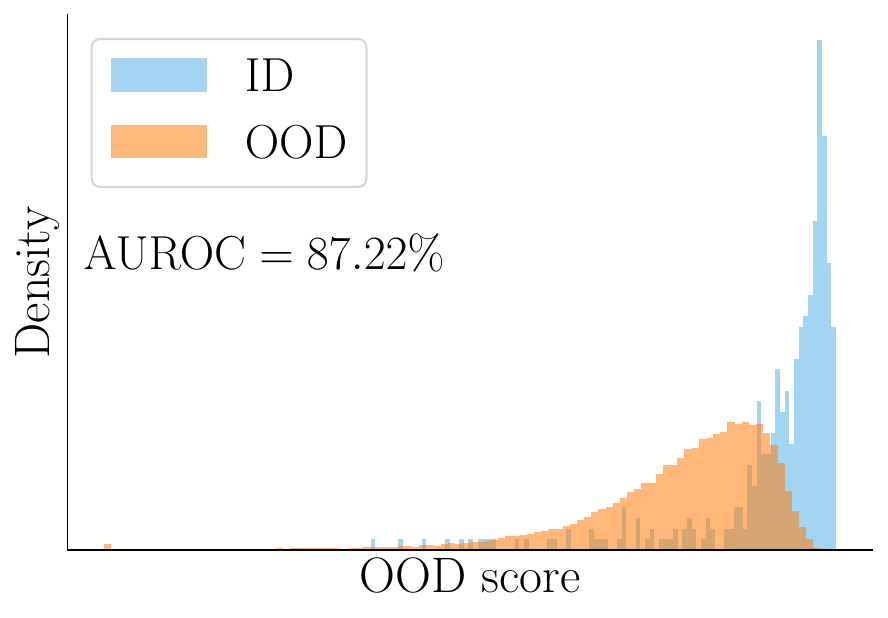}
        \caption{
            MDS++ (Shallow)
        }
    \end{subfigure}
    \hfill
    \begin{subfigure}[b]{0.3\linewidth}
        \centering
        \includegraphics[width=\linewidth]{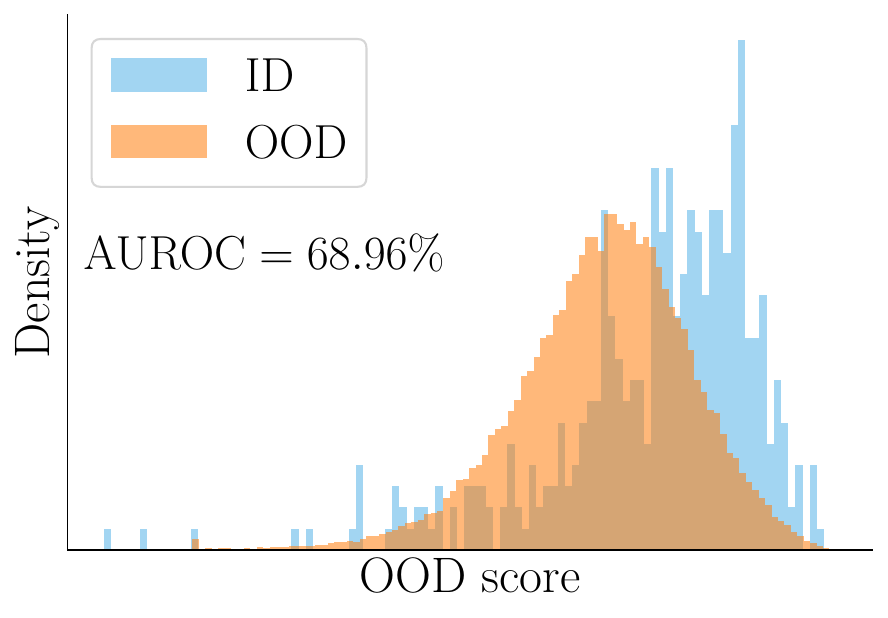}
        \caption{
            MDS++ (Deep)
        }
    \end{subfigure}
    \hfill
    \begin{subfigure}[b]{0.3\linewidth}
        \centering
        \includegraphics[width=\linewidth]{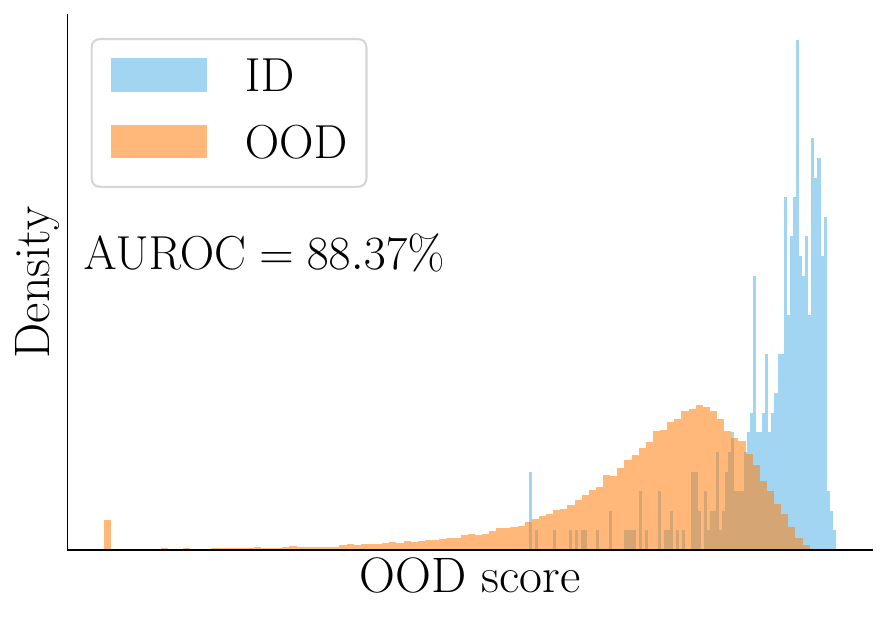}
        \caption{
            PRISM
        }
    \end{subfigure}
    
    \caption{
    Comparison of OOD score distribution across layer depths.
    Histograms represent data sampled from PhaKIR (ID) and Cholec80 (Near-OOD) using a ResNet-18. 
    (a) MDS++ applied to the first ResNet-18 block (shallow features).
    (b) MDS++ applied to the penultimate layer (deep features).
    (c) PRISM integrates the full architectural hierarchy (shallow to deep features).
    }
    \label{fig:score_dist_different_layers}
\end{figure}

%% file: figures/mdsens_calib_ablation.tex
\begin{figure}[tb]
    \centering
    \begin{subfigure}{0.325\linewidth}
        \centering
        \includegraphics[width=\linewidth]{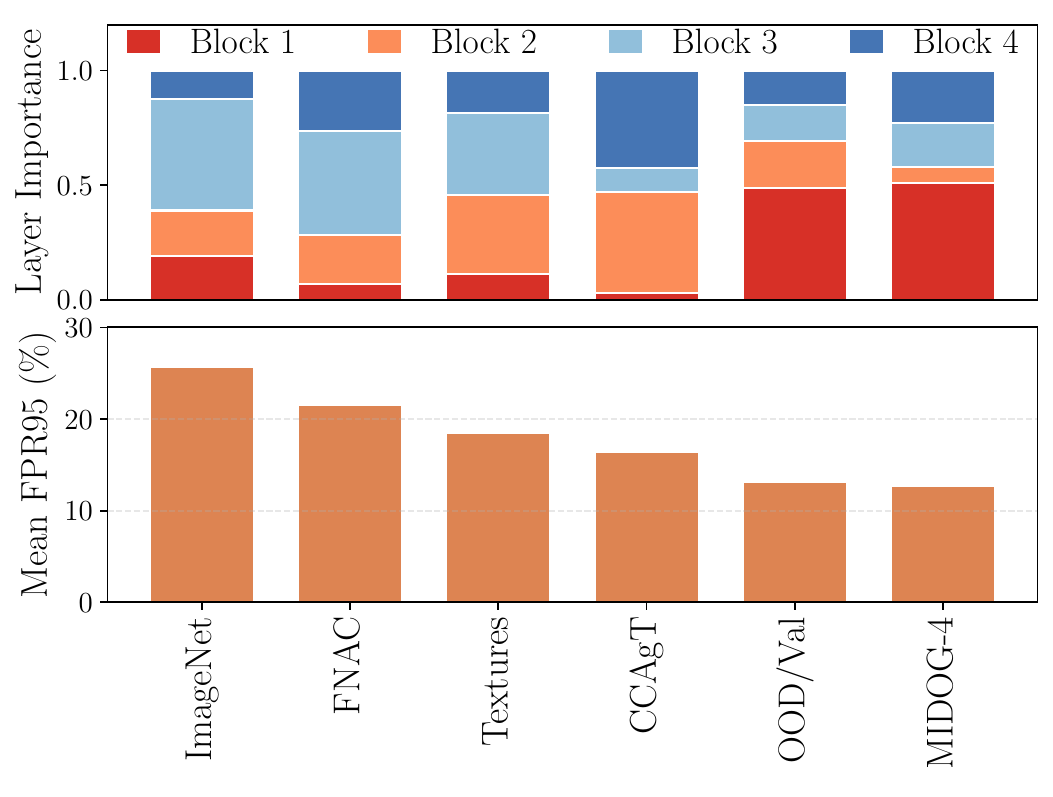}
        \caption{MIDOG (OpenMIBOOD)}
        \label{fig:ablation_midog}
    \end{subfigure}
    \hfill
    \begin{subfigure}{0.325\linewidth}
        \centering
        \includegraphics[width=\linewidth]{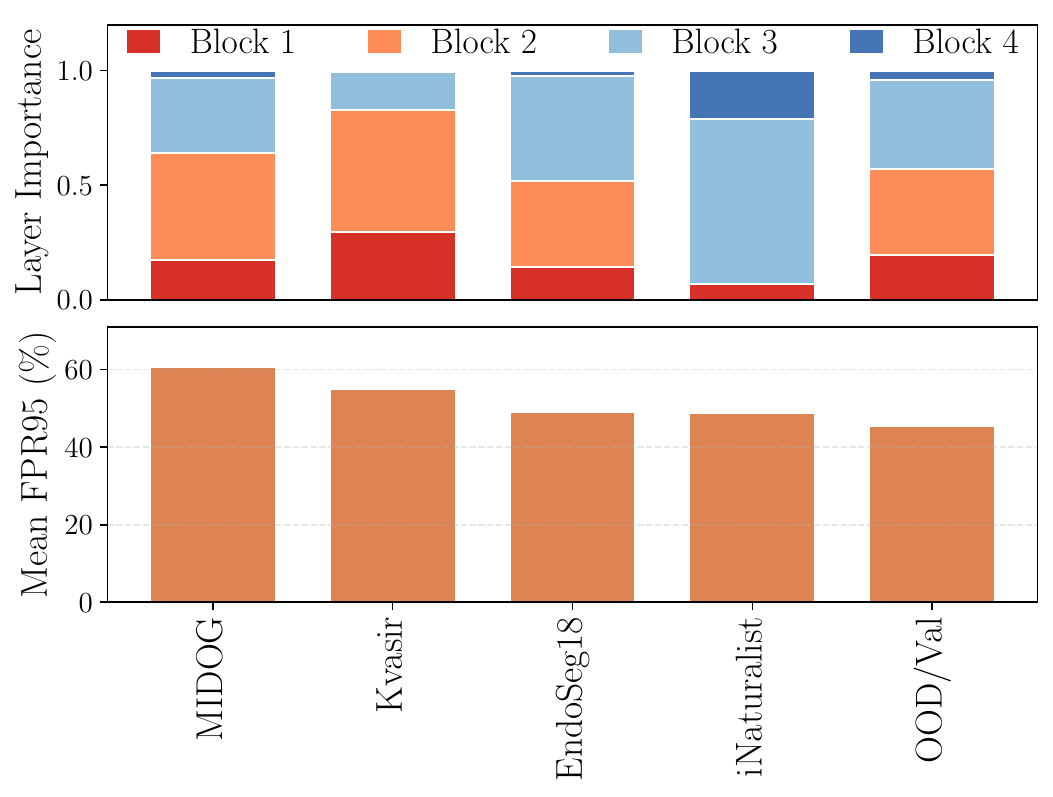}
        \caption{PhaKIR (OpenMIBOOD)}
        \label{fig:ablation_phakir}
    \end{subfigure}
    \hfill
    \begin{subfigure}{0.325\linewidth}
        \centering
        \includegraphics[width=\linewidth]{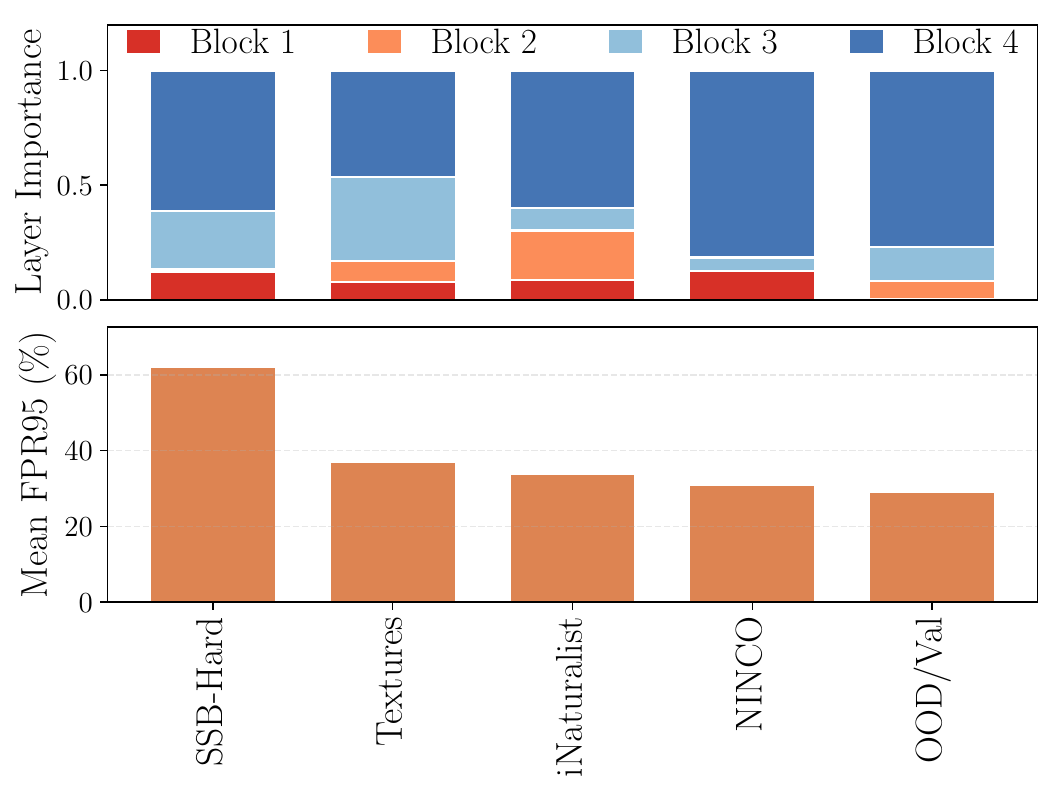}
        \caption{ImageNet-1k (OpenOOD)}
        \label{fig:ablation_imagenet}
    \end{subfigure}
    
    \caption{
    Sensitivity of layer aggregation to OOD validation data. 
    For three benchmark tasks: (a) MIDOG, (b) PhaKIR, and (c) ImageNet-1k, 
    we fit linear aggregation weights over ResNet blocks 1--4 using one OOD dataset as validation and evaluate the resulting model across all OOD test sets.
    The stacked bar plots (top) show the normalized layer weights learned for each calibration dataset, 
    while the bars (bottom) indicate the corresponding mean FPR95 averaged over all OOD datasets. 
    }
    \label{fig:calib_ablation_main}
\end{figure}

%% file: sec/04_experiments.tex
\section{Experiments}
\label{sec:experiments}
We evaluate PRISM for cross-domain robustness and architectural generalization.
\cref{sec:experimental_setup} describes the evaluation protocol and benchmarks. 
In \cref{sec:results}, we compare against 22 state-of-the-art post-hoc OOD detection methods across natural, medical, and industrial benchmarks \cite{zhang2024openood, gutbrod2025_openmibood, krumpl2026_iconic444}. 
\cref{sec:ablation_study} reports ablation results and analyzes the computational complexity.

\subsection{Experimental Setup}
\label{sec:experimental_setup}

\myparagraph{Benchmarks.}
We evaluate PRISM on three large-scale OOD detection benchmarks: OpenOOD \cite{yang2022openood, zhang2023openood, zhang2024openood}, OpenMIBOOD \cite{gutbrod2025_openmibood}, and ICONIC-444 \cite{krumpl2026_iconic444}. 
These benchmarks cover diverse acquisition settings and semantic granularity across natural images, industrial vision, histopathology, endoscopy, and MRI, enabling systematic evaluation of cross-domain robustness.

OpenOOD includes CIFAR-10/100 \cite{krizhevsky09cifar} and ImageNet-1k \cite{deng2009_imagenet} as ID datasets.
While CIFAR experiments are reported for completeness (see supplementary material), we focus on the large-scale ImageNet-1k task.
OpenMIBOOD evaluates cross-domain generalization across three medical tasks: histopathology (MIDOG \cite{aubreville2023_midog}), endoscopic video (PhaKIR \cite{RUECKERT2026_phakir}), and brain MRI (OASIS-3 \cite{lamontagne2019_oasis}). 
ICONIC-444 evaluates OOD detection in high-resolution industrial vision systems and comprises four ID tasks (Almond, Wheat, Kernels, and Food-grade).

OpenOOD and OpenMIBOOD define near- and far-OOD splits based on benchmark-specific protocols, while ICONIC-444 also introduces extreme-OOD splits. 
To enable consistent cross-domain comparison, we standardize the evaluation into near-, far-, and extreme-OOD categories across all benchmarks, strictly adhering to official splits and extending them only where not explicitly defined.
Across all tasks, we evaluate 21 OOD datasets spanning natural, medical, and industrial shifts (MNIST \cite{lecun-mnisthandwrittendigit-2010}, FMNIST \cite{xiao2017_fmnist}, Tiny ImageNet \cite{Torralba2008_tinyimagenet}, SVHN \cite{Netzer11_svhn}, Textures \cite{cimpoi14dtd}, Places365 \cite{zhou2017places}, SSB-Hard \cite{vaze2022openset}, NINCO \cite{bitterwolf2023_ninco}, iNaturalist \cite{Horn_2018_inat}, OpenImage-O \cite{wang2022_ood_vim}, CCAgT \cite{amorim2020_ccagt_1, atkinson_amorim_ccagt_2}, FNAC 2019 \cite{saikia2019_fnac}, Cholec80 \cite{twinanda2016endonet_cholec80}, EndoSeg15 \cite{bodenstedt2018_endoseg2015}, EndoSeg18 \cite{allan2020_endoseg2018}, Kvasir \cite{jha2020_kvasir}, CATARACTS \cite{ALHAJJ201924_cataracts}, ATLAS \cite{liew2022_atlas}, BraTS \cite{baid2021_brats1, bakas2017_brats2, menze2014_brats3}, MDS-Heart \cite{antonelli2022_mdsh1, simpson2019_mdsh2, tobon2015_mdsh3}, CHAOS \cite{CHAOSdata2019, KAVUR2021_chaos2}).
A detailed overview of ID tasks, OOD datasets, and their grouping into near-, far-, and extreme-OOD categories is provided in the supplementary material.

\myparagraph{Architectures.}
For the primary evaluation, we use the benchmark-defined backbones: ResNet-18~\cite{He2015_resnet} for CIFAR-10/100, PhaKIR, and ICONIC-444; ResNet-50 is used for ImageNet-1k and MIDOG; and task-specific R(2+1)D \cite{du2018_r21d} for the OASIS-3 brain MRI task.
To evaluate architectural robustness, we additionally consider ViT-B/16 \cite{dosovitskiy2021vit} and Swin-T \cite{liu2021Swin} on ImageNet-1k, and CCT-7/7x2 \cite{hassani202_cct} and ConvNeXt-S \cite{Liu_2022_CVPR} on ICONIC-444.

\myparagraph{Layer Selection Strategy \& Subspace Dimension.}
We use an architecture-specific layer-selection rule: for each backbone, we extract features from standard stage/block outputs that are uniformly spaced across depth (stage outputs for convolutional networks; regularly spaced blocks for transformers; layers listed in the supplement).
Unless stated otherwise, the PCA dimension is fixed to $d=512$ for all experiments, independent of dataset, domain, or backbone.
This avoids benchmark-specific tuning and keeps the protocol strictly ID-only across all evaluations.

\myparagraph{Evaluation Metrics.}
Following established protocol \cite{zhang2024openood, krumpl2026_iconic444, bitterwolf2023_ninco, hendrycks2016_ood_msp, liu2020_ood_ebo, sun2021_ood_react}, we report the false positive rate at recall $95\%$ (FPR95) and the area under the receiver operating
characteristic curve (AUROC).
When reporting aggregated results, we (i) first average over the OOD test sets within each OOD category/task and, then (ii) average across categories/tasks, giving each category/task equal weight.

\myparagraph{Evaluation Protocol.}
All inputs follow the benchmark's default preprocessing and are resized to the input shape required by the corresponding backbone.
We compare PRISM against 22 state-of-the-art post-hoc OOD detection methods. 
For methods requiring ID statistics, we use the training split of the ID dataset. 
For methods requiring hyperparameter selection, we follow each benchmark's official validation protocol.
MDSEns is the only method compared that requires OOD data for calibration; it is included for completeness (as a top-performing OpenMIBOOD \cite{gutbrod2025_openmibood} reference) and is marked accordingly in the results.
Additional implementation details are provided in the supplementary material.

\input{tables/main_openood_openmib_iconic_auroc}
\subsection{Benchmark Results}
\label{sec:results}

\myparagraph{Cross-Domain OOD Detection Performance.}
\cref{tab:main_openood_openmib_iconic_auroc} compares PRISM against 22 post-hoc OOD detection methods across all three domains.
PRISM achieves the best overall mean AUROC ($93.92\%$), improving over the second-best method (MDS++, $89.13\%$) by $4.79$ points.
It is also the only method that consistently remains among the top-performing approaches across all three domains.
To further assess robustness beyond dataset averages, \cref{fig:auroc_per_ood_group} reports AUROC averaged over near-, far-, and extreme-OOD categories. 
PRISM maintains high AUROC from near- to extreme-OOD, indicating robustness to increasing distributional shifts.

On OpenOOD, PRISM achieves the highest mean AUROC ($90.36\%$), surpassing both GRAM and MDS++, which each perform strongly only on specific subsets (\eg, CIFAR or ImageNet-1k).
On OpenMIBOOD, PRISM attains $94.86\%$, substantially improving over MDS++ ($84.09\%$), and emerges as the strongest ID-only alternative to the OOD-calibrated MDSEns ($98.41\%$).
On ICONIC-444, PRISM reaches $96.56\%$, closely matching the strongest industrial baseline (GRAM, $97.55\%$) and exceeding MDS++ ($96.12\%$).

In contrast, competing approaches exhibit pronounced domain-dependent variability. 
For example, MDSEns achieves excellent results on OpenMIBOOD but drops to $54.41\%$ on ImageNet-1k, consistent with the sensitivity of OOD-calibrated aggregation (\cf \cref{fig:calib_ablation_main}).
GRAM performs strongly on ICONIC-444 ($97.55\%$) yet degrades substantially on medical tasks ($75.12\%$).
Other methods, such as FeatNorm and DICE, show competitive performance on selected datasets but fail when transferred to other domains (\eg, DICE: $30.49\%$ on OASIS-3).
Overall, these observations suggest that many post-hoc detectors are implicitly tailored to particular benchmark regimes and rely on design choices (\eg, logit-only scoring, fixed-depth features, or OOD-calibrated aggregation) whose effectiveness varies across domains and OOD definitions.
Interestingly, we further observe that several logit-based methods (\eg, MLS, SCALE) degrade on extreme-OOD in some settings, underscoring that a larger semantic shift does not necessarily imply easier detection.
PRISM instead uses a simple ID-only model of a fused shallow-to-deep representation, reducing opportunities for domain-specific optimization and yielding more consistent cross-domain behavior.
This advantage persists beyond the mean AUROC: PRISM achieves the best overall average rank across datasets and shows significant improvements under paired per-dataset comparisons (see supplemental material, Fig.~1).
To better understand the remaining errors, we provide a qualitative analysis of false positives in the supplement.
Across OpenOOD, many failures arise from ID--OOD overlap, fine-grained semantic ambiguity, or context-dependent cases where the OOD object is visually close to, or only a small part of, an ImageNet-like scene.
In medical and industrial benchmarks, errors often reflect subtle near-OOD confusions caused by acquisition artifacts or shared coarse shape, texture, and color attributes between ID and OOD categories.
Overall, these failure modes are largely class-consistent and interpretable rather than arbitrary artifacts.

\subsection{Ablation Study}
\label{sec:ablation_study}

\input{tables/pca_and_component_ablation}
\myparagraph{Contribution of PRISM components.}
\cref{tab:pca_dim} analyzes the contribution of the projected Mahalanobis term ($s_{\text{proj}}$) and the residual energy term ($s_{\text{res}}$).
The two components exhibit domain-dependent behavior: on OpenOOD and ICONIC-444, the projected term performs strongest, whereas on OpenMIBOOD, the residual term dominates, indicating that OOD deviations manifest differently across domains.

Even when used individually, either term already improves over classical penultimate-layer baselines when computed on the unified multi-layer representation.
For comparison, averaged across all large-scale benchmarks, MDS obtains $87.00\%$ AUROC ($41.38\%$ FPR95), MDS++ $88.37\%$ ($39.44\%$), and ViM $85.59\%$ ($41.79\%$), all below the single-component PRISM variants.
This suggests that the gains do not arise merely from combining two scoring terms, but from modeling a fused representation that preserves complementary low-, mid-, and high-level cues.
The projected subspace model then derives the OOD score from ID feature geometry rather than a single layer, OOD-calibrated layer weights, or output confidence.

The two scores capture complementary geometric deviations. 
The projected Mahalanobis term detects violations of class-conditional structure within the principal ID subspace, while the residual term measures energy outside this subspace. 
Importantly, OOD deviations are not confined to residual directions: some samples lie outside the learned manifold, whereas others remain within it but violate class-conditional structure. 
In contrast to ViM, which models residual deviations at a single depth and combines them with a logit-based energy term for ID confidence, PRISM captures both in-subspace class-conditional deviations and off-subspace residual energy within a unified representation.

While each component can dominate on specific benchmarks, their combination achieves the highest overall average AUROC ($93.47\%$) and the most balanced FPR95 ($23.65\%$) across benchmarks, confirming that robust OOD detection requires modeling both types of deviation.

\myparagraph{Effect of Projection Dimension.}
We further analyze the sensitivity to the PCA dimensionality.
Performance remains stable across a broad range, with only minor variation in AUROC and FPR95 (see \cref{tab:pca_dim}). 
While the selected value of $d=512$ provides the best overall performance, the dependence on the projection dimension is generally weak, indicating that PRISM does not rely on precise tuning of subspace dimensionality.
This further supports the claim that PRISM operates as a stable cross-domain method with weak sensitivity to $d$.

\input{tables/imagenet_iconic_model_ablation_results}
\myparagraph{Architectural Generalization.}
To evaluate architectural robustness, we assess PRISM on diverse backbone families, including convolutional neural networks (ResNet-18/50 \cite{He2015_resnet}, ConvNeXt \cite{Liu_2022_CVPR}) and transformer-based models (ViT-B/16 \cite{dosovitskiy2021vit}, Swin-T \cite{liu2021Swin}).
As shown in \cref{tab:architectural_generalization}, PRISM consistently ranks among the top-performing methods across both OpenOOD and ICONIC-444, independent of the underlying architecture.

Although  \cref{tab:architectural_generalization} reports only the strongest methods per benchmark for clarity (full results are provided in the supplementary material), notable rank variations remain across architectures and domains. 
The leading competitors are predominantly geometric and statistic-based approaches operating on penultimate features, which tend to generalize better across architectural families, consistent with prior findings \cite{zhang2024openood, krumpl2026_onemodel}. 
Our results extend this observation by demonstrating stability not only across architectures but also across distinct domains.

Overall, PRISM maintains strong and consistent performance without specific tuning, confirming that unified multi-layer subspace modeling scales reliably across network designs.

\myparagraph{Computational Efficiency.}
In PRISM, intermediate feature maps are spatially reduced via global pooling, transforming tensors of size $B \times C_l \times H_l \times W_l$ into $B \times C_l$. 
This compression scales with $\calO(B H_l W_l)$ per layer but eliminates the dependence of the stacked representation $D = \sum_l C_l$ on the spatial resolution $(H, W)$. 
Consequently, all subsequent OOD scoring operations operate on resolution-independent feature vectors, ensuring that their computational cost does not increase with image size.

The additional computational cost of PRISM consists of three linear operations: 
(i) projection into a $d$-dimensional principal subspace with complexity $\calO(B D d)$, 
(ii) Mahalanobis distance computation in the reduced space with $\calO(B d K)$, and 
(iii) residual energy computation with $\calO(B D)$, where $B$ denotes the batch size and $K$ the number of classes. 
Since $d \ll D$ and all computations are implemented as matrix multiplications, these operations scale linearly with $B$ and contribute only a bounded relative overhead compared to the backbone forward pass, which itself scales with $\calO(B H W)$.

Overhead measures extra inference time beyond the backbone; MSP (max-softmax on logits) has near-zero overhead.
For a fair comparison, we optimize the MDS++ implementation analogously to PRISM and evaluate all methods under identical measurement settings.
Figure~\ref{fig:computational_runtime_overhead} reports the empirical runtime overhead on ImageNet-1k with ResNet-50.
Compared to MSP and optimized MDS++, PRISM introduces less than $1.0\%$ additional inference time for batch sizes $\geq 16$.
Inference time is measured over 1k iterations, preceded by 200 warm-up runs on a synthetic input.
These results indicate that multi-layer subspace modeling can be incorporated with negligible practical overhead while significantly boosting robustness.
\input{figures/ood_group_and_computational_overhead}

%% file: tables/main_openood_openmib_iconic_auroc.tex
\begin{table}[tb]
\centering\footnotesize
\setlength{\extrarowheight}{1.5pt}
\setlength{\tabcolsep}{4pt}
\caption{
AUROC ($\uparrow$) performance across diverse imaging domains.
Comparison of post-hoc OOD detection methods on natural (OpenOOD), medical (OpenMIBOOD), and industrial (ICONIC-444) benchmarks. 
Values represent the mean AUROC ($\%$) across the benchmark-specific OOD test sets.
Standard backbones are employed: ResNet-18 for CIFAR-10/100 (C10, C100), PhaKIR, and ICONIC-444; ResNet-50 for ImageNet-1k (IN-1k) and MIDOG; and R(2+1)D for OASIS-3. 
The \textbf{best} and \underline{second-best} results in each column are highlighted in bold and underlined, respectively.
* Results of MDSEns should be interpreted with caution, as discussed in \cref{sec:mahalanobis} and in prior benchmark analyses~\cite{gutbrod2025_openmibood}.
}
\label{tab:main_openood_openmib_iconic_auroc}
\resizebox{\textwidth}{!}{%
\begin{tabular}{lcccccccccccccc}
\toprule
\multirow{2}{*}{\textbf{Method}} & \multicolumn{4}{c}{\textbf{OpenOOD}} & \multicolumn{4}{c}{\textbf{OpenMIBOOD}} & \multicolumn{5}{c}{\textbf{ICONIC-444}} & \multirow{2}{*}{\textbf{Average}} \\
\cmidrule(lr){2-5}\cmidrule(lr){6-9}\cmidrule(lr){10-14}
 & C10 & C100 & IN-1k & \textbf{Mean} & MIDOG & PhaKIR & OASIS-3 & \textbf{Mean} & Almond & Wheat & Kernels & FG & \textbf{Mean} &  \\
\midrule
MSP~\cite{hendrycks2016_ood_msp} & $\phantom{0}87.97$ & $\phantom{0}78.55$ & $\phantom{0}77.94$ & $\phantom{0}81.49$ & $\phantom{0}64.23$ & $\phantom{0}39.99$ & $\phantom{0}69.23$ & $\phantom{0}57.82$ & $\phantom{0}75.04$ & $\phantom{0}39.86$ & $\phantom{0}86.74$ & $\phantom{0}90.61$ & $\phantom{0}73.06$ & $\phantom{0}70.79$ \\
MDS~\cite{lee2018_mahala} & $\phantom{0}88.66$ & $\phantom{0}63.83$ & $\phantom{0}89.83$ & $\phantom{0}80.77$ & $\phantom{0}78.14$ & $\phantom{0}73.66$ & $\phantom{0}98.73$ & $\phantom{0}83.51$ & $\phantom{0}97.00$ & $\phantom{0}98.64$ & $\phantom{0}96.65$ & $\phantom{0}88.96$ & $\phantom{0}95.31$ & $\phantom{0}86.53$ \\
MDSEns*~\cite{lee2018_mahala} & $\phantom{0}91.79$ & $\phantom{0}78.24$ & $\phantom{0}54.41$ & $\phantom{0}74.82$ & $\phantom{0}\textbf{96.87}$ & $\phantom{0}\textbf{98.49}$ & $\phantom{0}\underline{99.89}$ & $\phantom{0}\textbf{98.41}$ & $\phantom{0}93.38$ & $\phantom{0}98.44$ & $\phantom{0}95.82$ & $\phantom{0}87.59$ & $\phantom{0}93.81$ & $\phantom{0}89.01$ \\
GRAM~\cite{sastry20a_gram} & $\phantom{0}91.78$ & $\phantom{0}\underline{85.00}$ & $\phantom{0}85.61$ & $\phantom{0}\underline{87.46}$ & $\phantom{0}89.49$ & $\phantom{0}36.46$ & $\phantom{0}99.42$ & $\phantom{0}75.12$ & $\phantom{0}\underline{97.77}$ & $\phantom{0}98.70$ & $\phantom{0}97.82$ & $\phantom{0}\textbf{95.89}$ & $\phantom{0}\textbf{97.55}$ & $\phantom{0}86.71$ \\
EBO~\cite{liu2020_ood_ebo} & $\phantom{0}87.49$ & $\phantom{0}81.01$ & $\phantom{0}41.68$ & $\phantom{0}70.06$ & $\phantom{0}66.54$ & $\phantom{0}32.94$ & $\phantom{0}65.83$ & $\phantom{0}55.10$ & $\phantom{0}64.50$ & $\phantom{0}26.11$ & $\phantom{0}84.35$ & $\phantom{0}94.09$ & $\phantom{0}67.26$ & $\phantom{0}64.14$ \\
RMDS~\cite{Ren2021_rmds} & $\phantom{0}90.36$ & $\phantom{0}80.66$ & $\phantom{0}88.62$ & $\phantom{0}86.54$ & $\phantom{0}51.30$ & $\phantom{0}53.65$ & $\phantom{0}83.66$ & $\phantom{0}62.87$ & $\phantom{0}93.11$ & $\phantom{0}85.55$ & $\phantom{0}97.32$ & $\phantom{0}92.10$ & $\phantom{0}92.02$ & $\phantom{0}80.48$ \\
ReAct~\cite{sun2021_ood_react} & $\phantom{0}86.72$ & $\phantom{0}81.35$ & $\phantom{0}69.70$ & $\phantom{0}79.26$ & $\phantom{0}65.66$ & $\phantom{0}34.08$ & $\phantom{0}80.94$ & $\phantom{0}60.23$ & $\phantom{0}73.91$ & $\phantom{0}77.88$ & $\phantom{0}91.32$ & $\phantom{0}94.18$ & $\phantom{0}84.32$ & $\phantom{0}74.60$ \\
MLS~\cite{hendrycks2019_ood_mls} & $\phantom{0}87.43$ & $\phantom{0}80.95$ & $\phantom{0}70.21$ & $\phantom{0}79.53$ & $\phantom{0}65.71$ & $\phantom{0}32.94$ & $\phantom{0}65.87$ & $\phantom{0}54.84$ & $\phantom{0}65.08$ & $\phantom{0}26.11$ & $\phantom{0}84.35$ & $\phantom{0}93.65$ & $\phantom{0}67.30$ & $\phantom{0}67.22$ \\
ViM~\cite{wang2022_ood_vim} & $\phantom{0}91.67$ & $\phantom{0}78.38$ & $\phantom{0}85.54$ & $\phantom{0}85.19$ & $\phantom{0}79.12$ & $\phantom{0}64.46$ & $\phantom{0}99.48$ & $\phantom{0}81.02$ & $\phantom{0}97.36$ & $\phantom{0}98.28$ & $\phantom{0}97.85$ & $\phantom{0}84.68$ & $\phantom{0}94.54$ & $\phantom{0}86.92$ \\
KNN~\cite{sun2022knnood} & $\phantom{0}91.49$ & $\phantom{0}80.87$ & $\phantom{0}85.64$ & $\phantom{0}86.00$ & $\phantom{0}76.17$ & $\phantom{0}51.12$ & $\phantom{0}99.28$ & $\phantom{0}75.52$ & $\phantom{0}94.16$ & $\phantom{0}88.82$ & $\phantom{0}97.03$ & $\phantom{0}89.87$ & $\phantom{0}92.47$ & $\phantom{0}84.66$ \\
DICE~\cite{sun2022dice} & $\phantom{0}83.67$ & $\phantom{0}82.06$ & $\phantom{0}39.29$ & $\phantom{0}68.34$ & $\phantom{0}60.12$ & $\phantom{0}31.64$ & $\phantom{0}30.49$ & $\phantom{0}40.75$ & $\phantom{0}58.52$ & $\phantom{0}23.45$ & $\phantom{0}81.35$ & $\phantom{0}94.27$ & $\phantom{0}64.40$ & $\phantom{0}57.83$ \\
FeatNorm~\cite{Yu2023_featurenorm} & $\phantom{0}\underline{93.32}$ & $\phantom{0}73.56$ & $\phantom{0}73.38$ & $\phantom{0}80.09$ & $\phantom{0}39.76$ & $\phantom{0}61.58$ & $\phantom{0}62.56$ & $\phantom{0}54.63$ & $\phantom{0}43.41$ & $\phantom{0}34.25$ & $\phantom{0}28.43$ & $\phantom{0}37.89$ & $\phantom{0}35.99$ & $\phantom{0}56.90$ \\
ASH-b~\cite{djurisic2023ash} & $\phantom{0}72.99$ & $\phantom{0}81.08$ & $\phantom{0}26.17$ & $\phantom{0}60.08$ & $\phantom{0}68.94$ & $\phantom{0}51.10$ & $\phantom{0}84.37$ & $\phantom{0}68.14$ & $\phantom{0}66.86$ & $\phantom{0}33.33$ & $\phantom{0}74.22$ & $\phantom{0}91.12$ & $\phantom{0}66.38$ & $\phantom{0}64.87$ \\
ASH-s~\cite{djurisic2023ash} & $\phantom{0}79.48$ & $\phantom{0}82.48$ & $\phantom{0}69.01$ & $\phantom{0}76.99$ & $\phantom{0}65.04$ & $\phantom{0}53.94$ & $\phantom{0}87.59$ & $\phantom{0}68.86$ & $\phantom{0}66.94$ & $\phantom{0}61.40$ & $\phantom{0}84.67$ & $\phantom{0}\underline{94.70}$ & $\phantom{0}76.93$ & $\phantom{0}74.26$ \\
Residual~\cite{wang2022_ood_vim} & $\phantom{0}86.41$ & $\phantom{0}52.10$ & $\phantom{0}84.23$ & $\phantom{0}74.24$ & $\phantom{0}80.93$ & $\phantom{0}76.44$ & $\phantom{0}98.84$ & $\phantom{0}85.40$ & $\phantom{0}96.24$ & $\phantom{0}98.72$ & $\phantom{0}95.32$ & $\phantom{0}54.68$ & $\phantom{0}86.24$ & $\phantom{0}81.96$ \\
SHE~\cite{zhang2023_she} & $\phantom{0}90.00$ & $\phantom{0}81.40$ & $\phantom{0}79.51$ & $\phantom{0}83.64$ & $\phantom{0}76.00$ & $\phantom{0}55.71$ & $\phantom{0}96.79$ & $\phantom{0}76.17$ & $\phantom{0}93.12$ & $\phantom{0}93.99$ & $\phantom{0}95.90$ & $\phantom{0}91.89$ & $\phantom{0}93.72$ & $\phantom{0}84.51$ \\
GEN~\cite{Liu2023_GEN} & $\phantom{0}88.84$ & $\phantom{0}81.17$ & $\phantom{0}87.33$ & $\phantom{0}85.78$ & $\phantom{0}63.38$ & $\phantom{0}41.01$ & $\phantom{0}69.23$ & $\phantom{0}57.87$ & $\phantom{0}75.18$ & $\phantom{0}40.69$ & $\phantom{0}87.64$ & $\phantom{0}93.85$ & $\phantom{0}74.34$ & $\phantom{0}72.66$ \\
ATS~\cite{Krumpl2024_ats} & $\phantom{0}90.65$ & $\phantom{0}82.71$ & $\phantom{0}74.05$ & $\phantom{0}82.47$ & $\phantom{0}80.12$ & $\phantom{0}57.10$ & $\phantom{0}92.31$ & $\phantom{0}76.51$ & $\phantom{0}90.28$ & $\phantom{0}86.65$ & $\phantom{0}87.83$ & $\phantom{0}85.74$ & $\phantom{0}87.63$ & $\phantom{0}82.20$ \\
SCALE~\cite{xu2024scaling} & $\phantom{0}81.91$ & $\phantom{0}82.18$ & $\phantom{0}69.02$ & $\phantom{0}77.70$ & $\phantom{0}64.40$ & $\phantom{0}40.95$ & $\phantom{0}94.20$ & $\phantom{0}66.52$ & $\phantom{0}66.11$ & $\phantom{0}56.74$ & $\phantom{0}84.64$ & $\phantom{0}94.67$ & $\phantom{0}75.54$ & $\phantom{0}73.25$ \\
fDBD~\cite{liu2024_fdbd} & $\phantom{0}91.64$ & $\phantom{0}80.27$ & $\phantom{0}81.97$ & $\phantom{0}84.63$ & $\phantom{0}66.41$ & $\phantom{0}37.42$ & $\phantom{0}87.76$ & $\phantom{0}63.87$ & $\phantom{0}92.50$ & $\phantom{0}86.35$ & $\phantom{0}95.29$ & $\phantom{0}91.97$ & $\phantom{0}91.53$ & $\phantom{0}80.01$ \\
MDS++~\cite{mueller2025_mahalanobispp} & $\phantom{0}91.08$ & $\phantom{0}80.53$ & $\phantom{0}\underline{89.88}$ & $\phantom{0}87.16$ & $\phantom{0}80.72$ & $\phantom{0}71.69$ & $\phantom{0}99.88$ & $\phantom{0}84.09$ & $\phantom{0}97.04$ & $\phantom{0}\underline{98.79}$ & $\phantom{0}\underline{98.12}$ & $\phantom{0}90.54$ & $\phantom{0}96.12$ & $\phantom{0}\underline{89.13}$ \\
NCI~\cite{liu2025_nci} & $\phantom{0}89.09$ & $\phantom{0}81.18$ & $\phantom{0}84.60$ & $\phantom{0}84.96$ & $\phantom{0}73.34$ & $\phantom{0}44.74$ & $\phantom{0}79.38$ & $\phantom{0}65.82$ & $\phantom{0}85.54$ & $\phantom{0}80.21$ & $\phantom{0}95.38$ & $\phantom{0}94.50$ & $\phantom{0}88.91$ & $\phantom{0}79.90$ \\
\midrule
\textbf{PRISM (Ours)} & $\phantom{0}\textbf{94.50}$ & $\phantom{0}\textbf{86.62}$ & $\phantom{0}\textbf{89.96}$ & $\phantom{0}\textbf{90.36}$ & $\phantom{0}\underline{91.99}$ & $\phantom{0}\underline{92.60}$ & $\phantom{0}\textbf{99.98}$ & $\phantom{0}\underline{94.86}$ & $\phantom{0}\textbf{98.39}$ & $\phantom{0}\textbf{99.56}$ & $\phantom{0}\textbf{98.14}$ & $\phantom{0}90.13$ & $\phantom{0}\underline{96.56}$ & $\phantom{0}\textbf{93.92}$ \\
\bottomrule
\end{tabular}
}
\end{table}

%% file: tables/pca_and_component_ablation.tex
\begin{table}[tb]
\centering
\footnotesize
\caption{
Ablation of PRISM components and PCA dimensionality.
Performance is reported as AUROC ($\%$) ($\uparrow$) and FPR95 ($\%$) ($\downarrow$), averaged over all near-, far-, and extreme-OOD test sets across the three benchmarks.
For OpenOOD, results are shown on the large-scale ImageNet-1k task.
\textbf{(a)} Contribution of projected Mahalanobis distance ($s_{\text{proj}}$) and residual energy ($s_{\text{res}}$).
\textbf{(b)} Sensitivity to the PCA projection dimension.
\textbf{Best} and \underline{second-best} results are highlighted in bold and underlined, respectively.
}
\label{tab:ablation}

\begin{minipage}[t]{0.49\linewidth}
\centering
\footnotesize
\setlength{\extrarowheight}{1.5pt}
\setlength{\tabcolsep}{3pt}
(a) Component ablation.
\resizebox{\linewidth}{!}{
    \input{tables/component_ablation_groupfirst_taskbalanced}

}
\end{minipage}
\hfill
\begin{minipage}[t]{0.49\linewidth}
\centering
\centering
\footnotesize
\setlength{\extrarowheight}{1.5pt}
\setlength{\tabcolsep}{3pt}
(b) PCA dimensionality ablation.
\label{tab:pca_dim}
\resizebox{\linewidth}{!}{
    \input{tables/pca_ablation_groupfirst_taskbalanced}

}
\end{minipage}

\end{table}

%% file: tables/component_ablation_groupfirst_taskbalanced.tex
        \begin{tabular}{cccccccccc}
        \toprule
       \multirow{3}{*}{$s_{\text{proj}}$} & \multirow{3}{*}{$s_{\text{res}}$} & \multicolumn{2}{c}{\textbf{OpenOOD}} & \multicolumn{2}{c}{\textbf{OpenMIBOOD}} & \multicolumn{2}{c}{\textbf{ICONIC-444}} & \multicolumn{2}{c}{\textbf{Average}} \\
        \cmidrule(lr){3-4}\cmidrule(lr){5-6}\cmidrule(lr){7-8}\cmidrule(lr){9-10}
        &  & AUROC & FPR95 & AUROC & FPR95 & AUROC & FPR95 & AUROC & FPR95 \\
        &  & $\uparrow$ & $\downarrow$ & $\uparrow$ & $\downarrow$ & $\uparrow$ & $\downarrow$ & $\uparrow$ & $\downarrow$ \\
        \midrule
    $\checkmark$ &  & $\phantom{0}\textbf{89.94}$ & $\phantom{0}\underline{38.62}$ & $\phantom{0}91.42$ & $\phantom{0}32.32$ & $\phantom{0}\underline{95.21}$ & $\phantom{0}\underline{16.05}$ & $\phantom{0}\underline{92.19}$ & $\phantom{0}\underline{29.00}$ \\
 & $\checkmark$ & $\phantom{0}76.39$ & $\phantom{0}58.20$ & $\phantom{0}\textbf{95.72}$ & $\phantom{0}\textbf{17.21}$ & $\phantom{0}93.82$ & $\phantom{0}20.30$ & $\phantom{0}88.64$ & $\phantom{0}31.90$ \\
$\checkmark$ & $\checkmark$ & $\phantom{0}\underline{89.88}$ & $\phantom{0}\textbf{35.83}$ & $\phantom{0}\underline{94.86}$ & $\phantom{0}\underline{21.10}$ & $\phantom{0}\textbf{95.67}$ & $\phantom{0}\textbf{14.02}$ & $\phantom{0}\textbf{93.47}$ & $\phantom{0}\textbf{23.65}$ \\

        \bottomrule
        \end{tabular}

%% file: tables/pca_ablation_groupfirst_taskbalanced.tex
        \begin{tabular}{cccccccccc}
        \toprule
        \multirow{3}{*}{\textbf{PCA-Dim}} & \multicolumn{2}{c}{\textbf{OpenOOD}} & \multicolumn{2}{c}{\textbf{OpenMIBOOD}} & \multicolumn{2}{c}{\textbf{ICONIC-444}} & \multicolumn{2}{c}{\textbf{Average}} \\
        \cmidrule(lr){2-3}\cmidrule(lr){4-5}\cmidrule(lr){6-7}\cmidrule(lr){8-9}
          & AUROC & FPR95 & AUROC & FPR95 & AUROC & FPR95 & AUROC & FPR95 \\
          & $\uparrow$ & $\downarrow$ & $\uparrow$ & $\downarrow$ & $\uparrow$ & $\downarrow$ & $\uparrow$ & $\downarrow$ \\
 \midrule
    128 & $\phantom{0}88.94$ & $\phantom{0}39.81$ & $\phantom{0}94.83$ & $\phantom{0}22.91$ & $\phantom{0}95.56$ & $\phantom{0}14.54$ & $\phantom{0}93.11$ & $\phantom{0}25.75$ \\
 256 & $\phantom{0}89.67$ & $\phantom{0}36.93$ & $\phantom{0}\textbf{95.24}$ & $\phantom{0}\underline{21.19}$ & $\phantom{0}95.59$ & $\phantom{0}14.46$ & $\phantom{0}\textbf{93.50}$ & $\phantom{0}\underline{24.19}$ \\
 512 & $\phantom{0}\underline{89.88}$ & $\phantom{0}\textbf{35.83}$ & $\phantom{0}\underline{94.86}$ & $\phantom{0}\textbf{21.10}$ & $\phantom{0}\underline{95.67}$ & $\phantom{0}\underline{14.02}$ & $\phantom{0}\underline{93.47}$ & $\phantom{0}\textbf{23.65}$ \\
 768 & $\phantom{0}\textbf{89.97}$ & $\phantom{0}\underline{36.09}$ & $\phantom{0}93.47$ & $\phantom{0}25.32$ & $\phantom{0}\textbf{95.67}$ & $\phantom{0}\textbf{13.91}$ & $\phantom{0}93.04$ & $\phantom{0}25.11$ \\

        \bottomrule
        \end{tabular}

%% file: tables/imagenet_iconic_model_ablation_results.tex
\begin{table}[tb]
  \centering
  \caption{
  Performance comparison of post-hoc OOD detection methods across architectures.
  \textbf{(a)} OpenOOD (ImageNet-1k) using ResNet-50, Swin-T, and ViT-B/16.
  \textbf{(b)} ICONIC-444 using ResNet-18, CCT-7/7x2, and ConvNeXt-S.
  Methods are selected based on the mean AUROC across architectures within each benchmark.
  }
  \label{tab:architectural_generalization}
  \begin{minipage}[t]{0.49\columnwidth}
    \centering
    \footnotesize
    \setlength{\extrarowheight}{1.5pt}
    \setlength{\tabcolsep}{3pt}
    (a) ImageNet-1k (OpenOOD)
    \resizebox{\linewidth}{!}{%

\input{tables/openood_models_group_first}

    }
  \end{minipage}
  \hfill
  \begin{minipage}[t]{0.49\columnwidth}
    \centering
    \footnotesize
    \setlength{\extrarowheight}{1.5pt}
    \setlength{\tabcolsep}{3pt}
    (b) ICONIC-444
    \resizebox{\linewidth}{!}{%

\input{tables/iconic444_models_group_first}

    }
  \end{minipage}

\end{table}

%% file: tables/openood_models_group_first.tex
        \begin{tabular}{lcccccccc}
        \toprule
        \textbf{Method} & \multicolumn{2}{c}{\textbf{ResNet-50}} & \multicolumn{2}{c}{\textbf{Swin-T}} & \multicolumn{2}{c}{\textbf{ViT-B-16}} & \multicolumn{2}{c}{\textbf{Average}} \\
        \cmidrule(lr){2-3}\cmidrule(lr){4-5}\cmidrule(lr){6-7}\cmidrule(lr){8-9}
        & AUROC & FPR95 & AUROC & FPR95 & AUROC & FPR95 & AUROC & FPR95 \\
        & $\uparrow$ & $\downarrow$ & $\uparrow$ & $\downarrow$ & $\uparrow$ & $\downarrow$ & $\uparrow$ & $\downarrow$ \\
        \midrule
    MDS++~\cite{mueller2025_mahalanobispp} & $\phantom{0}\underline{89.88}$ & $\phantom{0}42.34$ & $\phantom{0}\underline{88.85}$ & $\phantom{0}52.15$ & $\phantom{0}\textbf{89.52}$ & $\phantom{0}\textbf{40.37}$ & $\phantom{0}\underline{89.42}$ & $\phantom{0}\underline{44.96}$ \\
RMDS~\cite{Ren2021_rmds} & $\phantom{0}88.62$ & $\phantom{0}47.71$ & $\phantom{0}88.02$ & $\phantom{0}55.59$ & $\phantom{0}88.38$ & $\phantom{0}53.87$ & $\phantom{0}88.34$ & $\phantom{0}52.39$ \\
MDS~\cite{lee2018_mahala} & $\phantom{0}89.83$ & $\phantom{0}\underline{36.38}$ & $\phantom{0}87.09$ & $\phantom{0}57.57$ & $\phantom{0}86.17$ & $\phantom{0}52.81$ & $\phantom{0}87.70$ & $\phantom{0}48.92$ \\
ViM~\cite{wang2022_ood_vim} & $\phantom{0}85.54$ & $\phantom{0}41.42$ & $\phantom{0}88.15$ & $\phantom{0}\underline{49.50}$ & $\phantom{0}83.45$ & $\phantom{0}50.45$ & $\phantom{0}85.71$ & $\phantom{0}47.12$ \\
KNN~\cite{sun2022knnood} & $\phantom{0}85.64$ & $\phantom{0}51.46$ & $\phantom{0}84.78$ & $\phantom{0}66.59$ & $\phantom{0}86.17$ & $\phantom{0}50.95$ & $\phantom{0}85.53$ & $\phantom{0}56.33$ \\
NCI~\cite{liu2025_nci} & $\phantom{0}84.60$ & $\phantom{0}52.40$ & $\phantom{0}86.15$ & $\phantom{0}61.21$ & $\phantom{0}84.62$ & $\phantom{0}66.59$ & $\phantom{0}85.13$ & $\phantom{0}60.07$ \\
\midrule
\textbf{PRISM} & $\phantom{0}\textbf{89.96}$ & $\phantom{0}\textbf{35.36}$ & $\phantom{0}\textbf{90.91}$ & $\phantom{0}\textbf{30.60}$ & $\phantom{0}\underline{88.76}$ & $\phantom{0}\underline{41.51}$ & $\phantom{0}\textbf{89.88}$ & $\phantom{0}\textbf{35.83}$ \\

        \bottomrule
        \end{tabular}

%% file: tables/iconic444_models_group_first.tex
 \begin{tabular}{lcccccccc}
        \toprule
        \textbf{Method} & \multicolumn{2}{c}{\textbf{ResNet-18}} & \multicolumn{2}{c}{\textbf{CCT-7/7x2}} & \multicolumn{2}{c}{\textbf{ConvNeXt-S}} & \multicolumn{2}{c}{\textbf{Average}} \\
        \cmidrule(lr){2-3}\cmidrule(lr){4-5}\cmidrule(lr){6-7}\cmidrule(lr){8-9}
        & AUROC & FPR95 & AUROC & FPR95 & AUROC & FPR95 & AUROC & FPR95 \\
        & $\uparrow$ & $\downarrow$ & $\uparrow$ & $\downarrow$ & $\uparrow$ & $\downarrow$ & $\uparrow$ & $\downarrow$ \\
        \midrule
    GRAM~\cite{sastry20a_gram} & $\phantom{0}\textbf{97.55}$ & $\phantom{0}\phantom{0}\textbf{8.41}$ & $\phantom{0}\underline{88.75}$ & $\phantom{0}\underline{34.64}$ & $\phantom{0}\textbf{96.29}$ & $\phantom{0}\textbf{15.93}$ & $\phantom{0}\underline{94.19}$ & $\phantom{0}\underline{19.66}$ \\
MDS++~\cite{mueller2025_mahalanobispp} & $\phantom{0}96.12$ & $\phantom{0}11.88$ & $\phantom{0}86.12$ & $\phantom{0}43.49$ & $\phantom{0}92.60$ & $\phantom{0}30.72$ & $\phantom{0}91.61$ & $\phantom{0}28.69$ \\
ViM~\cite{wang2022_ood_vim} & $\phantom{0}94.54$ & $\phantom{0}22.11$ & $\phantom{0}82.56$ & $\phantom{0}49.84$ & $\phantom{0}92.99$ & $\phantom{0}28.44$ & $\phantom{0}90.03$ & $\phantom{0}33.46$ \\
MDS~\cite{lee2018_mahala} & $\phantom{0}95.31$ & $\phantom{0}15.65$ & $\phantom{0}79.01$ & $\phantom{0}57.24$ & $\phantom{0}95.06$ & $\phantom{0}19.77$ & $\phantom{0}89.80$ & $\phantom{0}30.89$ \\
SHE~\cite{zhang2023_she} & $\phantom{0}93.72$ & $\phantom{0}23.65$ & $\phantom{0}82.29$ & $\phantom{0}49.38$ & $\phantom{0}88.27$ & $\phantom{0}39.21$ & $\phantom{0}88.09$ & $\phantom{0}37.41$ \\
Residual~\cite{wang2022_ood_vim} & $\phantom{0}86.24$ & $\phantom{0}34.92$ & $\phantom{0}82.93$ & $\phantom{0}52.94$ & $\phantom{0}93.54$ & $\phantom{0}28.81$ & $\phantom{0}87.57$ & $\phantom{0}38.89$ \\
\midrule
\textbf{PRISM} & $\phantom{0}\underline{96.56}$ & $\phantom{0}\phantom{0}\underline{9.87}$ & $\phantom{0}\textbf{94.65}$ & $\phantom{0}\textbf{16.08}$ & $\phantom{0}\underline{95.79}$ & $\phantom{0}\underline{16.12}$ & $\phantom{0}\textbf{95.67}$ & $\phantom{0}\textbf{14.02}$ \\

        \bottomrule
        \end{tabular}

%% file: figures/ood_group_and_computational_overhead.tex
\begin{figure}[t]
\centering

\begin{minipage}[t]{0.48\columnwidth}
  \centering
  \includegraphics[width=\linewidth]{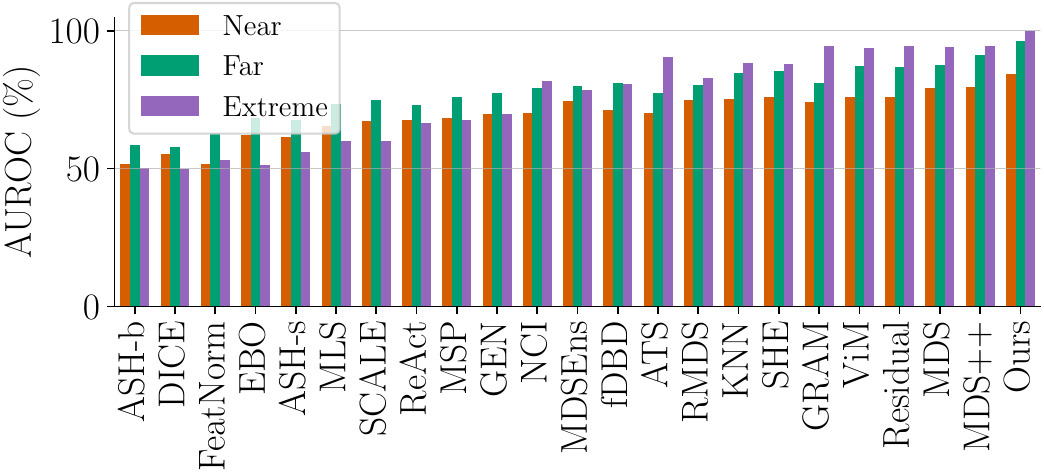}
  \captionof{figure}{
  AUROC ($\uparrow$) averaged over near-, far-, and extreme-OOD categories across all three benchmarks. 
  Methods are ordered by overall performance. 
  All values are reported as percentages.
  }
  \label{fig:auroc_per_ood_group}
\end{minipage}
\hfill
\begin{minipage}[t]{0.48\columnwidth}
  \centering
  \includegraphics[width=\linewidth]{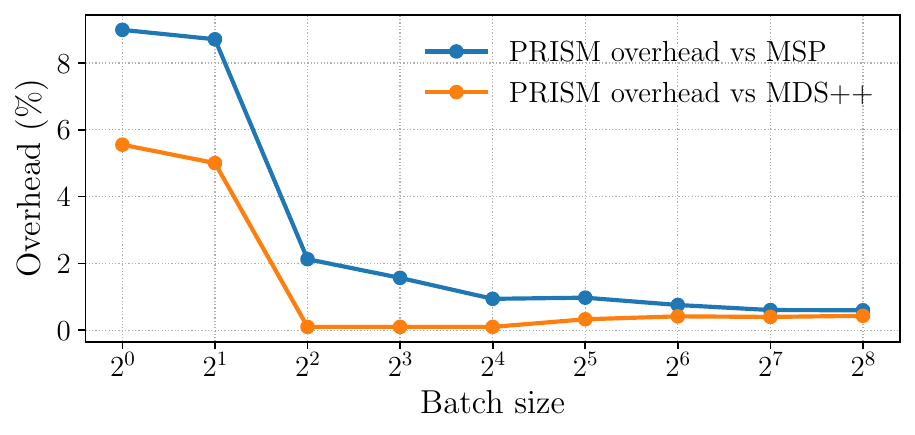}
  \captionof{figure}{
  Relative runtime overhead of PRISM on ImageNet-1k with ResNet-50.
  Runtime overhead as a percentage ($\%$) of the total inference time compared to the baseline MSP and MDS++ methods. 
  }
  \label{fig:computational_runtime_overhead}
\end{minipage}

\end{figure}

%% file: sec/05_conclusion.tex
\section{Conclusion}
\label{sec:conclusion}

We show that Out-of-Distribution (OOD) detection is influenced by representation depth: neither a single layer nor layer-wise score aggregation is consistently reliable across domains, and aggregation strategies calibrated on one OOD setting do not robustly transfer. %
This motivates preserving information across representation depths rather than relying on benchmark-specific depth preferences.

PRISM addresses this instability by fusing shallow-to-deep representations into a unified feature space and performing statistical modeling only after fusion, combining projected class-conditional Mahalanobis distance with orthogonal residual energy.
As a result, PRISM avoids per-benchmark layer selection, aggregation heuristics, and OOD-based calibration, and delivers consistent state-of-the-art performance across natural, medical, and industrial benchmarks.

We use a fixed, architecture-aware layer sampling strategy and global pooling to ensure a consistent ID-only evaluation protocol.
This design abstracts away spatial or token layouts, yielding a unified representation without architecture-specific fusion choices. 
More task-specific alternatives, such as tuned layer weights or spatially adaptive pooling, could improve individual benchmarks but would introduce additional priors and hyperparameters that may reduce transferability.
Our minimal fusion design, therefore, isolates the central hypothesis of PRISM: unified multi-layer ID geometry can improve cross-domain OOD robustness without task-specific tuning or OOD-dependent calibration.
Developing adaptive depth or pooling strategies that retain this transferability is an interesting direction for future work.

%% file: Supplementary_Material/sec/supplementary_sections.tex
\appendix
\maketitlesupplementary

\crefname{appendix}{Appendix}{Appendices}
\Crefname{appendix}{Appendix}{Appendices}
\crefalias{section}{appendix}

This supplementary document provides detailed experimental settings and implementation details (\cref{sec:experimental_details}) and comprehensive results that support the analyses presented in the main paper (\cref{sec:detailed_results}).

\section{Experimental Details}
\label{sec:experimental_details}

\subsection{Implementation Details}

\myparagraph{Software Setup.}
Our implementation builds upon the OpenOOD framework~\cite{yang2022openood, zhang2023openood, zhang2024openood}, which provides standardized evaluation protocols for post-hoc OOD detection. 
We integrate OpenMIBOOD~\cite{gutbrod2025_openmibood}, released as a fork of OpenOOD, and incorporate ICONIC-444~\cite{krumpl2026_iconic444} within the same framework to establish a unified evaluation pipeline. 
This ensures consistent data handling, preprocessing, and experimental protocols across natural, medical, and industrial domains.

In addition, we incorporate several post-hoc OOD detection methods, including FeatNorm~\cite{Yu2023_featurenorm}, ATS~\cite{Krumpl2024_ats}, and MDS++~\cite{mueller2025_mahalanobispp}. 
We further extend the framework with support for models from the \texttt{timm} repository~\cite{rw2019timm}, enabling evaluation of modern architectures such as ConvNeXt \cite{Liu_2022_CVPR} and Vision Transformers \cite{dosovitskiy2021vit}.

\myparagraph{Computational Environment.}
All experiments were conducted on two servers equipped with an Intel(R) Core(TM) i9-9900X CPU (3.50\,GHz) and NVIDIA GPUs, 
including three RTX A4000, one RTX 3080, and one RTX 2080 Ti. 
All runtime and throughput benchmarks reported in this paper were measured on an RTX A4000.
The systems run Ubuntu 22.04 with Python 3.11.3 and PyTorch 2.7.1, using CUDA 12.6 and cuDNN 9.5.1.

\subsection{Benchmark Setup}

\myparagraph{OOD Category Details.}
\cref{tab:benchmark_configuation} summarizes the in-distribution (ID) tasks and their corresponding near-, far-, and extreme-OOD splits across all benchmarks.

For OpenOOD and OpenMIBOOD, near- and far-OOD categories strictly follow the official benchmark definitions.
ICONIC-444 also defines an extreme-OOD category corresponding to large (semantically unrelated) cross-domain shifts.

For benchmarks without explicitly defined extreme-OOD categories, we introduce extreme-OOD datasets to standardize evaluation across near-, far-, and extreme-OOD categories. 
These extensions do not modify any official benchmark categories; they simply enable a consistent robustness analysis, since extreme-OOD, often assumed to be easiest due to a larger semantic shift, can still be challenging for some OOD detectors.

For 3D medical data (OASIS-3), 2D OOD datasets are converted to grayscale and replicated along the channel dimension to match the network input format.

\input{Supplementary_Material/tables/benchmark_config}

\input{Supplementary_Material/tables/model_source}

\myparagraph{Medical Preprocessing (OpenMIBOOD / OASIS-3).}
We use the official OpenMIBOOD preprocessing for OASIS-3 without modification.
This includes resampling all scans to $1\text{mm}^{3}$ isotropic voxel spacing and skull-stripping with HD-BET \cite{isensee2019_hdbet}.
Any additional processing is limited to format compatibility with the evaluated backbones: inputs are resized to the required spatial resolution, and 2D OOD datasets are converted to grayscale and replicated across channels to match the network input format.

\myparagraph{Architectures.}
\cref{tab:model_overview} summarizes the models and weight sources.
The officially released checkpoints for OpenOOD\footnote{https://github.com/Jingkang50/OpenOOD} (CIFAR-10/100) and OpenMIBOOD\footnote{https://github.com/remic-othr/OpenMIBOOD} are used. 
For ImageNet-1k, we employ supervised pretrained models from \emph{torchvision} \cite{Paszke19_pytorch} and \emph{timm} \cite{rw2019timm}. 
For ICONIC-444, models are trained following the official benchmark protocol.

\myparagraph{OOD Detection Methods.}
All experiments are conducted on OpenOOD, OpenMIBOOD, and ICONIC-444 using a unified evaluation pipeline. 
Because most OOD detection methods were originally developed and tuned on natural-image benchmarks (CIFAR \cite{krizhevsky09cifar}, ImageNet-1k \cite{deng2009_imagenet}), we perform a benchmark-specific hyperparameter search for methods that require hyperparameter tuning.

We follow the official hyperparameter selection procedures of each benchmark, using the provided ID validation and OOD validation splits where applicable.
To ensure fair comparison across benchmarks, we merge the benchmark-specific search ranges such that all parameter values from the original protocols are covered. 
The resulting search spaces and calibration requirements are summarized in \cref{tab:ood_hyperparameter}.

We restrict our primary evaluation to post-hoc methods that do not require real OOD validation data for calibration. 
The only exception is MDSEns~\cite{lee2018_mahala}, which requires OOD data to calibrate layer aggregation weights; we include it for completeness as a strong baseline in OpenMIBOOD. 
FeatNorm~\cite{Yu2023_featurenorm} generates synthetic OOD samples via jigsaw augmentation applied to ID data and therefore does not require external OOD datasets. 
All remaining methods are calibrated exclusively on ID training data.
For ImageNet-1k, we use $20\%$ of the official training split for ID calibration, following prior work (\eg, \cite{wang2022_ood_vim, Krumpl2024_ats}), to reduce computational cost while maintaining stable calibration.

For PRISM, PCA is fitted on the ID calibration split using class-balanced subsampling (up to 50k samples). 
In the resulting PCA space, we estimate class-conditional means and a shared covariance matrix, using ridge regularization $\lambda = 10^{-3}$, and retain $d=512$ components unless noted otherwise. 
For ResNet-like (including 3D ResNet variants) and ConvNeXt backbones, we extract the final output of each stage (\ie, each spatial-resolution block), yielding four stage outputs in total.
For ViT and Swin-T backbones, we use a uniformly spaced subset of blocks (every second block), resulting in 7 and 5 tapped representations, respectively; for the compact transformer, we tap all blocks (7 representations).

\input{Supplementary_Material/tables/ood_methods_hyper_params}

\section{Detailed Results}
\label{sec:detailed_results}

This section discusses the per-dataset and per-category results reported in the supplementary tables and complements the main paper analysis.
Overall, the detailed breakdown confirms three consistent observations: (i) there is no single OOD detection method that is uniformly best across all benchmarks and OOD categories, (ii) PRISM remains consistently among the top-performing methods across highly heterogeneous settings, and (iii) the benchmark-dependent layer-depth trends from the main paper are clearly reflected in the detailed rankings.

\myparagraph{No Universally Best Baseline Across Domains.}
\cref{tab:cifar10_full_supp,tab:cifar100_full_supp,tab:imagenet_full_supp_all_models} report full per-dataset results for OpenOOD (across all evaluated ID tasks and architectures).
For OpenMIBOOD, we report per-task results averaged at the OOD-category level in \cref{tab:openmibood_full_per_ood_category}.
For ICONIC-444, where tasks are defined over a shared underlying dataset, we report task-averaged results in \cref{tab:iconic444_taskavg_models}.

A consistent takeaway from the OpenOOD results (\cref{tab:cifar10_full_supp,tab:cifar100_full_supp,tab:imagenet_full_supp_all_models}) is that strong performance on one natural-image setting does not necessarily transfer to others: method rankings can change substantially across OOD datasets and across near-, far-, and extreme- OOD categories.
In other words, methods that are highly competitive on a subset of near-OOD datasets can be overtaken on far- or extreme-OOD datasets, and vice versa.

On CIFAR-10/100 (see \cref{tab:cifar10_full_supp,tab:cifar100_full_supp}), PRISM achieves the strongest aggregated performance (CIFAR-10: $94.50\%$ AUROC; CIFAR-100: $86.62\%$ AUROC), improving over competitive baselines in the overall mean.
At the same time, the per-dataset columns show that no method achieves the best performance across all OOD datasets: different baselines obtain isolated wins on specific near- or far-OOD datasets, confirming substantial method re-ranking at the dataset level.

A similar pattern holds on ImagNet-1k (see \cref{tab:imagenet_full_supp_all_models}).
In the ResNet-50 setting, PRISM achieves a strong overall mean of $89.96\%$ AUROC and $35.36\%$ FPR95, and remains competitive across the diverse OOD sets.
However, the detailed breakdown also highlights depth dependence on challenging near-OOD: on SSB-Hard, PRISM attains $68.82\%$ AUROC / $82.70\%$ FPR95, while deep-feature baselines such as RMDS achieve higher AUROC and substantially lower FPR95 on this specific near-OOD set (\eg, $75.47\%$ AUROC; $70.78\%$ FPR95).
Conversely, PRISM is very strong on several far- and extreme-OOD sets, illustrating that the best representation depth and score vary with OOD characteristics.
Overall, these results reinforce our central motivation: when deployment OOD characteristics are unknown \emph{a priori}, integrating information across representation depth provides a robust default and reduces sensitivity to OOD-specific depth preferences.

For OpenMIBOOD, \cref{tab:openmibood_full_per_ood_category} reports task-wise results for MIDOG, PhaKIR, and OASIS-3, broken down by the three OOD-categories, along with task-wise means and the overall mean.
In contrast to the natural-image benchmarks, the detailed results on these medical tasks exhibit depth dependence that aligns with the main manuscript: OOD-relevant cues are often captured in shallow-to-intermediate representations, and methods that exploit multi-depth information perform better.
Consistent with this, PRISM achieves strong performance across all three tasks under an ID-only protocol, and MDSEns is also among the strongest methods.
At the same time, the per-task breakdown highlights substantial cross-task re-ranking: for example, GRAM performs competitively on MIDOG ($83.20\%$ AUROC) but drops sharply on PhaKIR ($36.46\%$ AUROC).
Finally, as emphasized in the main paper, MDSEns requires real OOD validation data to calibrate aggregation weights and should therefore be interpreted separately from purely ID-calibrated post-hoc methods.

For ICONIC-444, \cref{tab:iconic444_taskavg_models} reports results averaged over the four tasks (Almond, Wheat, Kernels, Food-grade) and broken down by OOD-category and specific OOD test sets, for three backbones (ResNet-18, CCT-7/7x2, ConvNeXt-S).
Here, we observe behavior similar to the medical benchmark, with methods that leverage intermediate layers also showing better performance. 
While GRAM can be extremely strong on ResNet-18 (\eg, $97.55$ AUROC; $8.41\%$ FPR95), it drops substantially on CCT-7/7x2 ($88.75\%$ AUROC;  $34.64\%$), indicating sensitivity to architecture.
In contrast, PRISM remains consistently strong across all three backbones, yielding the best architecture-average AUROC of $95.67\%$.

\input{Supplementary_Material/tables/cifar10_full_supp}

\input{Supplementary_Material/tables/cifar100_full_supp}

\input{Supplementary_Material/tables/openood_full_supp_all_models_v2}

\input{Supplementary_Material/tables/iconic444_taskavg_models}

\myparagraph{Rank-based Statistical Comparison.}
To quantify performance consistency across heterogeneous settings, we additionally report the mean AUROC rank of each method across $\text{N}=140$ OOD datasets (see \cref{fig:mean_rank_ci95}).
Each evaluation unit corresponds to a single OOD dataset, yielding (N=140) paired units across all methods.
Within each unit, methods are ranked by AUROC (higher is better), and mean ranks are computed across units.

A Friedman test rejects the null hypothesis of equal performance across methods ($\chi^2 = 1185.59$, $p \ll 0.001$).
PRISM achieves the best mean rank ($3.04$, $95\%$ CI $[2.48,\ 3.61]$), substantially ahead of the next-best methods GRAM (mean rank $6.29$, $95\%$ CI $[5.33,\ 7.25]$) and MDS++ (mean rank $6.33$, $95\%$ CI $[5.74,\ 6.92]$).
Using a Nemenyi post-hoc test at $\alpha=0.05$, the critical difference is ($\text{CD}=2.08$); the rank gaps between PRISM and both GRAM ($\Delta=3.25$) and MDS++ ($\Delta=3.29$) exceed this threshold, indicating statistically significant improvements in average rank.
These results support our claim that PRISM is not only strong on average but also consistently top-ranked across diverse OOD datasets.

\input{Supplementary_Material/figures/mean_rank_ci95}

\myparagraph{Qualitative Error Analysis Across Benchmarks}
To complement the quantitative results, we qualitatively inspect high-confidence false accepts (OOD samples assigned high ID-likeness) across all three benchmark families.
For the ImageNet-based OpenOOD setting, we report examples from SSB-hard, Textures, and NINCO (see \cref{fig:openood_ssb_hard_samples,fig:openood_textures_samples,fig:openood_ninco_samples}); for the medical and industrial settings, we show near- and far-OOD examples on OpenMIBOOD/PhaKIR and ICONIC-444/almond (see \cref{fig:openmibood_phakir_samples,fig:iconic444_almond_samples}).

Across the ImageNet-based OpenOOD test sets, many of the most confident false accepts are consistent with either (i) ID--OOD overlap/contamination or (ii) inherently fine-grained semantic ambiguity, rather than a clear distribution shift. 
Prior work on NINCO reports substantial overlap for commonly used ImageNet OOD sets such as SSB-hard and Textures and motivates the construction of a more carefully curated alternative \cite{bitterwolf2023_ninco}.
This effect is most pronounced on SSB-hard, where we observe a larger performance gap, with multiple OOD samples that are visually near-indistinguishable from canonical ImageNet concepts (\eg, \emph{touch football} $\rightarrow$ \emph{football helmet}, \emph{cruise liner} $\rightarrow$ \emph{ocean liner}, \emph{hot air balloon} $\rightarrow$ \emph{balloon}). 
The SSB-hard example \emph{black olives} further illustrates a granularity mismatch between OOD labels and ImageNet semantics: although the annotation focuses on an ingredient, the overall image depicts a typical \emph{pizza} instance, so the detector's assignment to the ImageNet class \emph{pizza} is semantically reasonable. 
Even on the Textures dataset (where performance is strong), false accepts often correspond to texture attributes that naturally occur in ImageNet classes (\eg, \emph{striped}/\emph{freckled} aligning with \emph{zebra}/\emph{banana}), highlighting that OOD can become ill-defined when the ID taxonomy is comparatively coarse while OOD labels are highly fine-grained. 
NINCO is explicitly curated to reduce shortcut cues, and indeed many remaining false accepts are more genuinely fine-grained or context-dependent; nevertheless, we still observe occasional overlap-like cases in this large-scale, scene-centric setting (\eg, \emph{waffles} $\rightarrow$ \emph{waffle iron} when a prominent waffle is present, or \emph{hippopus} $\rightarrow$ \emph{snorkel} when a snorkel is visible in the background), and several of the hardest NINCO examples correspond to highly fine-grained categories and small/secondary objects.
In addition, some NINCO false accepts arise because the nominally OOD object occupies only a small region of the image (\eg, \emph{glass of milk} mapped to the ImageNet class \emph{ice lolly}), while the overall scene is dominated by the person interacting with the object (a child drinking milk \vs eating an ice lolly), making the ID/OOD decision inherently context-dependent.

In contrast, on OpenMIBOOD/PhaKIR and ICONIC-444/almond, the selected near- and far-OOD false accepts are largely free of obvious ID contamination and instead reflect real near-OOD confusions: subtle appearance changes, blur and acquisition artifacts, or highly similar shape/texture/color between ID and OOD categories (see \cref{fig:openmibood_phakir_samples,fig:iconic444_almond_samples}). Beyond fine-grained visual similarity, these errors also reveal how the underlying ID task structure influences OOD behavior. 
For PhaKIR, the near-OOD sets remain laparoscopic scenes but introduce distribution shifts via different instruments and procedures (Cholec80, EndoSeg15), whereas the far-OOD set includes a different medical field without surgical tools (Kvasir). 
Consequently, borderline errors often reflect whether the image exhibits instrument-specific cues learned from the ID task: images with weak or absent instrument evidence tend to be absorbed by the broad \emph{non-instrument} mode, while images with tool-like structures may be mapped to the closest instrument-related ID class.
A similar effect appears on ICONIC-444/almond: among the seven ID classes, \emph{almond shell fragment} is visually more distinctive than the fine-grained almond categories, and \emph{almond blanched} is the only predominantly white class, making OOD samples that share these coarse attributes harder to separate from ID. 
Taken together with the ImageNet-based analysis above, these results highlight that different application domains induce qualitatively different failure modes (overlap/ambiguity \vs fine-grained confusions and category-structure effects), reinforcing the importance of evaluating OOD detectors on a diverse set of benchmarks for reliable interpretation. 
Across all settings, PRISM's false accepts remain largely class-consistent and interpretable, supporting its robustness as a post-hoc detector.

\input{Supplementary_Material/figures/openood_imagenet_samples}
\input{Supplementary_Material/figures/openood_textures_samples}
\input{Supplementary_Material/figures/openood_ninco_samples}
\input{Supplementary_Material/figures/openmibood_phakir_samples}
\input{Supplementary_Material/figures/iconic444_almond_samples}

\myparagraph{Feature Fusion Ablations and Efficiency.}
For completeness, we report an ablation of PRISM and representative baselines, where all numbers denote the average over all large-scale benchmarks (AUROC/FPR95).
While strong methods such as MDS++ and ViM achieve $(88.37\%/39.44\%)$ and $(85.59\%/41.79\%)$, PRISM improves to $(93.47\%/23.65\%)$, indicating that aggregating shallow-to-deep information into a single hierarchical embedding yields a more stable ID geometry than single-layer or logit-based scores (\eg, MSP $(70.60\%/74.10\%)$, EBO $(60.70\%/78.22\%)$).
In addition, we ablate the pre-PCA feature normalization: removing per-layer $\ell_2$ normalization before stacking and PCA substantially degrades performance to $(87.87\%/36.82\%)$, showing that normalization is critical for stable fusion across layers with different feature scales.
Ablations confirm that both fusion and subspace modeling contribute: a Mahalanobis detector on the fused vector is slightly weaker $(92.81\%/24.84\%)$, and a per-layer MDS++ variant with uniform aggregation degrades further $(90.12\%/32.06\%)$. 
We also tested combining the projected Mahalanobis and residual components using an additive rule after score normalization to account for different magnitudes, but this performs slightly worse ($93.24\%/24.21\%$). 
We therefore adopt multiplicative fusion, which avoids the additional ID statistics required for normalization while yielding the best overall performance.
Beyond OOD detection performance, PRISM's PCA compression yields a compact representation (\eg, $d=512$), thereby reducing the complexity of covariance estimation compared to operating on the full stacked feature dimension. 
Finally, PRISM avoids the overhead of layer-wise modeling: per-layer Mahalanobis approaches must estimate and evaluate class-conditional statistics separately for each layer (\eg, for ImageNet-1k, 1000 class means per layer, plus repeated distance computations), whereas PRISM fits a single set of class-conditional statistics in the fused space after one PCA projection, making it simpler and typically more efficient.

\myparagraph{Summary.}
Taken together, the detailed tables and rank-based analysis reinforce the main practical conclusion: there is no universally best post-hoc OOD detector across domains and OOD types, but PRISM provides a robust ID-only default by integrating complementary information across representation depth, thereby reducing sensitivity to benchmark-specific depth preferences and improving consistency across natural, medical, and industrial settings.
Additionally, this comprehensive evaluation demonstrates the importance of a diverse evaluation framework (\eg, application domains, OOD categories) for validating and comparing OOD detection methods.

\input{Supplementary_Material/tables/openmibood_tasks_groupfirst}

%% file: Supplementary_Material/tables/benchmark_config.tex
\begin{table}[tb]
\centering
\footnotesize
\setlength{\extrarowheight}{1.3pt}
\setlength{\tabcolsep}{4.0pt}
\caption{
Detailed benchmark configuration. 
For each benchmark and in-distribution (ID) dataset, we list the corresponding near-, far-, and extreme-OOD evaluation sets.
}
\label{tab:benchmark_configuation}
\resizebox{\linewidth}{!}{%
\begin{tabular}{l c c c c}
\toprule
\textbf{Benchmark} & \textbf{ID Dataset} & \textbf{Near-OOD} & \textbf{Far-OOD} & \textbf{Extreme-OOD} \\
\midrule
\multirow{6}{*}{OpenOOD \cite{zhang2024openood}} 
&  \makecell[l]{CIFAR-10 \cite{krizhevsky09cifar}}    & \makecell[l]{CIFAR-100 \cite{krizhevsky09cifar},\\ Tiny ImageNet \cite{Torralba2008_tinyimagenet}} & \makecell[l]{MNIST \cite{lecun-mnisthandwrittendigit-2010}, SVHN \cite{Netzer11_svhn}, \\ Textures \cite{cimpoi14dtd}, Places365 \cite{zhou2017places}} & \makecell[l]{MIDOG \cite{aubreville2023_midog}, ICONIC-444 \cite{krumpl2026_iconic444}} \\ \cmidrule(lr){2-5}
&  \makecell[l]{CIFAR-100 \cite{krizhevsky09cifar}}   & \makecell[l]{CIFAR-10 \cite{krizhevsky09cifar},\\ Tiny ImageNet \cite{Torralba2008_tinyimagenet}} & \makecell[l]{MNIST \cite{lecun-mnisthandwrittendigit-2010}, SVHN \cite{Netzer11_svhn}, \\ Textures \cite{cimpoi14dtd}, Places365 \cite{zhou2017places}} & \makecell[l]{MIDOG \cite{aubreville2023_midog}, ICONIC-444 \cite{krumpl2026_iconic444}} \\  \cmidrule(lr){2-5}
&  \makecell[l]{ImageNet-1k \cite{deng2009_imagenet}} & \makecell[l]{ SSB-Hard \cite{vaze2022openset}, NINCO \cite{bitterwolf2023_ninco} } & \makecell[l]{ iNaturalist \cite{Horn_2018_inat}, Textures \cite{cimpoi14dtd}, \\ OpenImage-O \cite{wang2022_ood_vim}} & \makecell[l]{MNIST \cite{lecun-mnisthandwrittendigit-2010}, FMNIST \cite{xiao2017_fmnist}} \\ 
\midrule
\multirow{6}{*}{OpenMIBOOD \cite{gutbrod2025_openmibood}}
&  \makecell[l]{MIDOG \cite{aubreville2023_midog}}   & \makecell[l]{other MIDOG domain shifts \\ not in ID} & \makecell[l]{CCAgT \cite{amorim2020_ccagt_1, atkinson_amorim_ccagt_2}, FNAC 2019 \cite{saikia2019_fnac}} & \makecell[l]{PhaKIR \cite{RUECKERT2026_phakir}, ImageNet \cite{deng2009_imagenet},\\ iNaturalist \cite{Horn_2018_inat}, Textures \cite{cimpoi14dtd}} \\  \cmidrule(lr){2-5}
&  \makecell[l]{PhaKIR \cite{RUECKERT2026_phakir}}  & \makecell[l]{Cholec80 \cite{twinanda2016endonet_cholec80}, EndoSeg15 \cite{bodenstedt2018_endoseg2015},\\ EndoSeg18 \cite{allan2020_endoseg2018}} & \makecell[l]{Kvasir \cite{jha2020_kvasir}, CATARACTS \cite{ALHAJJ201924_cataracts}} & \makecell[l]{MIDOG \cite{aubreville2023_midog}, ImageNet \cite{deng2009_imagenet},\\ iNaturalist \cite{Horn_2018_inat}, Textures \cite{cimpoi14dtd}} \\  \cmidrule(lr){2-5}
&  \makecell[l]{OASIS-3 \cite{lamontagne2019_oasis}} & \makecell[l]{ATLAS \cite{liew2022_atlas}, BraTS \cite{baid2021_brats1, bakas2017_brats2, menze2014_brats3},\\ OASIS-CT \cite{lamontagne2019_oasis}} & \makecell[l]{MDS-Heart \cite{antonelli2022_mdsh1, simpson2019_mdsh2, tobon2015_mdsh3},\\ CHAOS \cite{CHAOSdata2019, KAVUR2021_chaos2}} & \makecell[l]{MIDOG \cite{aubreville2023_midog}, MNIST \cite{lecun-mnisthandwrittendigit-2010},\\ FMNIST \cite{xiao2017_fmnist}, Textures \cite{cimpoi14dtd}} \\
\midrule
\multirow{8}{*}{ICONIC-444 \cite{krumpl2026_iconic444}}
& \makecell[l]{Almond}    & \makecell[l]{ICONIC-444 Food (w/o ID)} & \makecell[l]{ICONIC-444 Non-Food} & \makecell[l]{ImageNet \cite{deng2009_imagenet}, iNaturalist \cite{Horn_2018_inat},\\ Places365 \cite{zhou2017places}, Textures \cite{cimpoi14dtd}} \\  \cmidrule(lr){2-5}
& \makecell[l]{Wheat}     & \makecell[l]{ICONIC-444 Food (w/o ID)} & \makecell[l]{ICONIC-444 Non-Food} & \makecell[l]{ImageNet \cite{deng2009_imagenet}, iNaturalist \cite{Horn_2018_inat},\\ Places365 \cite{zhou2017places}, Textures \cite{cimpoi14dtd}} \\  \cmidrule(lr){2-5}
& \makecell[l]{Kernels}   & \makecell[l]{ICONIC-444 Food (w/o ID)} & \makecell[l]{ICONIC-444 Non-Food} & \makecell[l]{ImageNet \cite{deng2009_imagenet}, iNaturalist \cite{Horn_2018_inat},\\ Places365 \cite{zhou2017places}, Textures \cite{cimpoi14dtd}} \\  \cmidrule(lr){2-5}
& \makecell[l]{Food-grade} & \makecell[l]{ICONIC-444 Food (w/o ID)} & \makecell[l]{ICONIC-444 Non-Food} & \makecell[l]{ImageNet \cite{deng2009_imagenet}, iNaturalist \cite{Horn_2018_inat},\\ Places365 \cite{zhou2017places}, Textures \cite{cimpoi14dtd}} \\
\bottomrule
\end{tabular}
}
\end{table}

%% file: Supplementary_Material/tables/model_source.tex
\begin{table}[t]
\centering
\footnotesize
\setlength{\tabcolsep}{3pt}
\caption{Overview of architectures and their corresponding official checkpoint sources or training protocol across all benchmarks.}
\label{tab:model_overview}
\resizebox{\linewidth}{!}{%
\begin{tabular}{l l l l}
\toprule
\textbf{Benchmark} & \textbf{ID Dataset / Task} & \textbf{Architecture} & \textbf{Checkpoint Source / Training} \\
\midrule
\multirow{6}{*}{OpenOOD \cite{zhang2024openood}}
& CIFAR-10 \cite{krizhevsky09cifar}     & ResNet-18 \cite{He2015_resnet} & Official OpenOOD checkpoint \\ \cmidrule(lr){2-4}
& CIFAR-100 \cite{krizhevsky09cifar}    & ResNet-18 \cite{He2015_resnet} & Official OpenOOD checkpoint \\ \cmidrule(lr){2-4}
& \multirow{3}{*}{ImageNet-1k \cite{deng2009_imagenet}}
                                       & ResNet-50 \cite{He2015_resnet}  & \texttt{timm: resnet50.tv2\_in1k} \\
&                                      & Swin-T \cite{liu2021Swin}        & \texttt{torchvision: Swin\_T\_Weights.IMAGENET1K\_V1} \\
&                                      & ViT-B/16 \cite{He_2022_MAE}      & \texttt{timm: vit\_base\_patch16\_224.mae\_in1k} \\
\midrule
\multirow{4}{*}{OpenMIBOOD \cite{gutbrod2025_openmibood}}
& MIDOG \cite{aubreville2023_midog}     & ResNet-50 \cite{He2015_resnet}  & Official OpenMIBOOD checkpoint \\ \cmidrule(lr){2-4}
& PhaKIR \cite{RUECKERT2026_phakir}     & ResNet-18 \cite{He2015_resnet}  & Official OpenMIBOOD checkpoint \\ \cmidrule(lr){2-4}
& OASIS-3 \cite{lamontagne2019_oasis}   & R(2+1)D \cite{du2018_r21d}       & Official OpenMIBOOD checkpoint \\
\midrule
\multirow{13}{*}{ICONIC-444 \cite{krumpl2026_iconic444}}
& \multirow{3}{*}{Almond}    & ResNet-18 \cite{He2015_resnet}     & Trained following ICONIC-444 protocol \\
&                            & CCT-7/7x2 \cite{hassani202_cct}     & Trained following ICONIC-444 protocol \\
&                            & ConvNeXt-S \cite{Liu_2022_CVPR}     & Trained following ICONIC-444 protocol \\ \cmidrule(lr){2-4}
& \multirow{3}{*}{Wheat}     & ResNet-18 \cite{He2015_resnet}     & Trained following ICONIC-444 protocol \\
&                            & CCT-7/7x2 \cite{hassani202_cct}     & Trained following ICONIC-444 protocol \\ 
&                            & ConvNeXt-S \cite{Liu_2022_CVPR}     & Trained following ICONIC-444 protocol \\ \cmidrule(lr){2-4}
& \multirow{3}{*}{Kernels}   & ResNet-18 \cite{He2015_resnet}     & Trained following ICONIC-444 protocol \\
&                            & CCT-7/7x2 \cite{hassani202_cct}     & Trained following ICONIC-444 protocol \\
&                            & ConvNeXt-S \cite{Liu_2022_CVPR}     & Trained following ICONIC-444 protocol \\ \cmidrule(lr){2-4}
& \multirow{3}{*}{Food-grade} & ResNet-18 \cite{He2015_resnet}     & Trained following ICONIC-444 protocol \\
&                            & CCT-7/7x2 \cite{hassani202_cct}     & Trained following ICONIC-444 protocol \\
&                            & ConvNeXt-S \cite{Liu_2022_CVPR}     & Trained following ICONIC-444 protocol \\
\bottomrule
\end{tabular}%
}
\end{table}

%% file: Supplementary_Material/tables/ood_methods_hyper_params.tex
\begin{table}[t]
\centering
\footnotesize
\setlength{\tabcolsep}{5pt}
\caption{Hyperparameter search space and calibration requirements for the considered OOD detection method.}
\label{tab:ood_hyperparameter}

\resizebox{\linewidth}{!}{%
\begin{tabular}{l cc l l l l}
\toprule
\multirow{2}{*}{\textbf{Method}} &
\multicolumn{2}{c}{\textbf{Cal. data}} &
\multicolumn{4}{c}{\textbf{Hyperparameter search space}} \\
\cmidrule(lr){2-3}\cmidrule(lr){4-7}
& ID & OOD &
Param 1 & Range & Param 2 & Range \\
\midrule

MSP~\cite{hendrycks2016_ood_msp}        &  &   & ---        & --- & --- & --- \\
MDS~\cite{lee2018_mahala}        & \cmark &   & ---        & --- & --- & --- \\
MDSEns~\cite{lee2018_mahala}     & \cmark & \cmarkred & noise      & [0, 0.0025, 0.0014, 0.005] & --- & --- \\
GRAM~\cite{sastry20a_gram}       & \cmark &  & ---        & --- & --- & --- \\
EBO~\cite{liu2020_ood_ebo}        &  &  & temperature& [0.1, 0.5, 1.0, 1.5, 2.0] & --- & --- \\
RMDS~\cite{Ren2021_rmds}       & \cmark &  & ---        & --- & --- & --- \\
ReAct~\cite{sun2021_ood_react}      & \cmark &  & percentile & [85, 90, 95, 99] & --- & --- \\
MLS~\cite{hendrycks2019_ood_mls}        &  &  & ---        & --- & --- & --- \\
ViM~\cite{wang2022_ood_vim}        & \cmark &  & dim        & [1, 16, 32, 64, 128, 256, 512] & --- & --- \\
KNN~\cite{sun2022knnood}        & \cmark &  & k          & [1, 2, 5, 10, 25, 50, 100, 200, 500, 750, 1000] & --- & --- \\
DICE~\cite{sun2022dice}       & \cmark &  & percentile & [60, 65, 70, 75, 80, 85, 90, 95] & --- & --- \\
FeatNorm~\cite{Yu2023_featurenorm}   & \cmark & \cmarkred & ---        & --- & --- & --- \\
ASH-b~\cite{djurisic2023ash}      &  &  & percentile & [65, 70, 75, 80, 85, 90, 95] & --- & --- \\
ASH-s~\cite{djurisic2023ash}      &  &  & percentile & [65, 70, 75, 80, 85, 90, 95] & --- & --- \\
Residual~\cite{wang2022_ood_vim}   & \cmark &  & dim        & [64, 128, 256, 512, 1024] & --- & --- \\
SHE~\cite{zhang2023_she}        & \cmark &  & metric     & [inner\_product, euclidean, cosine] & --- & --- \\
GEN~\cite{Liu2023_GEN}        &  &  & gamma      & [0.01, 0.1, 0.5, 1.0, 2.0, 5.0, 10.0] & M & [1, 2, 3, 4, 5, 6, 7, 50, 100, 200, 500, 1000] \\
ATS~\cite{Krumpl2024_ats}        & \cmark &  & ---        & --- & --- & --- \\
SCALE~\cite{xu2024scaling}      &  &  & percentile & [65, 70, 75, 80, 85, 90, 95] & --- & --- \\
fDBD~\cite{liu2024_fdbd}       & \cmark &  & normalized & [True, False] & --- & --- \\
MDS++~\cite{mueller2025_mahalanobispp}      & \cmark &  & ---        & --- & --- & --- \\
NCI~\cite{liu2025_nci}        & \cmark &  & alpha      & [0.0001, 0.001, 0.01, 0.1] & --- & --- \\
\rowcolor{gray!15}\textbf{PRISM (Ours)} & \cmark &  & --- & --- & --- & --- \\
\bottomrule
\end{tabular}%
}
\end{table}

%% file: Supplementary_Material/tables/cifar10_full_supp.tex
\begin{table}[tb]
\centering\scriptsize
\setlength{\extrarowheight}{1.5pt}
\setlength{\tabcolsep}{3.0pt}
\caption{
Performance of OOD detection methods for a ResNet-18 backbone trained on CIFAR-10.
}
\label{tab:cifar10_full_supp}
\resizebox{\textwidth}{!}{%
\begin{tabular}{lcccccccccccccccccc}
\toprule
\multirow{4}{*}{\textbf{Method}} & \multicolumn{4}{c}{\textbf{Near}} & \multicolumn{8}{c}{\textbf{Far}} & \multicolumn{4}{c}{\textbf{Extreme}} & \multicolumn{2}{c}{} \\
\cmidrule(lr){2-5}\cmidrule(lr){6-13}\cmidrule(lr){14-17}
 & \multicolumn{2}{c}{\textbf{CIFAR-100}} & \multicolumn{2}{c}{\textbf{Tiny ImageNet}} & \multicolumn{2}{c}{\textbf{MNIST}} & \multicolumn{2}{c}{\textbf{Places365}} & \multicolumn{2}{c}{\textbf{SVHN}} & \multicolumn{2}{c}{\textbf{Textures}} & \multicolumn{2}{c}{\textbf{ICONIC-444}} & \multicolumn{2}{c}{\textbf{MIDOG}} & \multicolumn{2}{c}{\textbf{Average}} \\
 & AUROC & FPR95 & AUROC & FPR95 & AUROC & FPR95 & AUROC & FPR95 & AUROC & FPR95 & AUROC & FPR95 & AUROC & FPR95 & AUROC & FPR95 & AUROC & FPR95 \\
 & $\uparrow$ & $\downarrow$ & $\uparrow$ & $\downarrow$ & $\uparrow$ & $\downarrow$ & $\uparrow$ & $\downarrow$ & $\uparrow$ & $\downarrow$ & $\uparrow$ & $\downarrow$ & $\uparrow$ & $\downarrow$ & $\uparrow$ & $\downarrow$ & $\uparrow$ & $\downarrow$ \\
\midrule
MSP~\cite{hendrycks2016_ood_msp} & $\phantom{0}87.16$ & $\phantom{0}60.60$ & $\phantom{0}88.85$ & $\phantom{0}56.85$ & $\phantom{0}92.53$ & $\phantom{0}47.03$ & $\phantom{0}89.03$ & $\phantom{0}56.15$ & $\phantom{0}91.31$ & $\phantom{0}53.33$ & $\phantom{0}89.94$ & $\phantom{0}54.96$ & $\phantom{0}81.41$ & $\phantom{0}61.25$ & $\phantom{0}89.02$ & $\phantom{0}61.68$ & $\phantom{0}87.97$ & $\phantom{0}57.69$ \\
MDS~\cite{lee2018_mahala} & $\phantom{0}83.62$ & $\phantom{0}69.54$ & $\phantom{0}84.85$ & $\phantom{0}69.34$ & $\phantom{0}89.94$ & $\phantom{0}63.84$ & $\phantom{0}85.01$ & $\phantom{0}68.45$ & $\phantom{0}91.37$ & $\phantom{0}55.65$ & $\phantom{0}92.68$ & $\phantom{0}40.55$ & $\phantom{0}95.33$ & $\phantom{0}29.45$ & $\phantom{0}88.68$ & $\phantom{0}65.56$ & $\phantom{0}88.66$ & $\phantom{0}58.02$ \\
MDSEns~\cite{lee2018_mahala} & $\phantom{0}84.75$ & $\phantom{0}67.18$ & $\phantom{0}86.19$ & $\phantom{0}66.71$ & $\phantom{0}93.94$ & $\phantom{0}43.02$ & $\phantom{0}86.15$ & $\phantom{0}67.42$ & $\phantom{0}91.30$ & $\phantom{0}52.09$ & $\phantom{0}91.69$ & $\phantom{0}40.01$ & $\textbf{100.00}$ & $\phantom{0}\phantom{0}\textbf{0.00}$ & $\phantom{0}98.28$ & $\phantom{0}10.08$ & $\phantom{0}91.79$ & $\phantom{0}40.87$ \\
GRAM~\cite{sastry20a_gram} & $\phantom{0}82.67$ & $\phantom{0}55.88$ & $\phantom{0}84.71$ & $\phantom{0}52.17$ & $\phantom{0}96.27$ & $\phantom{0}26.99$ & $\phantom{0}86.02$ & $\phantom{0}50.06$ & $\phantom{0}\underline{98.21}$ & $\phantom{0}\phantom{0}\underline{8.04}$ & $\phantom{0}94.65$ & $\phantom{0}\underline{23.26}$ & $\phantom{0}\underline{99.73}$ & $\phantom{0}\phantom{0}\textbf{0.00}$ & $\phantom{0}96.00$ & $\phantom{0}23.56$ & $\phantom{0}91.78$ & $\phantom{0}30.96$ \\
EBO~\cite{liu2020_ood_ebo} & $\phantom{0}86.19$ & $\phantom{0}51.19$ & $\phantom{0}88.70$ & $\phantom{0}\textbf{44.42}$ & $\phantom{0}94.26$ & $\phantom{0}27.10$ & $\phantom{0}89.51$ & $\phantom{0}\textbf{41.72}$ & $\phantom{0}91.28$ & $\phantom{0}40.86$ & $\phantom{0}89.40$ & $\phantom{0}43.58$ & $\phantom{0}78.39$ & $\phantom{0}61.25$ & $\phantom{0}89.47$ & $\phantom{0}48.08$ & $\phantom{0}87.49$ & $\phantom{0}46.93$ \\
RMDS~\cite{Ren2021_rmds} & $\phantom{0}88.80$ & $\phantom{0}55.17$ & $\phantom{0}90.74$ & $\phantom{0}50.04$ & $\phantom{0}93.20$ & $\phantom{0}43.56$ & $\phantom{0}91.54$ & $\phantom{0}45.46$ & $\phantom{0}91.63$ & $\phantom{0}54.03$ & $\phantom{0}92.16$ & $\phantom{0}47.34$ & $\phantom{0}88.02$ & $\phantom{0}64.05$ & $\phantom{0}90.33$ & $\phantom{0}60.43$ & $\phantom{0}90.36$ & $\phantom{0}54.15$ \\
ReAct~\cite{sun2021_ood_react} & $\phantom{0}85.86$ & $\phantom{0}52.29$ & $\phantom{0}88.25$ & $\phantom{0}46.51$ & $\phantom{0}92.96$ & $\phantom{0}31.58$ & $\phantom{0}90.32$ & $\phantom{0}\underline{42.35}$ & $\phantom{0}88.90$ & $\phantom{0}46.98$ & $\phantom{0}89.31$ & $\phantom{0}45.34$ & $\phantom{0}78.57$ & $\phantom{0}64.97$ & $\phantom{0}86.91$ & $\phantom{0}57.44$ & $\phantom{0}86.72$ & $\phantom{0}50.72$ \\
MLS~\cite{hendrycks2019_ood_mls} & $\phantom{0}86.19$ & $\phantom{0}51.67$ & $\phantom{0}88.63$ & $\phantom{0}45.52$ & $\phantom{0}93.99$ & $\phantom{0}29.16$ & $\phantom{0}89.36$ & $\phantom{0}43.03$ & $\phantom{0}91.21$ & $\phantom{0}41.72$ & $\phantom{0}89.38$ & $\phantom{0}44.13$ & $\phantom{0}78.50$ & $\phantom{0}60.52$ & $\phantom{0}89.31$ & $\phantom{0}49.64$ & $\phantom{0}87.43$ & $\phantom{0}47.73$ \\
ViM~\cite{wang2022_ood_vim} & $\phantom{0}87.91$ & $\phantom{0}55.30$ & $\phantom{0}89.70$ & $\phantom{0}51.47$ & $\phantom{0}94.55$ & $\phantom{0}34.24$ & $\phantom{0}89.91$ & $\phantom{0}50.20$ & $\phantom{0}94.16$ & $\phantom{0}35.90$ & $\phantom{0}\underline{95.13}$ & $\phantom{0}27.53$ & $\phantom{0}93.22$ & $\phantom{0}30.00$ & $\phantom{0}92.29$ & $\phantom{0}47.05$ & $\phantom{0}91.67$ & $\phantom{0}42.96$ \\
KNN~\cite{sun2022knnood} & $\phantom{0}\textbf{89.88}$ & $\phantom{0}\textbf{50.85}$ & $\phantom{0}\textbf{91.74}$ & $\phantom{0}45.46$ & $\phantom{0}94.55$ & $\phantom{0}34.84$ & $\phantom{0}\underline{92.07}$ & $\phantom{0}42.77$ & $\phantom{0}92.77$ & $\phantom{0}47.90$ & $\phantom{0}93.40$ & $\phantom{0}40.74$ & $\phantom{0}89.20$ & $\phantom{0}58.96$ & $\phantom{0}91.69$ & $\phantom{0}55.11$ & $\phantom{0}91.49$ & $\phantom{0}48.92$ \\
DICE~\cite{sun2022dice} & $\phantom{0}82.57$ & $\phantom{0}56.46$ & $\phantom{0}85.20$ & $\phantom{0}50.64$ & $\phantom{0}92.77$ & $\phantom{0}31.08$ & $\phantom{0}85.85$ & $\phantom{0}47.73$ & $\phantom{0}88.96$ & $\phantom{0}44.75$ & $\phantom{0}84.85$ & $\phantom{0}51.38$ & $\phantom{0}72.08$ & $\phantom{0}65.47$ & $\phantom{0}85.95$ & $\phantom{0}54.18$ & $\phantom{0}83.67$ & $\phantom{0}52.37$ \\
FeatNorm~\cite{Yu2023_featurenorm} & $\phantom{0}83.37$ & $\phantom{0}67.94$ & $\phantom{0}87.61$ & $\phantom{0}58.36$ & $\phantom{0}\textbf{99.43}$ & $\phantom{0}\phantom{0}\textbf{2.08}$ & $\phantom{0}89.14$ & $\phantom{0}54.11$ & $\phantom{0}\textbf{98.81}$ & $\phantom{0}\phantom{0}\textbf{6.42}$ & $\phantom{0}94.98$ & $\phantom{0}24.50$ & $\phantom{0}99.45$ & $\phantom{0}\phantom{0}\underline{2.51}$ & $\phantom{0}\underline{98.34}$ & $\phantom{0}\phantom{0}\underline{9.37}$ & $\phantom{0}\underline{93.32}$ & $\phantom{0}\underline{30.29}$ \\
ASH-b~\cite{djurisic2023ash} & $\phantom{0}73.71$ & $\phantom{0}63.60$ & $\phantom{0}76.12$ & $\phantom{0}60.05$ & $\phantom{0}82.52$ & $\phantom{0}49.00$ & $\phantom{0}79.99$ & $\phantom{0}54.95$ & $\phantom{0}72.51$ & $\phantom{0}66.59$ & $\phantom{0}77.06$ & $\phantom{0}59.55$ & $\phantom{0}64.24$ & $\phantom{0}72.38$ & $\phantom{0}67.83$ & $\phantom{0}76.30$ & $\phantom{0}72.99$ & $\phantom{0}64.56$ \\
ASH-s~\cite{djurisic2023ash} & $\phantom{0}78.68$ & $\phantom{0}55.87$ & $\phantom{0}81.20$ & $\phantom{0}51.63$ & $\phantom{0}88.37$ & $\phantom{0}38.23$ & $\phantom{0}84.26$ & $\phantom{0}46.55$ & $\phantom{0}80.16$ & $\phantom{0}54.51$ & $\phantom{0}81.90$ & $\phantom{0}50.12$ & $\phantom{0}72.44$ & $\phantom{0}61.54$ & $\phantom{0}77.19$ & $\phantom{0}61.22$ & $\phantom{0}79.48$ & $\phantom{0}54.16$ \\
Residual~\cite{wang2022_ood_vim} & $\phantom{0}80.44$ & $\phantom{0}73.83$ & $\phantom{0}80.96$ & $\phantom{0}74.29$ & $\phantom{0}87.80$ & $\phantom{0}67.02$ & $\phantom{0}81.11$ & $\phantom{0}74.14$ & $\phantom{0}89.65$ & $\phantom{0}60.03$ & $\phantom{0}90.76$ & $\phantom{0}45.70$ & $\phantom{0}95.46$ & $\phantom{0}29.68$ & $\phantom{0}86.93$ & $\phantom{0}64.86$ & $\phantom{0}86.41$ & $\phantom{0}61.02$ \\
SHE~\cite{zhang2023_she} & $\phantom{0}88.66$ & $\phantom{0}56.87$ & $\phantom{0}90.32$ & $\phantom{0}52.53$ & $\phantom{0}92.76$ & $\phantom{0}45.87$ & $\phantom{0}90.60$ & $\phantom{0}51.09$ & $\phantom{0}91.30$ & $\phantom{0}54.10$ & $\phantom{0}91.93$ & $\phantom{0}48.81$ & $\phantom{0}87.62$ & $\phantom{0}60.56$ & $\phantom{0}90.14$ & $\phantom{0}60.45$ & $\phantom{0}90.00$ & $\phantom{0}55.06$ \\
GEN~\cite{Liu2023_GEN} & $\phantom{0}87.87$ & $\phantom{0}51.88$ & $\phantom{0}89.88$ & $\phantom{0}46.35$ & $\phantom{0}94.16$ & $\phantom{0}31.42$ & $\phantom{0}90.25$ & $\phantom{0}44.75$ & $\phantom{0}92.22$ & $\phantom{0}42.56$ & $\phantom{0}90.94$ & $\phantom{0}43.85$ & $\phantom{0}81.34$ & $\phantom{0}60.02$ & $\phantom{0}90.18$ & $\phantom{0}51.97$ & $\phantom{0}88.84$ & $\phantom{0}48.59$ \\
ATS~\cite{Krumpl2024_ats} & $\phantom{0}80.07$ & $\phantom{0}62.42$ & $\phantom{0}82.13$ & $\phantom{0}61.13$ & $\phantom{0}\underline{99.02}$ & $\phantom{0}\phantom{0}\underline{4.34}$ & $\phantom{0}80.67$ & $\phantom{0}62.17$ & $\phantom{0}95.29$ & $\phantom{0}21.37$ & $\phantom{0}92.65$ & $\phantom{0}30.34$ & $\textbf{100.00}$ & $\phantom{0}\phantom{0}\textbf{0.00}$ & $\phantom{0}97.90$ & $\phantom{0}11.08$ & $\phantom{0}90.65$ & $\phantom{0}32.29$ \\
SCALE~\cite{xu2024scaling} & $\phantom{0}80.91$ & $\phantom{0}54.93$ & $\phantom{0}83.55$ & $\phantom{0}49.93$ & $\phantom{0}90.18$ & $\phantom{0}36.66$ & $\phantom{0}86.53$ & $\phantom{0}44.02$ & $\phantom{0}83.56$ & $\phantom{0}52.63$ & $\phantom{0}83.60$ & $\phantom{0}49.48$ & $\phantom{0}74.30$ & $\phantom{0}62.15$ & $\phantom{0}80.77$ & $\phantom{0}59.34$ & $\phantom{0}81.91$ & $\phantom{0}52.96$ \\
fDBD~\cite{liu2024_fdbd} & $\phantom{0}\underline{89.57}$ & $\phantom{0}\underline{51.03}$ & $\phantom{0}\underline{91.60}$ & $\phantom{0}\underline{45.11}$ & $\phantom{0}94.60$ & $\phantom{0}32.51$ & $\phantom{0}\textbf{92.08}$ & $\phantom{0}42.95$ & $\phantom{0}92.67$ & $\phantom{0}44.85$ & $\phantom{0}93.15$ & $\phantom{0}40.69$ & $\phantom{0}89.78$ & $\phantom{0}55.10$ & $\phantom{0}92.62$ & $\phantom{0}48.10$ & $\phantom{0}91.64$ & $\phantom{0}46.64$ \\
MDS++~\cite{mueller2025_mahalanobispp} & $\phantom{0}87.28$ & $\phantom{0}55.83$ & $\phantom{0}88.95$ & $\phantom{0}52.56$ & $\phantom{0}95.41$ & $\phantom{0}27.55$ & $\phantom{0}89.22$ & $\phantom{0}50.20$ & $\phantom{0}94.53$ & $\phantom{0}32.74$ & $\phantom{0}94.69$ & $\phantom{0}29.06$ & $\phantom{0}91.99$ & $\phantom{0}41.26$ & $\phantom{0}91.34$ & $\phantom{0}51.65$ & $\phantom{0}91.08$ & $\phantom{0}45.18$ \\
NCI~\cite{liu2025_nci} & $\phantom{0}87.83$ & $\phantom{0}54.88$ & $\phantom{0}89.49$ & $\phantom{0}50.34$ & $\phantom{0}91.93$ & $\phantom{0}41.49$ & $\phantom{0}90.34$ & $\phantom{0}49.41$ & $\phantom{0}90.58$ & $\phantom{0}51.64$ & $\phantom{0}91.96$ & $\phantom{0}45.17$ & $\phantom{0}85.71$ & $\phantom{0}58.57$ & $\phantom{0}89.12$ & $\phantom{0}55.40$ & $\phantom{0}89.09$ & $\phantom{0}52.17$ \\
\midrule
\rowcolor{gray!15}\textbf{PRISM (Ours)}  & $\phantom{0}88.38$ & $\phantom{0}53.91$ & $\phantom{0}89.94$ & $\phantom{0}50.73$ & $\phantom{0}97.87$ & $\phantom{0}12.54$ & $\phantom{0}90.11$ & $\phantom{0}48.93$ & $\phantom{0}95.79$ & $\phantom{0}25.72$ & $\phantom{0}\textbf{96.73}$ & $\phantom{0}\textbf{18.60}$ & $\textbf{100.00}$ & $\phantom{0}\phantom{0}\textbf{0.00}$ & $\phantom{0}\textbf{98.43}$ & $\phantom{0}\phantom{0}\textbf{9.10}$ & $\phantom{0}\textbf{94.50}$ & $\phantom{0}\textbf{27.77}$ \\

\bottomrule
\end{tabular}
}
\end{table}

%% file: Supplementary_Material/tables/cifar100_full_supp.tex
\begin{table}[tb]
\centering\scriptsize
\setlength{\extrarowheight}{1.5pt}
\setlength{\tabcolsep}{3.0pt}
\caption{
Performance of OOD detection methods for a ResNet-18 backbone trained on CIFAR-100.
}
\label{tab:cifar100_full_supp}
\resizebox{\textwidth}{!}{%
\begin{tabular}{lcccccccccccccccccc}
\toprule
\multirow{4}{*}{\textbf{Method}} & \multicolumn{4}{c}{\textbf{Near}} & \multicolumn{8}{c}{\textbf{Far}} & \multicolumn{4}{c}{\textbf{Extreme}} & \multicolumn{2}{c}{} \\
\cmidrule(lr){2-5}\cmidrule(lr){6-13}\cmidrule(lr){14-17}
 & \multicolumn{2}{c}{\textbf{CIFAR-10}} & \multicolumn{2}{c}{\textbf{Tiny ImageNet}} & \multicolumn{2}{c}{\textbf{MNIST}} & \multicolumn{2}{c}{\textbf{Places365}} & \multicolumn{2}{c}{\textbf{SVHN}} & \multicolumn{2}{c}{\textbf{Textures}} & \multicolumn{2}{c}{\textbf{ICONIC-444}} & \multicolumn{2}{c}{\textbf{MIDOG}} & \multicolumn{2}{c}{\textbf{Average}} \\
 & AUROC & FPR95 & AUROC & FPR95 & AUROC & FPR95 & AUROC & FPR95 & AUROC & FPR95 & AUROC & FPR95 & AUROC & FPR95 & AUROC & FPR95 & AUROC & FPR95 \\
 & $\uparrow$ & $\downarrow$ & $\uparrow$ & $\downarrow$ & $\uparrow$ & $\downarrow$ & $\uparrow$ & $\downarrow$ & $\uparrow$ & $\downarrow$ & $\uparrow$ & $\downarrow$ & $\uparrow$ & $\downarrow$ & $\uparrow$ & $\downarrow$ & $\uparrow$ & $\downarrow$ \\
\midrule
MSP~\cite{hendrycks2016_ood_msp} & $\phantom{0}78.49$ & $\phantom{0}79.85$ & $\phantom{0}82.12$ & $\phantom{0}74.53$ & $\phantom{0}75.64$ & $\phantom{0}87.65$ & $\phantom{0}79.27$ & $\phantom{0}79.31$ & $\phantom{0}79.02$ & $\phantom{0}78.67$ & $\phantom{0}77.57$ & $\phantom{0}80.73$ & $\phantom{0}83.35$ & $\phantom{0}74.86$ & $\phantom{0}71.59$ & $\phantom{0}90.70$ & $\phantom{0}78.55$ & $\phantom{0}80.52$ \\
MDS~\cite{lee2018_mahala} & $\phantom{0}56.00$ & $\phantom{0}95.91$ & $\phantom{0}61.57$ & $\phantom{0}94.29$ & $\phantom{0}67.97$ & $\phantom{0}93.46$ & $\phantom{0}63.18$ & $\phantom{0}93.24$ & $\phantom{0}70.60$ & $\phantom{0}90.24$ & $\phantom{0}76.13$ & $\phantom{0}73.09$ & $\phantom{0}68.61$ & $\phantom{0}87.86$ & $\phantom{0}57.87$ & $\phantom{0}97.63$ & $\phantom{0}63.83$ & $\phantom{0}91.78$ \\
MDSEns~\cite{lee2018_mahala} & $\phantom{0}57.16$ & $\phantom{0}95.38$ & $\phantom{0}65.66$ & $\phantom{0}92.47$ & $\phantom{0}88.55$ & $\phantom{0}61.05$ & $\phantom{0}67.00$ & $\phantom{0}91.07$ & $\phantom{0}83.36$ & $\phantom{0}71.00$ & $\phantom{0}85.88$ & $\phantom{0}51.37$ & $\phantom{0}93.00$ & $\phantom{0}\phantom{0}7.36$ & $\phantom{0}\underline{91.24}$ & $\phantom{0}\underline{47.92}$ & $\phantom{0}78.24$ & $\phantom{0}63.40$ \\
GRAM~\cite{sastry20a_gram} & $\phantom{0}73.46$ & $\phantom{0}81.52$ & $\phantom{0}78.05$ & $\phantom{0}76.86$ & $\phantom{0}85.76$ & $\phantom{0}61.63$ & $\phantom{0}77.04$ & $\phantom{0}78.10$ & $\phantom{0}92.35$ & $\phantom{0}37.95$ & $\phantom{0}\underline{89.68}$ & $\phantom{0}\underline{47.46}$ & $\phantom{0}99.94$ & $\phantom{0}\phantom{0}\textbf{0.00}$ & $\phantom{0}86.16$ & $\phantom{0}64.74$ & $\phantom{0}\underline{85.00}$ & $\phantom{0}\underline{55.95}$ \\
EBO~\cite{liu2020_ood_ebo} & $\phantom{0}79.20$ & $\phantom{0}80.07$ & $\phantom{0}82.93$ & $\phantom{0}74.12$ & $\phantom{0}78.35$ & $\phantom{0}87.39$ & $\phantom{0}79.75$ & $\phantom{0}79.98$ & $\phantom{0}82.31$ & $\phantom{0}76.56$ & $\phantom{0}78.74$ & $\phantom{0}80.17$ & $\phantom{0}86.88$ & $\phantom{0}69.64$ & $\phantom{0}77.50$ & $\phantom{0}86.72$ & $\phantom{0}81.01$ & $\phantom{0}78.77$ \\
RMDS~\cite{Ren2021_rmds} & $\phantom{0}77.80$ & $\phantom{0}82.72$ & $\phantom{0}82.57$ & $\phantom{0}75.46$ & $\phantom{0}79.41$ & $\phantom{0}84.70$ & $\phantom{0}\textbf{83.38}$ & $\phantom{0}\textbf{67.17}$ & $\phantom{0}85.08$ & $\phantom{0}62.70$ & $\phantom{0}83.69$ & $\phantom{0}64.86$ & $\phantom{0}80.21$ & $\phantom{0}91.13$ & $\phantom{0}77.58$ & $\phantom{0}78.95$ & $\phantom{0}80.66$ & $\phantom{0}78.00$ \\
ReAct~\cite{sun2021_ood_react} & $\phantom{0}78.77$ & $\phantom{0}80.46$ & $\phantom{0}82.89$ & $\phantom{0}74.25$ & $\phantom{0}77.97$ & $\phantom{0}86.69$ & $\phantom{0}79.90$ & $\phantom{0}80.61$ & $\phantom{0}83.19$ & $\phantom{0}75.80$ & $\phantom{0}80.06$ & $\phantom{0}79.88$ & $\phantom{0}87.14$ & $\phantom{0}67.96$ & $\phantom{0}78.72$ & $\phantom{0}83.71$ & $\phantom{0}81.35$ & $\phantom{0}77.98$ \\
MLS~\cite{hendrycks2019_ood_mls} & $\phantom{0}79.21$ & $\phantom{0}80.01$ & $\phantom{0}82.93$ & $\phantom{0}74.02$ & $\phantom{0}78.31$ & $\phantom{0}87.35$ & $\phantom{0}79.77$ & $\phantom{0}79.95$ & $\phantom{0}82.20$ & $\phantom{0}77.03$ & $\phantom{0}78.73$ & $\phantom{0}80.23$ & $\phantom{0}86.70$ & $\phantom{0}70.98$ & $\phantom{0}77.35$ & $\phantom{0}87.41$ & $\phantom{0}80.95$ & $\phantom{0}79.12$ \\
ViM~\cite{wang2022_ood_vim} & $\phantom{0}74.42$ & $\phantom{0}87.40$ & $\phantom{0}79.69$ & $\phantom{0}81.79$ & $\phantom{0}78.47$ & $\phantom{0}83.85$ & $\phantom{0}79.16$ & $\phantom{0}81.19$ & $\phantom{0}82.25$ & $\phantom{0}77.33$ & $\phantom{0}84.08$ & $\phantom{0}64.09$ & $\phantom{0}80.72$ & $\phantom{0}81.69$ & $\phantom{0}73.48$ & $\phantom{0}94.04$ & $\phantom{0}78.38$ & $\phantom{0}83.02$ \\
KNN~\cite{sun2022knnood} & $\phantom{0}77.97$ & $\phantom{0}80.88$ & $\phantom{0}\underline{83.64}$ & $\phantom{0}73.83$ & $\phantom{0}80.98$ & $\phantom{0}83.34$ & $\phantom{0}80.19$ & $\phantom{0}78.31$ & $\phantom{0}83.78$ & $\phantom{0}70.15$ & $\phantom{0}80.97$ & $\phantom{0}74.57$ & $\phantom{0}87.66$ & $\phantom{0}70.16$ & $\phantom{0}72.99$ & $\phantom{0}89.70$ & $\phantom{0}80.87$ & $\phantom{0}77.96$ \\
DICE~\cite{sun2022dice} & $\phantom{0}77.86$ & $\phantom{0}81.36$ & $\phantom{0}80.72$ & $\phantom{0}79.76$ & $\phantom{0}79.59$ & $\phantom{0}81.30$ & $\phantom{0}77.91$ & $\phantom{0}83.10$ & $\phantom{0}84.95$ & $\phantom{0}69.24$ & $\phantom{0}77.66$ & $\phantom{0}79.44$ & $\phantom{0}87.85$ & $\phantom{0}58.32$ & $\phantom{0}85.86$ & $\phantom{0}63.26$ & $\phantom{0}82.06$ & $\phantom{0}73.21$ \\
FeatNorm~\cite{Yu2023_featurenorm} & $\phantom{0}45.61$ & $\phantom{0}98.07$ & $\phantom{0}49.98$ & $\phantom{0}95.65$ & $\phantom{0}\underline{95.57}$ & $\phantom{0}\underline{31.64}$ & $\phantom{0}53.69$ & $\phantom{0}94.83$ & $\phantom{0}\textbf{93.44}$ & $\phantom{0}\textbf{33.19}$ & $\phantom{0}81.03$ & $\phantom{0}57.58$ & $\phantom{0}93.96$ & $\phantom{0}31.80$ & $\phantom{0}89.97$ & $\phantom{0}61.99$ & $\phantom{0}73.56$ & $\phantom{0}66.02$ \\
ASH-b~\cite{djurisic2023ash} & $\phantom{0}76.48$ & $\phantom{0}79.92$ & $\phantom{0}79.92$ & $\phantom{0}75.88$ & $\phantom{0}76.68$ & $\phantom{0}81.00$ & $\phantom{0}78.77$ & $\phantom{0}79.23$ & $\phantom{0}85.72$ & $\phantom{0}65.63$ & $\phantom{0}80.98$ & $\phantom{0}73.49$ & $\phantom{0}88.88$ & $\phantom{0}58.07$ & $\phantom{0}80.10$ & $\phantom{0}81.96$ & $\phantom{0}81.08$ & $\phantom{0}74.25$ \\
ASH-s~\cite{djurisic2023ash} & $\phantom{0}79.18$ & $\phantom{0}\textbf{79.24}$ & $\phantom{0}82.49$ & $\phantom{0}74.55$ & $\phantom{0}80.14$ & $\phantom{0}81.67$ & $\phantom{0}80.49$ & $\phantom{0}78.05$ & $\phantom{0}85.62$ & $\phantom{0}66.17$ & $\phantom{0}81.38$ & $\phantom{0}74.13$ & $\phantom{0}90.14$ & $\phantom{0}57.60$ & $\phantom{0}79.26$ & $\phantom{0}83.80$ & $\phantom{0}82.48$ & $\phantom{0}74.20$ \\
Residual~\cite{wang2022_ood_vim} & $\phantom{0}44.58$ & $\phantom{0}97.13$ & $\phantom{0}48.03$ & $\phantom{0}96.24$ & $\phantom{0}59.21$ & $\phantom{0}94.95$ & $\phantom{0}49.42$ & $\phantom{0}95.73$ & $\phantom{0}57.54$ & $\phantom{0}93.38$ & $\phantom{0}66.23$ & $\phantom{0}78.07$ & $\phantom{0}59.00$ & $\phantom{0}89.63$ & $\phantom{0}44.77$ & $\phantom{0}98.26$ & $\phantom{0}52.10$ & $\phantom{0}93.72$ \\
SHE~\cite{zhang2023_she} & $\phantom{0}77.23$ & $\phantom{0}81.37$ & $\phantom{0}83.02$ & $\phantom{0}74.52$ & $\phantom{0}80.83$ & $\phantom{0}80.76$ & $\phantom{0}80.33$ & $\phantom{0}\underline{76.65}$ & $\phantom{0}84.71$ & $\phantom{0}66.19$ & $\phantom{0}84.10$ & $\phantom{0}67.17$ & $\phantom{0}88.48$ & $\phantom{0}68.39$ & $\phantom{0}74.68$ & $\phantom{0}88.09$ & $\phantom{0}81.40$ & $\phantom{0}76.29$ \\
GEN~\cite{Liu2023_GEN} & $\phantom{0}\textbf{79.29}$ & $\phantom{0}79.98$ & $\phantom{0}83.27$ & $\phantom{0}73.19$ & $\phantom{0}78.46$ & $\phantom{0}87.92$ & $\phantom{0}80.16$ & $\phantom{0}79.33$ & $\phantom{0}82.38$ & $\phantom{0}76.78$ & $\phantom{0}79.21$ & $\phantom{0}79.70$ & $\phantom{0}87.26$ & $\phantom{0}69.68$ & $\phantom{0}77.11$ & $\phantom{0}87.27$ & $\phantom{0}81.17$ & $\phantom{0}78.66$ \\
ATS~\cite{Krumpl2024_ats} & $\phantom{0}71.49$ & $\phantom{0}86.63$ & $\phantom{0}75.61$ & $\phantom{0}82.71$ & $\phantom{0}\textbf{95.62}$ & $\phantom{0}\textbf{27.40}$ & $\phantom{0}70.80$ & $\phantom{0}87.44$ & $\phantom{0}\underline{92.87}$ & $\phantom{0}\underline{36.83}$ & $\phantom{0}84.84$ & $\phantom{0}61.11$ & $\phantom{0}\underline{99.96}$ & $\phantom{0}\phantom{0}\underline{0.01}$ & $\phantom{0}77.13$ & $\phantom{0}73.85$ & $\phantom{0}82.71$ & $\phantom{0}58.27$ \\
SCALE~\cite{xu2024scaling} & $\phantom{0}\underline{79.24}$ & $\phantom{0}\underline{79.41}$ & $\phantom{0}82.74$ & $\phantom{0}74.12$ & $\phantom{0}79.69$ & $\phantom{0}83.26$ & $\phantom{0}80.50$ & $\phantom{0}77.92$ & $\phantom{0}84.92$ & $\phantom{0}68.42$ & $\phantom{0}80.80$ & $\phantom{0}75.44$ & $\phantom{0}89.22$ & $\phantom{0}61.43$ & $\phantom{0}78.93$ & $\phantom{0}84.28$ & $\phantom{0}82.18$ & $\phantom{0}75.29$ \\
fDBD~\cite{liu2024_fdbd} & $\phantom{0}78.39$ & $\phantom{0}80.34$ & $\phantom{0}\textbf{83.96}$ & $\phantom{0}\textbf{72.24}$ & $\phantom{0}78.76$ & $\phantom{0}87.99$ & $\phantom{0}79.84$ & $\phantom{0}79.32$ & $\phantom{0}80.80$ & $\phantom{0}80.36$ & $\phantom{0}81.27$ & $\phantom{0}78.04$ & $\phantom{0}86.14$ & $\phantom{0}77.99$ & $\phantom{0}72.81$ & $\phantom{0}93.99$ & $\phantom{0}80.27$ & $\phantom{0}81.24$ \\
MDS++~\cite{mueller2025_mahalanobispp} & $\phantom{0}71.70$ & $\phantom{0}86.55$ & $\phantom{0}76.54$ & $\phantom{0}83.72$ & $\phantom{0}82.39$ & $\phantom{0}76.32$ & $\phantom{0}76.65$ & $\phantom{0}81.45$ & $\phantom{0}87.03$ & $\phantom{0}59.78$ & $\phantom{0}87.07$ & $\phantom{0}53.92$ & $\phantom{0}91.80$ & $\phantom{0}45.06$ & $\phantom{0}76.55$ & $\phantom{0}84.20$ & $\phantom{0}80.53$ & $\phantom{0}72.54$ \\
NCI~\cite{liu2025_nci} & $\phantom{0}78.36$ & $\phantom{0}80.30$ & $\phantom{0}83.54$ & $\phantom{0}\underline{72.90}$ & $\phantom{0}79.56$ & $\phantom{0}85.51$ & $\phantom{0}\underline{80.85}$ & $\phantom{0}78.13$ & $\phantom{0}83.16$ & $\phantom{0}75.23$ & $\phantom{0}83.79$ & $\phantom{0}72.30$ & $\phantom{0}86.63$ & $\phantom{0}74.21$ & $\phantom{0}74.84$ & $\phantom{0}92.89$ & $\phantom{0}81.18$ & $\phantom{0}79.31$ \\
\midrule
\rowcolor{gray!15}\textbf{PRISM (Ours)} & $\phantom{0}70.59$ & $\phantom{0}89.05$ & $\phantom{0}77.65$ & $\phantom{0}82.26$ & $\phantom{0}92.42$ & $\phantom{0}45.95$ & $\phantom{0}77.21$ & $\phantom{0}81.22$ & $\phantom{0}91.19$ & $\phantom{0}41.48$ & $\phantom{0}\textbf{93.21}$ & $\phantom{0}\textbf{30.89}$ & $\textbf{100.00}$ & $\phantom{0}\phantom{0}\textbf{0.00}$ & $\phantom{0}\textbf{94.48}$ & $\phantom{0}\textbf{31.36}$ & $\phantom{0}\textbf{86.62}$ & $\phantom{0}\textbf{50.40}$ \\

\bottomrule
\end{tabular}
}
\end{table}

%% file: Supplementary_Material/tables/openood_full_supp_all_models_v2.tex
\begin{table}[tb]
\centering\scriptsize
\setlength{\extrarowheight}{1.5pt}
\setlength{\tabcolsep}{3pt}
\caption{
Performance of OOD detection methods on ImageNet-1k using three backbones: ResNet-50, ViT-B/16, and Swin-T.
Metrics are reported in $\%$ (higher is better for AUROC; lower is better for FPR95, indicated by $\uparrow/\downarrow$).
The \textbf{best} and \underline{second-best} results for each OOD dataset (\ie, each column) are shown in bold or underlined, respectively.
}
\label{tab:imagenet_full_supp_all_models}
\resizebox{\textwidth}{!}{%
\begin{tabular}{clcccccccccccccccc}
\toprule
 & \multirow{4}{*}{\textbf{Method}} & \multicolumn{4}{c}{\textbf{Near}} & \multicolumn{6}{c}{\textbf{Far}} & \multicolumn{4}{c}{\textbf{Extreme}} & \multicolumn{2}{c}{\textbf{}} \\
\cmidrule(lr){3-6}\cmidrule(lr){7-12}\cmidrule(lr){13-16}
 &  & \multicolumn{2}{c}{\textbf{SSB-Hard}} & \multicolumn{2}{c}{\textbf{NINCO}} & \multicolumn{2}{c}{\textbf{iNaturalist}} & \multicolumn{2}{c}{\textbf{Textures}} & \multicolumn{2}{c}{\textbf{OpenImage-O}} & \multicolumn{2}{c}{\textbf{MNIST}} & \multicolumn{2}{c}{\textbf{FashionMNIST}} & \multicolumn{2}{c}{\textbf{Average}} \\
 &  & AUROC & FPR95 & AUROC & FPR95 & AUROC & FPR95 & AUROC & FPR95 & AUROC & FPR95 & AUROC & FPR95 & AUROC & FPR95 & AUROC & FPR95 \\
 &  & $\uparrow$ & $\downarrow$ & $\uparrow$ & $\downarrow$ & $\uparrow$ & $\downarrow$ & $\uparrow$ & $\downarrow$ & $\uparrow$ & $\downarrow$ & $\uparrow$ & $\downarrow$ & $\uparrow$ & $\downarrow$ & $\uparrow$ & $\downarrow$ \\
\midrule
\multirow{23}{*}{\rotatebox{90}{\textbf{ResNet-50}}} & MSP~\cite{hendrycks2016_ood_msp} & $\phantom{0}68.32$ & $\phantom{0}85.50$ & $\phantom{0}76.73$ & $\phantom{0}76.15$ & $\phantom{0}84.80$ & $\phantom{0}59.68$ & $\phantom{0}77.34$ & $\phantom{0}70.50$ & $\phantom{0}82.29$ & $\phantom{0}67.07$ & $\phantom{0}78.39$ & $\phantom{0}78.01$ & $\phantom{0}81.22$ & $\phantom{0}62.81$ & $\phantom{0}77.94$ & $\phantom{0}72.33$ \\
 & MDS~\cite{lee2018_mahala} & $\phantom{0}70.44$ & $\phantom{0}81.32$ & $\phantom{0}82.58$ & $\phantom{0}70.13$ & $\phantom{0}\textbf{95.25}$ & $\phantom{0}\textbf{26.25}$ & $\phantom{0}93.57$ & $\phantom{0}26.16$ & $\phantom{0}\textbf{92.14}$ & $\phantom{0}\underline{43.12}$ & $\phantom{0}99.75$ & $\phantom{0}\phantom{0}0.03$ & $\phantom{0}98.90$ & $\phantom{0}\phantom{0}3.13$ & $\phantom{0}89.83$ & $\phantom{0}\underline{36.38}$ \\
\bottomrule
\end{tabular}
}
\end{table}

%% file: Supplementary_Material/tables/iconic444_taskavg_models.tex
\begin{table}[tb]
\centering\scriptsize
\setlength{\extrarowheight}{1.5pt}
\setlength{\tabcolsep}{2.5pt}
\caption{
Average OOD detection performance per method and OOD test set using three backbones: Resnet-18, CCT-7/7x2, and ConvNeXt-S.
Results are reported as the mean over the four tasks (Almond, Wheat, Kernels, and Food-grade).
Arrows ($\uparrow$/$\downarrow$) indicate whether higher or lower values are better.
Additionally, the \textbf{best} and \underline{second-best} results in each column are highlighted in bold and underlined, respectively. 
All values are reported as percentages.
}
\label{tab:iconic444_taskavg_models}
\resizebox{\textwidth}{!}{%
\begin{tabular}{clcccccccccccccccc}
\toprule
 & \multirow{4}{*}{\textbf{Method}} & \multicolumn{2}{c}{\textbf{Near}} & \multicolumn{2}{c}{\textbf{Far}} & \multicolumn{10}{c}{\textbf{Extreme}} & \multicolumn{2}{c}{} \\
\cmidrule(lr){3-4}\cmidrule(lr){5-6}\cmidrule(lr){7-16}
 &  & \multicolumn{2}{c}{\textbf{Mean}} & \multicolumn{2}{c}{\textbf{Mean}} & \multicolumn{2}{c}{\textbf{ImageNet-1k}} & \multicolumn{2}{c}{\textbf{iNaturalist}} & \multicolumn{2}{c}{\textbf{Textures}} & \multicolumn{2}{c}{\textbf{Places365}} & \multicolumn{2}{c}{\textbf{Mean}} & \multicolumn{2}{c}{\textbf{Average}} \\
 &  & AUROC & FPR95 & AUROC & FPR95 & AUROC & FPR95 & AUROC & FPR95 & AUROC & FPR95 & AUROC & FPR95 & AUROC & FPR95 & AUROC & FPR95 \\
 &  & $\uparrow$ & $\downarrow$ & $\uparrow$ & $\downarrow$ & $\uparrow$ & $\downarrow$ & $\uparrow$ & $\downarrow$ & $\uparrow$ & $\downarrow$ & $\uparrow$ & $\downarrow$ & $\uparrow$ & $\downarrow$ & $\uparrow$ & $\downarrow$ \\
\midrule
\multirow{23}{*}{\rotatebox{90}{\textbf{ResNet-18}}} & MSP~\cite{hendrycks2016_ood_msp} & $\phantom{0}72.86$ & $\phantom{0}72.12$ & $\phantom{0}74.85$ & $\phantom{0}64.22$ & $\phantom{0}76.06$ & $\phantom{0}55.12$ & $\phantom{0}64.79$ & $\phantom{0}62.83$ & $\phantom{0}68.46$ & $\phantom{0}64.08$ & $\phantom{0}76.59$ & $\phantom{0}53.52$ & $\phantom{0}71.48$ & $\phantom{0}58.89$ & $\phantom{0}73.06$ & $\phantom{0}65.07$ \\
 & MDS~\cite{lee2018_mahala} & $\phantom{0}89.14$ & $\phantom{0}32.15$ & $\phantom{0}97.03$ & $\phantom{0}14.25$ & $\phantom{0}99.73$ & $\phantom{0}\phantom{0}0.54$ & $\phantom{0}99.86$ & $\phantom{0}\phantom{0}0.15$ & $\phantom{0}99.72$ & $\phantom{0}\phantom{0}0.99$ & $\phantom{0}99.74$ & $\phantom{0}\phantom{0}0.45$ & $\phantom{0}99.76$ & $\phantom{0}\phantom{0}0.53$ & $\phantom{0}95.31$ & $\phantom{0}15.65$ \\
\bottomrule
\end{tabular}
}
\end{table}

%% file: Supplementary_Material/figures/mean_rank_ci95.tex
\begin{figure}[tb]
\centering
\includegraphics[width=0.7\linewidth]{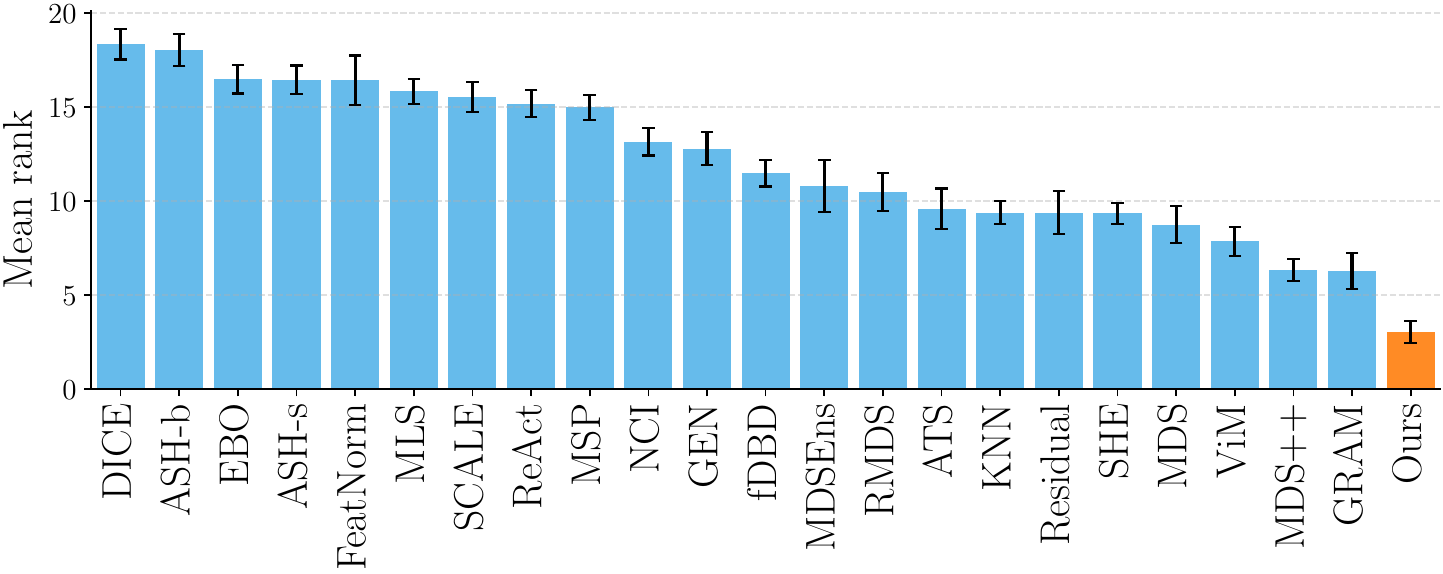}
\caption{
Mean-rank comparison (AUROC) across OOD datasets.
Average rank of OOD detection methods computed over OOD datasets.
Error bars show $95\%$ CIs; lower rank is better.
}
\label{fig:mean_rank_ci95}
\end{figure}

%% file: Supplementary_Material/figures/openood_imagenet_samples.tex
\begin{figure}[tb]
\centering
 \includegraphics[width=\linewidth]{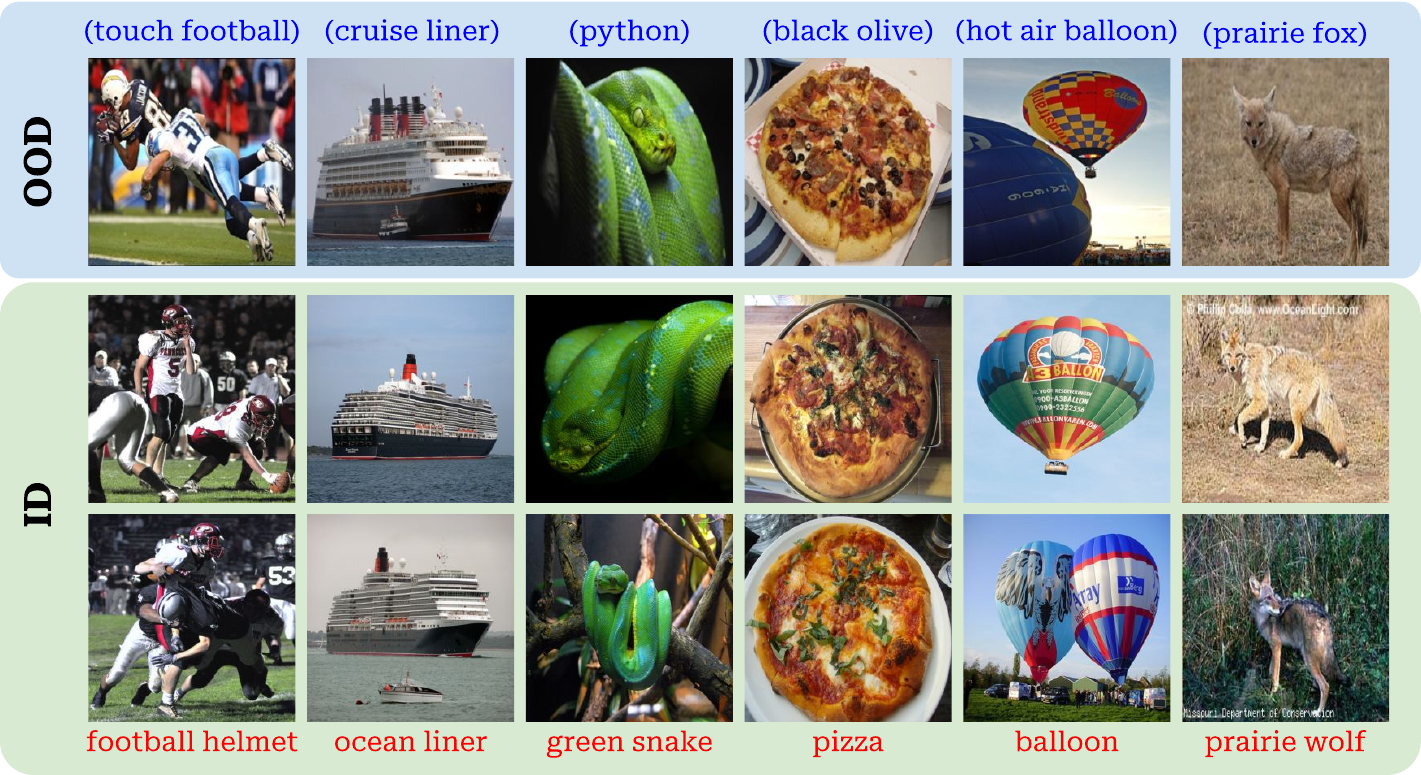}
\caption{
Qualitative analysis for OpenOOD (ImageNet-1k / SSB-hard).
For each column, the top row shows an OOD sample from the SSB-hard split, with its original OOD class label annotated in \textcolor{oodclassblue}{\textbf{blue}} (parentheses). 
The two rows below show two ID ImageNet examples from the ID class predicted by the Mahalanobis component of our detector (\ie, the class achieving the minimum Mahalanobis distance in PCA space), with the corresponding predicted ID class label annotated in \textcolor{idclassred}{\textbf{red}}.
}
\label{fig:openood_ssb_hard_samples}
\end{figure}

%% file: Supplementary_Material/figures/openood_textures_samples.tex
\begin{figure}[tb]
\centering
 \includegraphics[width=\linewidth]{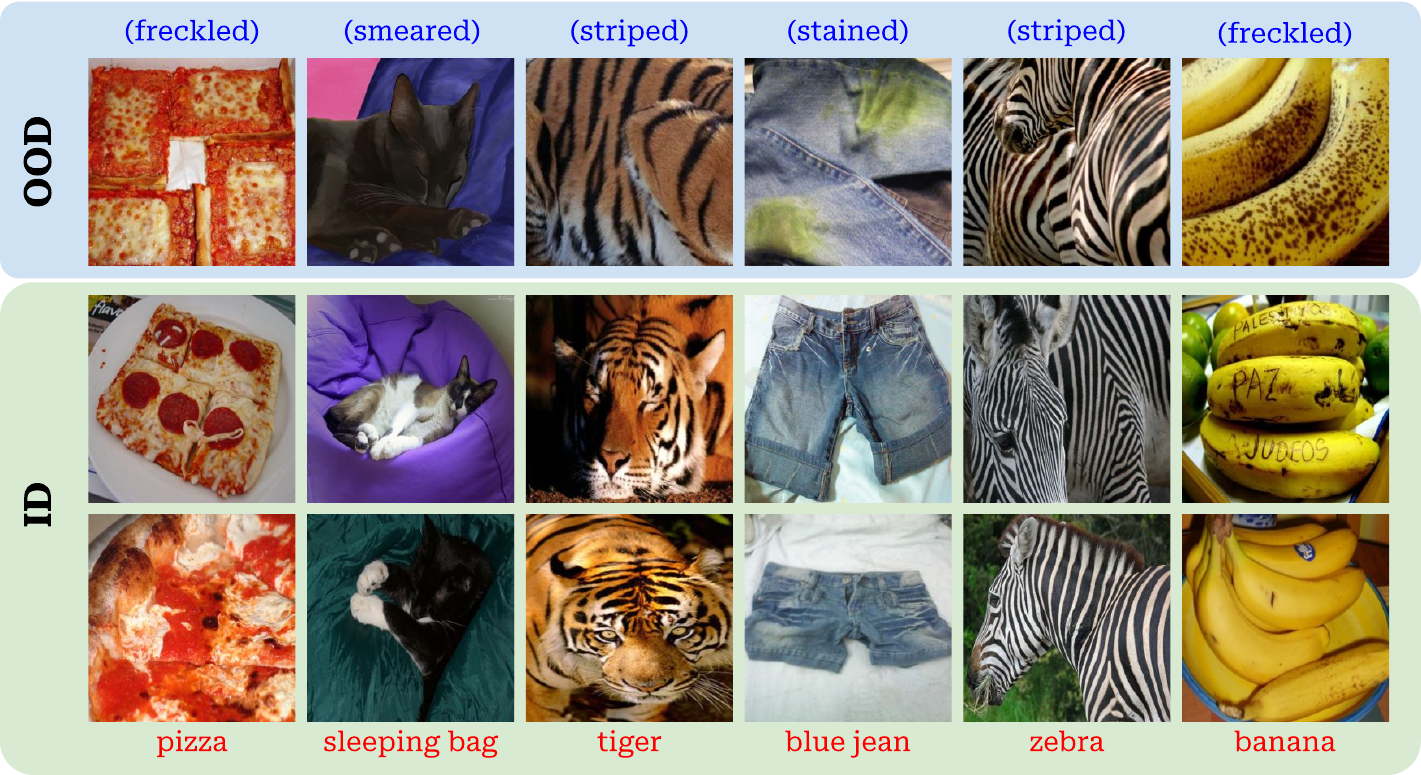}
\caption{
Qualitative analysis for OpenOOD (ImageNet-1k / Textures).
For each column, the top row shows an OOD sample from the Textures split, with its original OOD class label annotated in \textcolor{oodclassblue}{\textbf{blue}} (parentheses). 
The two rows below show two ID ImageNet examples from the ID class predicted by the Mahalanobis component of our detector (\ie, the class achieving the minimum Mahalanobis distance in PCA space), with the corresponding predicted ID class label annotated in \textcolor{idclassred}{\textbf{red}}.
}
\label{fig:openood_textures_samples}
\end{figure}

%% file: Supplementary_Material/figures/openood_ninco_samples.tex
\begin{figure}[tb]
\centering
 \includegraphics[width=\linewidth]{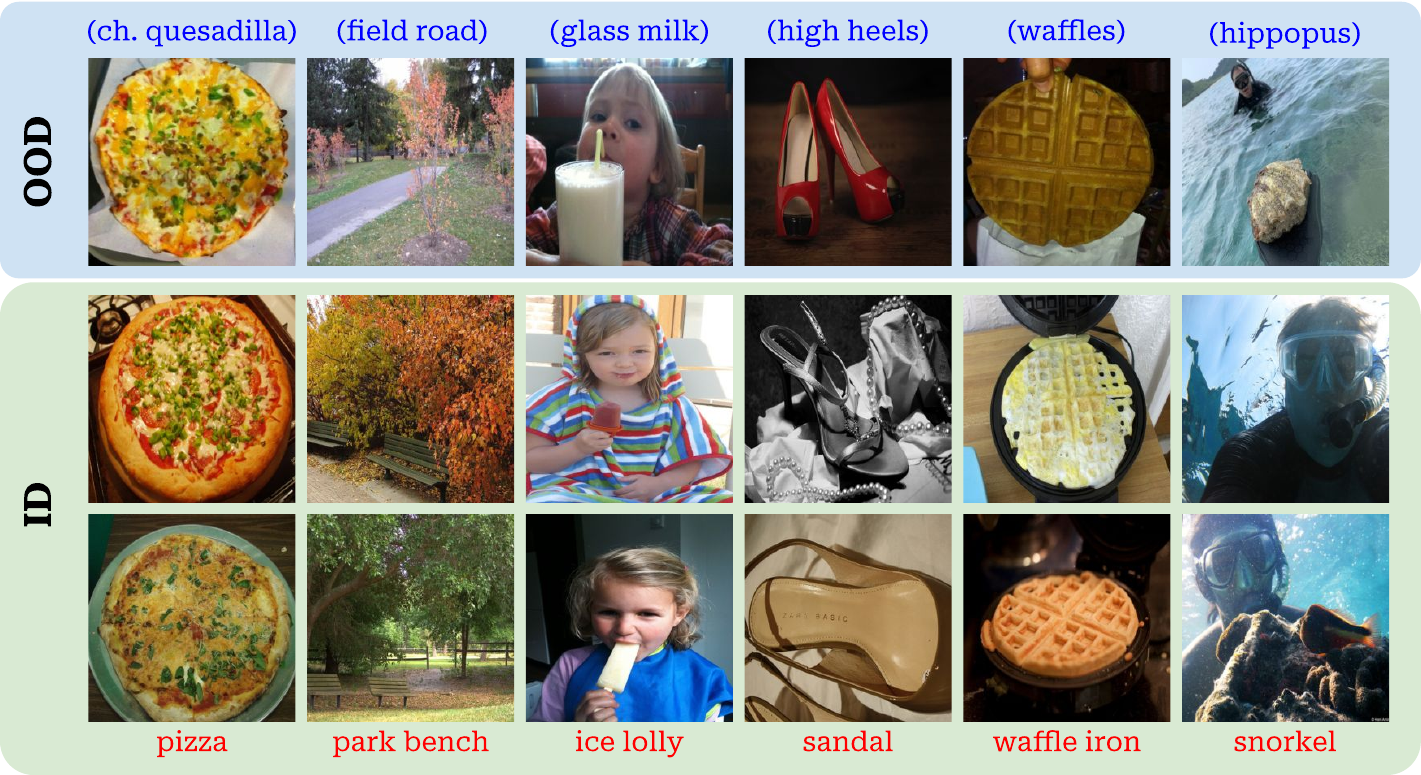}
\caption{
Qualitative analysis for OpenOOD (ImageNet-1k / NINCO).
For each column, the top row shows an OOD sample from the NINCO split, with its original OOD class label annotated in \textcolor{oodclassblue}{\textbf{blue}} (parentheses). 
The two rows below show two ID ImageNet examples from the ID class predicted by the Mahalanobis component of our detector (\ie, the class achieving the minimum Mahalanobis distance in PCA space), with the corresponding predicted ID class label annotated in \textcolor{idclassred}{\textbf{red}}.
}
\label{fig:openood_ninco_samples}
\end{figure}

%% file: Supplementary_Material/figures/openmibood_phakir_samples.tex
\begin{figure}[tb]
\centering
 \includegraphics[width=\linewidth]{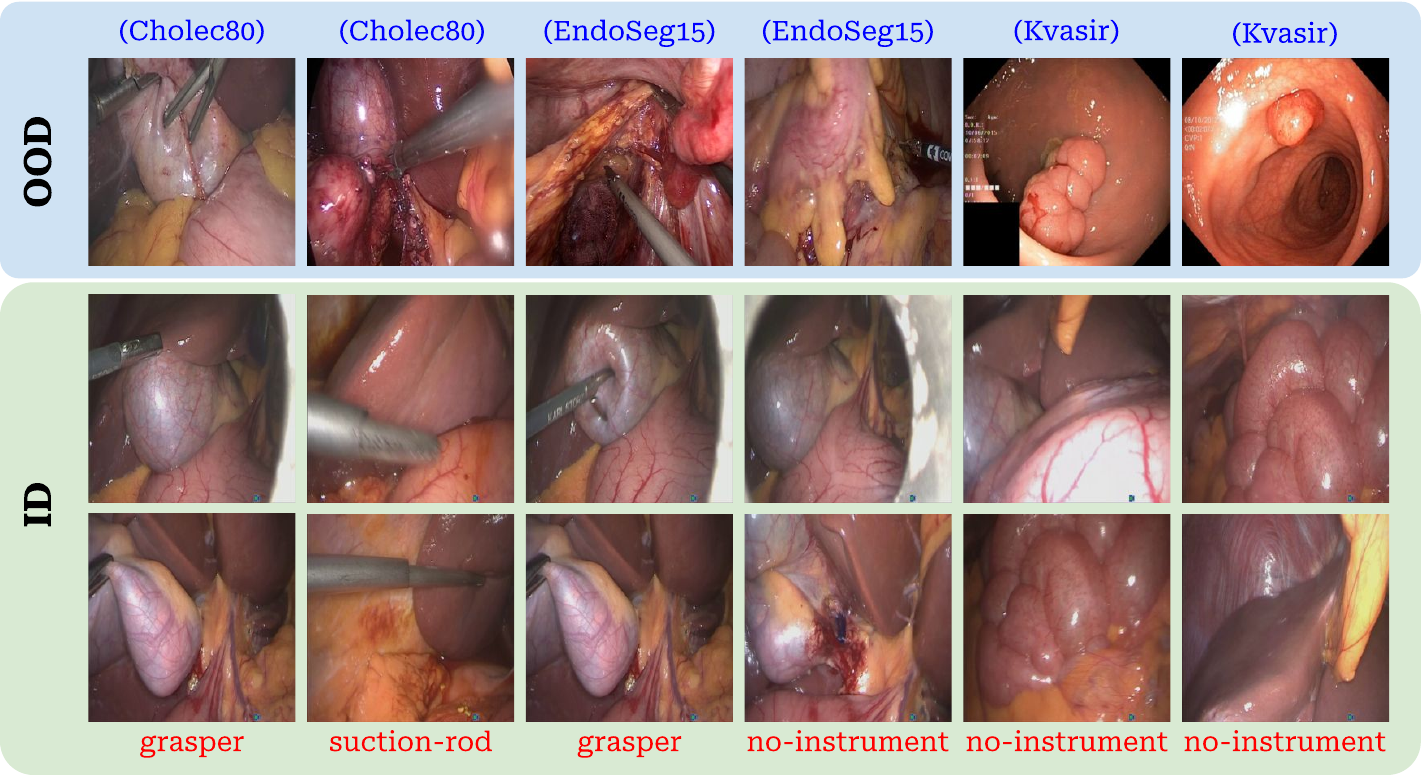}
\caption{
Qualitative analysis on OpenMIBOOD/PhaKIR. 
For each column, the top row shows an OOD sample annotated with the source OOD dataset (\textcolor{oodclassblue}{\textbf{blue}}). 
The two rows below show two ID examples from the Mahalanobis-predicted ID class (\textcolor{idclassred}{\textbf{red}}; argmin Mahalanobis distance in PCA space), illustrating which ID category each OOD sample is mapped to.
}
\label{fig:openmibood_phakir_samples}
\end{figure}

%% file: Supplementary_Material/figures/iconic444_almond_samples.tex
\begin{figure}[tb]
\centering
 \includegraphics[width=\linewidth]{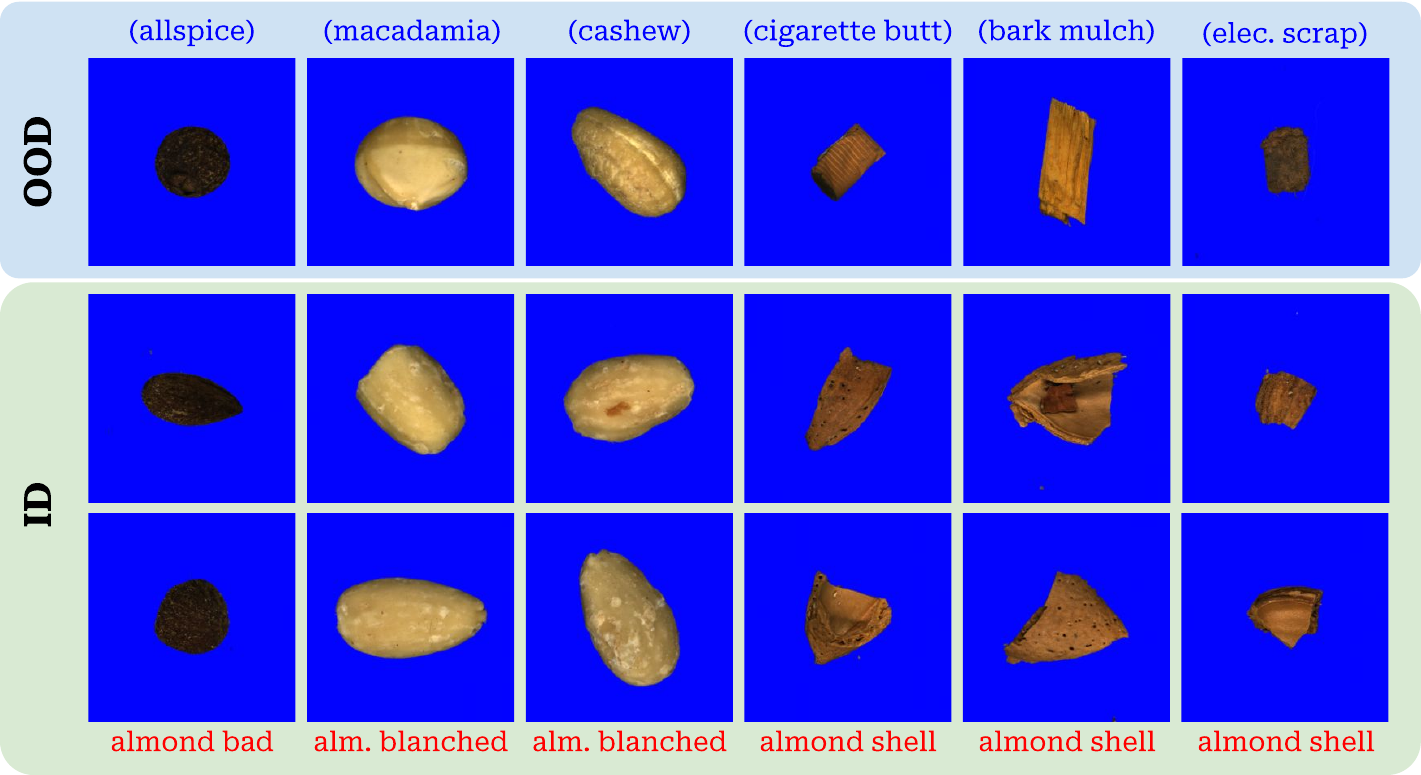}
\caption{
ICONIC-444/almond qualitative examples. 
For each column, the top row shows an OOD sample with its OOD class label (\textcolor{oodclassblue}{\textbf{blue}}), and the two rows below show two ID examples from the Mahalanobis-predicted ID class (\textcolor{idclassred}{\textbf{red}}; argmin Mahalanobis distance in PCA space).
}
\label{fig:iconic444_almond_samples}
\end{figure}

%% file: Supplementary_Material/tables/openmibood_tasks_groupfirst.tex
\begin{landscape}
\begin{table}[tb]
\centering\footnotesize
\setlength{\extrarowheight}{1.5pt}
\setlength{\tabcolsep}{3pt}
\caption{
Average OOD detection performance for the OpenMIBOOD benchmark, measured as FPR at $95\%$ TPR, reported per method and OOD category (near, far, and extreme.
Arrows ($\uparrow$/$\downarrow$) indicate whether higher or lower values are better.
Additionally, the \textbf{best} and \underline{second-best} results in each column are highlighted in bold and underlined, respectively. 
All values are reported as percentages.
*: Results of MDSEns should be interpreted with caution, as discussed in the main manuscript and in prior benchmark analyses~\cite{gutbrod2025_openmibood}.
}
\label{tab:openmibood_full_per_ood_category}
\resizebox{\linewidth}{!}{%
\begin{tabular}{lcccccccccccccccccccccccccc}
\toprule
 \multirow{4}{*}{\textbf{Method}} & \multicolumn{8}{c}{\textbf{MIDOG}} & \multicolumn{8}{c}{\textbf{PhaKIR}} & \multicolumn{8}{c}{\textbf{OASIS-3}} & \multicolumn{2}{c}{} \\
\cmidrule(lr){2-9}\cmidrule(lr){10-17}\cmidrule(lr){18-25}
 & \multicolumn{2}{c}{\textbf{Near}} & \multicolumn{2}{c}{\textbf{Far}} & \multicolumn{2}{c}{\textbf{Extreme}} & \multicolumn{2}{c}{\textbf{Mean}} & \multicolumn{2}{c}{\textbf{Near}} & \multicolumn{2}{c}{\textbf{Far}} & \multicolumn{2}{c}{\textbf{Extreme}} & \multicolumn{2}{c}{\textbf{Mean}} & \multicolumn{2}{c}{\textbf{Near}} & \multicolumn{2}{c}{\textbf{Far}} & \multicolumn{2}{c}{\textbf{Extreme}} & \multicolumn{2}{c}{\textbf{Mean}} & \multicolumn{2}{c}{\textbf{Average}} \\
 & AUROC & FPR95 & AUROC & FPR95 & AUROC & FPR95 & AUROC & FPR95 & AUROC & FPR95 & AUROC & FPR95 & AUROC & FPR95 & AUROC & FPR95 & AUROC & FPR95 & AUROC & FPR95 & AUROC & FPR95 & AUROC & FPR95 & AUROC & FPR95 \\
 & $\uparrow$ & $\downarrow$ & $\uparrow$ & $\downarrow$ & $\uparrow$ & $\downarrow$ & $\uparrow$ & $\downarrow$ & $\uparrow$ & $\downarrow$ & $\uparrow$ & $\downarrow$ & $\uparrow$ & $\downarrow$ & $\uparrow$ & $\downarrow$ & $\uparrow$ & $\downarrow$ & $\uparrow$ & $\downarrow$ & $\uparrow$ & $\downarrow$ & $\uparrow$ & $\downarrow$ & $\uparrow$ & $\downarrow$ \\
\midrule
MSP~\cite{hendrycks2016_ood_msp} & $\phantom{0}55.90$ & $\phantom{0}90.51$ & $\phantom{0}80.91$ & $\phantom{0}71.45$ & $\phantom{0}55.87$ & $\phantom{0}91.13$ & $\phantom{0}64.23$ & $\phantom{0}84.36$ & $\phantom{0}50.14$ & $\phantom{0}96.21$ & $\phantom{0}32.50$ & $\phantom{0}97.65$ & $\phantom{0}37.32$ & $\phantom{0}97.96$ & $\phantom{0}39.99$ & $\phantom{0}97.27$ & $\phantom{0}54.56$ & $\phantom{0}93.62$ & $\phantom{0}79.16$ & $\phantom{0}89.17$ & $\phantom{0}73.97$ & $\phantom{0}83.24$ & $\phantom{0}69.23$ & $\phantom{0}88.68$ & $\phantom{0}57.82$ & $\phantom{0}90.10$ \\
MDS~\cite{lee2018_mahala} & $\phantom{0}63.21$ & $\phantom{0}81.00$ & $\phantom{0}90.91$ & $\phantom{0}41.60$ & $\phantom{0}80.30$ & $\phantom{0}62.95$ & $\phantom{0}78.14$ & $\phantom{0}61.85$ & $\phantom{0}76.49$ & $\phantom{0}70.30$ & $\phantom{0}51.48$ & $\phantom{0}91.14$ & $\phantom{0}93.02$ & $\phantom{0}21.38$ & $\phantom{0}73.66$ & $\phantom{0}60.94$ & $\phantom{0}96.19$ & $\phantom{0}30.57$ & $\textbf{100.00}$ & $\phantom{0}\phantom{0}\textbf{0.00}$ & $\textbf{100.00}$ & $\phantom{0}\phantom{0}\textbf{0.00}$ & $\phantom{0}98.73$ & $\phantom{0}10.19$ & $\phantom{0}83.51$ & $\phantom{0}44.33$ \\
MDSEns*~\cite{lee2018_mahala} & $\phantom{0}\textbf{90.60}$ & $\phantom{0}\textbf{21.83}$ & $\phantom{0}\textbf{100.00}$ & $\phantom{0}\phantom{0}\textbf{0.00}$ & $\phantom{0}\textbf{100.00}$ & $\phantom{0}\phantom{0}\textbf{0.00}$ & $\phantom{0}\textbf{96.87}$ & $\phantom{0}\phantom{0}\textbf{7.28}$ & $\phantom{0}\textbf{97.05}$ & $\phantom{0}\textbf{21.24}$ & $\phantom{0}\textbf{98.41}$ & $\phantom{0}\textbf{15.85}$ & $\phantom{0}\textbf{100.00}$ & $\phantom{0}\phantom{0}\textbf{0.00}$ & $\phantom{0}\textbf{98.49}$ & $\phantom{0}\textbf{12.36}$ & $\phantom{0}\underline{99.67}$ & $\phantom{0}\phantom{0}2.31$ & $\textbf{100.00}$ & $\phantom{0}\phantom{0}\textbf{0.00}$ & $\textbf{100.00}$ & $\phantom{0}\phantom{0}\textbf{0.00}$ & $\phantom{0}\underline{99.89}$ & $\phantom{0}\phantom{0}0.77$ & $\phantom{0}\textbf{98.41}$ & $\phantom{0}\phantom{0}\textbf{6.80}$ \\
GRAM~\cite{sastry20a_gram} & $\phantom{0}\underline{83.20}$ & $\phantom{0}95.05$ & $\phantom{0}92.44$ & $\phantom{0}95.10$ & $\phantom{0}92.83$ & $\phantom{0}93.39$ & $\phantom{0}89.49$ & $\phantom{0}94.51$ & $\phantom{0}14.24$ & $\phantom{0}99.19$ & $\phantom{0}12.18$ & $\phantom{0}98.67$ & $\phantom{0}82.95$ & $\phantom{0}61.61$ & $\phantom{0}36.46$ & $\phantom{0}86.49$ & $\phantom{0}98.27$ & $\phantom{0}\phantom{0}8.90$ & $\textbf{100.00}$ & $\phantom{0}\phantom{0}\textbf{0.00}$ & $\textbf{100.00}$ & $\phantom{0}\phantom{0}\textbf{0.00}$ & $\phantom{0}99.42$ & $\phantom{0}\phantom{0}2.97$ & $\phantom{0}75.12$ & $\phantom{0}61.32$ \\
EBO~\cite{liu2020_ood_ebo} & $\phantom{0}56.85$ & $\phantom{0}88.48$ & $\phantom{0}84.45$ & $\phantom{0}52.10$ & $\phantom{0}58.31$ & $\phantom{0}86.15$ & $\phantom{0}66.54$ & $\phantom{0}75.58$ & $\phantom{0}40.17$ & $\phantom{0}95.41$ & $\phantom{0}34.33$ & $\phantom{0}96.37$ & $\phantom{0}24.31$ & $\phantom{0}98.74$ & $\phantom{0}32.94$ & $\phantom{0}96.84$ & $\phantom{0}50.43$ & $\phantom{0}97.38$ & $\phantom{0}70.69$ & $\phantom{0}91.67$ & $\phantom{0}76.36$ & $\phantom{0}71.99$ & $\phantom{0}65.83$ & $\phantom{0}87.01$ & $\phantom{0}55.10$ & $\phantom{0}86.48$ \\
RMDS~\cite{Ren2021_rmds} & $\phantom{0}52.23$ & $\phantom{0}94.64$ & $\phantom{0}60.68$ & $\phantom{0}89.41$ & $\phantom{0}40.97$ & $\phantom{0}96.78$ & $\phantom{0}51.30$ & $\phantom{0}93.61$ & $\phantom{0}67.71$ & $\phantom{0}90.67$ & $\phantom{0}35.47$ & $\phantom{0}92.69$ & $\phantom{0}57.76$ & $\phantom{0}88.66$ & $\phantom{0}53.65$ & $\phantom{0}90.67$ & $\phantom{0}68.97$ & $\phantom{0}56.84$ & $\phantom{0}98.40$ & $\phantom{0}\phantom{0}2.78$ & $\phantom{0}83.62$ & $\phantom{0}29.53$ & $\phantom{0}83.66$ & $\phantom{0}29.71$ & $\phantom{0}62.87$ & $\phantom{0}71.33$ \\
ReAct~\cite{sun2021_ood_react} & $\phantom{0}57.13$ & $\phantom{0}87.41$ & $\phantom{0}84.53$ & $\phantom{0}54.04$ & $\phantom{0}55.33$ & $\phantom{0}84.63$ & $\phantom{0}65.66$ & $\phantom{0}75.36$ & $\phantom{0}46.12$ & $\phantom{0}95.83$ & $\phantom{0}27.47$ & $\phantom{0}95.83$ & $\phantom{0}28.66$ & $\phantom{0}98.95$ & $\phantom{0}34.08$ & $\phantom{0}96.87$ & $\phantom{0}69.50$ & $\phantom{0}81.51$ & $\phantom{0}76.70$ & $\phantom{0}77.78$ & $\phantom{0}96.63$ & $\phantom{0}\underline{16.81}$ & $\phantom{0}80.94$ & $\phantom{0}58.70$ & $\phantom{0}60.23$ & $\phantom{0}76.98$ \\
MLS~\cite{hendrycks2019_ood_mls} & $\phantom{0}56.58$ & $\phantom{0}88.40$ & $\phantom{0}82.98$ & $\phantom{0}59.64$ & $\phantom{0}57.58$ & $\phantom{0}87.96$ & $\phantom{0}65.71$ & $\phantom{0}78.67$ & $\phantom{0}40.16$ & $\phantom{0}95.37$ & $\phantom{0}34.33$ & $\phantom{0}96.35$ & $\phantom{0}24.32$ & $\phantom{0}98.72$ & $\phantom{0}32.94$ & $\phantom{0}96.81$ & $\phantom{0}50.53$ & $\phantom{0}96.93$ & $\phantom{0}70.78$ & $\phantom{0}91.67$ & $\phantom{0}76.31$ & $\phantom{0}71.74$ & $\phantom{0}65.87$ & $\phantom{0}86.78$ & $\phantom{0}54.84$ & $\phantom{0}87.42$ \\
ViM~\cite{wang2022_ood_vim} & $\phantom{0}64.35$ & $\phantom{0}82.79$ & $\phantom{0}93.11$ & $\phantom{0}39.62$ & $\phantom{0}79.90$ & $\phantom{0}66.45$ & $\phantom{0}79.12$ & $\phantom{0}62.95$ & $\phantom{0}62.70$ & $\phantom{0}91.04$ & $\phantom{0}43.26$ & $\phantom{0}87.68$ & $\phantom{0}87.43$ & $\phantom{0}26.51$ & $\phantom{0}64.46$ & $\phantom{0}68.41$ & $\phantom{0}98.45$ & $\phantom{0}\phantom{0}9.01$ & $\textbf{100.00}$ & $\phantom{0}\phantom{0}\textbf{0.00}$ & $\textbf{100.00}$ & $\phantom{0}\phantom{0}\textbf{0.00}$ & $\phantom{0}99.48$ & $\phantom{0}\phantom{0}3.00$ & $\phantom{0}81.02$ & $\phantom{0}44.79$ \\
KNN~\cite{sun2022knnood} & $\phantom{0}61.63$ & $\phantom{0}85.31$ & $\phantom{0}90.18$ & $\phantom{0}48.92$ & $\phantom{0}76.69$ & $\phantom{0}75.30$ & $\phantom{0}76.17$ & $\phantom{0}69.85$ & $\phantom{0}55.42$ & $\phantom{0}94.39$ & $\phantom{0}37.76$ & $\phantom{0}93.98$ & $\phantom{0}60.19$ & $\phantom{0}76.17$ & $\phantom{0}51.12$ & $\phantom{0}88.18$ & $\phantom{0}97.83$ & $\phantom{0}13.66$ & $\textbf{100.00}$ & $\phantom{0}\phantom{0}\textbf{0.00}$ & $\textbf{100.00}$ & $\phantom{0}\phantom{0}\textbf{0.00}$ & $\phantom{0}99.28$ & $\phantom{0}\phantom{0}4.55$ & $\phantom{0}75.52$ & $\phantom{0}54.19$ \\
DICE~\cite{sun2022dice} & $\phantom{0}54.15$ & $\phantom{0}92.69$ & $\phantom{0}79.08$ & $\phantom{0}66.50$ & $\phantom{0}47.12$ & $\phantom{0}96.12$ & $\phantom{0}60.12$ & $\phantom{0}85.10$ & $\phantom{0}53.32$ & $\phantom{0}93.31$ & $\phantom{0}22.64$ & $\phantom{0}98.59$ & $\phantom{0}18.95$ & $\phantom{0}96.47$ & $\phantom{0}31.64$ & $\phantom{0}96.12$ & $\phantom{0}26.57$ & $\phantom{0}99.95$ & $\phantom{0}23.85$ & $100.00$ & $\phantom{0}41.04$ & $\phantom{0}89.00$ & $\phantom{0}30.49$ & $\phantom{0}96.32$ & $\phantom{0}40.75$ & $\phantom{0}92.51$ \\
FeatNorm~\cite{Yu2023_featurenorm} & $\phantom{0}53.07$ & $\phantom{0}93.02$ & $\phantom{0}39.99$ & $\phantom{0}95.08$ & $\phantom{0}26.20$ & $\phantom{0}98.62$ & $\phantom{0}39.76$ & $\phantom{0}95.57$ & $\phantom{0}41.03$ & $\phantom{0}99.90$ & $\phantom{0}78.48$ & $\phantom{0}91.68$ & $\phantom{0}65.24$ & $\phantom{0}76.73$ & $\phantom{0}61.58$ & $\phantom{0}89.44$ & $\phantom{0}37.97$ & $\phantom{0}86.72$ & $\phantom{0}62.00$ & $\phantom{0}50.00$ & $\phantom{0}87.71$ & $\phantom{0}31.41$ & $\phantom{0}62.56$ & $\phantom{0}56.04$ & $\phantom{0}54.63$ & $\phantom{0}80.35$ \\
ASH-b~\cite{djurisic2023ash} & $\phantom{0}57.79$ & $\phantom{0}86.88$ & $\phantom{0}86.47$ & $\phantom{0}49.53$ & $\phantom{0}62.57$ & $\phantom{0}84.61$ & $\phantom{0}68.94$ & $\phantom{0}73.68$ & $\phantom{0}40.05$ & $\phantom{0}95.17$ & $\phantom{0}60.02$ & $\phantom{0}95.00$ & $\phantom{0}53.23$ & $\phantom{0}96.05$ & $\phantom{0}51.10$ & $\phantom{0}95.41$ & $\phantom{0}70.07$ & $\phantom{0}84.28$ & $\phantom{0}88.06$ & $\phantom{0}52.78$ & $\phantom{0}94.96$ & $\phantom{0}25.91$ & $\phantom{0}84.37$ & $\phantom{0}54.32$ & $\phantom{0}68.14$ & $\phantom{0}74.47$ \\
ASH-s~\cite{djurisic2023ash} & $\phantom{0}56.02$ & $\phantom{0}88.10$ & $\phantom{0}82.70$ & $\phantom{0}58.52$ & $\phantom{0}56.40$ & $\phantom{0}87.36$ & $\phantom{0}65.04$ & $\phantom{0}78.00$ & $\phantom{0}40.15$ & $\phantom{0}94.20$ & $\phantom{0}65.08$ & $\phantom{0}94.28$ & $\phantom{0}56.59$ & $\phantom{0}94.13$ & $\phantom{0}53.94$ & $\phantom{0}94.20$ & $\phantom{0}76.14$ & $\phantom{0}55.99$ & $\phantom{0}91.06$ & $\phantom{0}45.00$ & $\phantom{0}95.57$ & $\phantom{0}\phantom{0}\textbf{0.00}$ & $\phantom{0}87.59$ & $\phantom{0}33.66$ & $\phantom{0}68.86$ & $\phantom{0}68.62$ \\
Residual~\cite{wang2022_ood_vim} & $\phantom{0}65.77$ & $\phantom{0}82.53$ & $\phantom{0}92.35$ & $\phantom{0}43.33$ & $\phantom{0}84.66$ & $\phantom{0}59.09$ & $\phantom{0}80.93$ & $\phantom{0}61.65$ & $\phantom{0}76.99$ & $\phantom{0}\underline{68.69}$ & $\phantom{0}57.33$ & $\phantom{0}88.69$ & $\phantom{0}95.00$ & $\phantom{0}17.56$ & $\phantom{0}76.44$ & $\phantom{0}58.31$ & $\phantom{0}96.53$ & $\phantom{0}32.77$ & $\textbf{100.00}$ & $\phantom{0}\phantom{0}\textbf{0.00}$ & $\textbf{100.00}$ & $\phantom{0}\phantom{0}\textbf{0.00}$ & $\phantom{0}98.84$ & $\phantom{0}10.92$ & $\phantom{0}85.40$ & $\phantom{0}43.63$ \\
SHE~\cite{zhang2023_she} & $\phantom{0}61.80$ & $\phantom{0}85.86$ & $\phantom{0}91.09$ & $\phantom{0}47.57$ & $\phantom{0}75.11$ & $\phantom{0}76.72$ & $\phantom{0}76.00$ & $\phantom{0}70.05$ & $\phantom{0}50.35$ & $\phantom{0}96.99$ & $\phantom{0}47.42$ & $\phantom{0}94.65$ & $\phantom{0}69.35$ & $\phantom{0}78.95$ & $\phantom{0}55.71$ & $\phantom{0}90.20$ & $\phantom{0}90.71$ & $\phantom{0}47.42$ & $\phantom{0}\underline{99.67}$ & $\phantom{0}\phantom{0}\underline{2.50}$ & $\textbf{100.00}$ & $\phantom{0}\phantom{0}\textbf{0.00}$ & $\phantom{0}96.79$ & $\phantom{0}16.64$ & $\phantom{0}76.17$ & $\phantom{0}58.96$ \\
GEN~\cite{Liu2023_GEN} & $\phantom{0}55.46$ & $\phantom{0}91.46$ & $\phantom{0}79.78$ & $\phantom{0}77.84$ & $\phantom{0}54.88$ & $\phantom{0}92.81$ & $\phantom{0}63.38$ & $\phantom{0}87.37$ & $\phantom{0}51.53$ & $\phantom{0}93.13$ & $\phantom{0}32.67$ & $\phantom{0}97.71$ & $\phantom{0}38.83$ & $\phantom{0}94.93$ & $\phantom{0}41.01$ & $\phantom{0}95.26$ & $\phantom{0}54.56$ & $\phantom{0}93.62$ & $\phantom{0}79.16$ & $\phantom{0}89.17$ & $\phantom{0}73.97$ & $\phantom{0}83.24$ & $\phantom{0}69.23$ & $\phantom{0}88.68$ & $\phantom{0}57.87$ & $\phantom{0}90.43$ \\
ATS~\cite{Krumpl2024_ats} & $\phantom{0}67.56$ & $\phantom{0}79.61$ & $\phantom{0}93.88$ & $\phantom{0}35.88$ & $\phantom{0}78.90$ & $\phantom{0}52.69$ & $\phantom{0}80.12$ & $\phantom{0}56.06$ & $\phantom{0}59.06$ & $\phantom{0}89.61$ & $\phantom{0}34.76$ & $\phantom{0}95.52$ & $\phantom{0}77.49$ & $\phantom{0}47.33$ & $\phantom{0}57.10$ & $\phantom{0}77.49$ & $\phantom{0}84.00$ & $\phantom{0}35.05$ & $\phantom{0}93.00$ & $\phantom{0}25.00$ & $\phantom{0}\underline{99.92}$ & $\phantom{0}\phantom{0}\textbf{0.00}$ & $\phantom{0}92.31$ & $\phantom{0}20.02$ & $\phantom{0}76.51$ & $\phantom{0}51.19$ \\
SCALE~\cite{xu2024scaling} & $\phantom{0}55.71$ & $\phantom{0}88.96$ & $\phantom{0}82.03$ & $\phantom{0}60.32$ & $\phantom{0}55.47$ & $\phantom{0}88.29$ & $\phantom{0}64.40$ & $\phantom{0}79.19$ & $\phantom{0}38.97$ & $\phantom{0}94.73$ & $\phantom{0}46.82$ & $\phantom{0}95.16$ & $\phantom{0}37.08$ & $\phantom{0}97.65$ & $\phantom{0}40.95$ & $\phantom{0}95.85$ & $\phantom{0}84.36$ & $\phantom{0}36.90$ & $\phantom{0}98.23$ & $\phantom{0}10.00$ & $\textbf{100.00}$ & $\phantom{0}\phantom{0}\textbf{0.00}$ & $\phantom{0}94.20$ & $\phantom{0}15.63$ & $\phantom{0}66.52$ & $\phantom{0}63.56$ \\
fDBD~\cite{liu2024_fdbd} & $\phantom{0}58.33$ & $\phantom{0}87.80$ & $\phantom{0}83.03$ & $\phantom{0}61.69$ & $\phantom{0}57.89$ & $\phantom{0}84.45$ & $\phantom{0}66.41$ & $\phantom{0}77.98$ & $\phantom{0}50.11$ & $\phantom{0}95.68$ & $\phantom{0}27.54$ & $\phantom{0}95.60$ & $\phantom{0}34.62$ & $\phantom{0}98.61$ & $\phantom{0}37.42$ & $\phantom{0}96.63$ & $\phantom{0}75.34$ & $\phantom{0}87.83$ & $\phantom{0}93.34$ & $\phantom{0}50.00$ & $\phantom{0}94.60$ & $\phantom{0}42.14$ & $\phantom{0}87.76$ & $\phantom{0}59.99$ & $\phantom{0}63.87$ & $\phantom{0}78.20$ \\
MDS++~\cite{mueller2025_mahalanobispp} & $\phantom{0}63.06$ & $\phantom{0}84.21$ & $\phantom{0}94.91$ & $\phantom{0}29.42$ & $\phantom{0}84.18$ & $\phantom{0}63.58$ & $\phantom{0}80.72$ & $\phantom{0}59.07$ & $\phantom{0}56.52$ & $\phantom{0}98.38$ & $\phantom{0}66.57$ & $\phantom{0}89.18$ & $\phantom{0}91.98$ & $\phantom{0}35.58$ & $\phantom{0}71.69$ & $\phantom{0}74.38$ & $\phantom{0}99.63$ & $\phantom{0}\phantom{0}\underline{1.78}$ & $\textbf{100.00}$ & $\phantom{0}\phantom{0}\textbf{0.00}$ & $\textbf{100.00}$ & $\phantom{0}\phantom{0}\textbf{0.00}$ & $\phantom{0}99.88$ & $\phantom{0}\phantom{0}\underline{0.59}$ & $\phantom{0}84.09$ & $\phantom{0}44.68$ \\
NCI~\cite{liu2025_nci} & $\phantom{0}60.81$ & $\phantom{0}84.84$ & $\phantom{0}90.46$ & $\phantom{0}43.62$ & $\phantom{0}68.73$ & $\phantom{0}77.93$ & $\phantom{0}73.34$ & $\phantom{0}68.80$ & $\phantom{0}49.63$ & $\phantom{0}95.07$ & $\phantom{0}27.52$ & $\phantom{0}98.38$ & $\phantom{0}57.08$ & $\phantom{0}97.35$ & $\phantom{0}44.74$ & $\phantom{0}96.93$ & $\phantom{0}70.33$ & $\phantom{0}98.06$ & $\phantom{0}79.46$ & $\phantom{0}97.22$ & $\phantom{0}88.34$ & $\phantom{0}90.75$ & $\phantom{0}79.38$ & $\phantom{0}95.34$ & $\phantom{0}65.82$ & $\phantom{0}87.03$ \\
\midrule
\rowcolor{gray!15}\textbf{PRISM (Ours)} & $\phantom{0}76.23$ & $\phantom{0}\underline{77.99}$ & $\phantom{0}\underline{99.91}$ & $\phantom{0}\phantom{0}\underline{0.60}$ & $\phantom{0}\underline{99.84}$ & $\phantom{0}\phantom{0}\underline{0.78}$ & $\phantom{0}\underline{91.99}$ & $\phantom{0}\underline{26.46}$ & $\phantom{0}\underline{85.37}$ & $\phantom{0}75.29$ & $\phantom{0}\underline{92.45}$ & $\phantom{0}\underline{35.04}$ & $\phantom{0}\underline{99.99}$ & $\phantom{0}\phantom{0}\underline{0.01}$ & $\phantom{0}\underline{92.60}$ & $\phantom{0}\underline{36.78}$ & $\phantom{0}\textbf{99.94}$ & $\phantom{0}\phantom{0}\textbf{0.21}$ & $\textbf{100.00}$ & $\phantom{0}\phantom{0}\textbf{0.00}$ & $\textbf{100.00}$ & $\phantom{0}\phantom{0}\textbf{0.00}$ & $\phantom{0}\textbf{99.98}$ & $\phantom{0}\phantom{0}\textbf{0.07}$ & $\phantom{0}\underline{94.86}$ & $\phantom{0}\underline{21.10}$ \\

\bottomrule
\end{tabular}
}
\end{table}
\end{landscape}

%% file: main.bib
@String(PAMI = {IEEE Trans. Pattern Anal. Mach. Intell.})

@String(IJCV = {Int. J. Comput. Vis.})

@String(CVPR= {IEEE Conf. Comput. Vis. Pattern Recog.})

@String(ICCV= {Int. Conf. Comput. Vis.})

@String(ECCV= {Eur. Conf. Comput. Vis.})

@String(NIPS= {Adv. Neural Inform. Process. Syst.})

@String(ICLR = {Proc. ICLR})

@String(WACV = {})

@String(PAMI  = {IEEE TPAMI})

@String(IJCV  = {IJCV})

@String(CVPR  = {CVPR})

@String(ICCV  = {ICCV})

@String(ECCV  = {ECCV})

@String(NIPS  = {NeurIPS})

@String(ICLR  = {ICLR})

@String(ICLRW  = {ICLRW})

@String(ICLR = {Int. Conf. Learn. Represent.})

@String(PAMI  = {TPAMI})

@String(CVPR  = {Proc. CVPR})

@String(ICCV  = {Proc. ICCV})

@String(ECCV  = {Proc. ECCV})

@String(ICLR  = {Proc. ICLR})

@String(ICLRW  = {Proc. ICLR Workshops})

@String(WACV = {Proc. WACV})

@String(IEEE = {IEEE})

@String(ICML = {Proc. ICML})

@String(NIPSW = {NeurIPS Workshops})

@STRING(NIPSDB= {NeurIPS Datasets and Benchmarks Track})

@String(CBMS = {Proc. CBMS})

@String(TMI   = {IEEE Trans. Med. Imaging})

@String(MedIA = {Med. Image Anal.})

@inproceedings{moosavidezfooli17,
      title={{Universal adversarial perturbations}}, 
      author={Seyed-Mohsen Moosavi-Dezfooli and Alhussein Fawzi and Omar Fawzi and Pascal Frossard},
      booktitle = CVPR,
      year={2017},
}

@inproceedings{Nguyen2014,
    title = {{Deep Neural Networks are Easily Fooled: High Confidence Predictions for Unrecognizable Images}},
    author = {Anh M Nguyen and Jason Yosinski and Jeff Clune},  
    booktitle = CVPR,
    year = {2015},
}

@inproceedings{hein2019relu,
  title={{Why ReLU networks yield high-confidence predictions far away from the training data and how to mitigate the problem}},
  author={Hein, Matthias and Andriushchenko, Maksym and Bitterwolf, Julian},
  booktitle = CVPR,
  year={2019}
}

@article{yang2021_ood_survey,
    title = {{Generalized Out-of-Distribution Detection: A Survey}},
    author = {Yang, Jingkang and Zhou, Kaiyang and Li, Yixuan and Liu, Ziwei},
    journal = {arXiv preprint arXiv:2110.11334},
    year = {2021},  
}

@inproceedings{shiyu17,
    title = {{Enhancing The Reliability of Out-of-distribution Image Detection in Neural Networks}},
    author = {Liang, Shiyu and Li, Yixuan and Srikant, R.},
    booktitle=ICLR,
    year={2018},
}

@inproceedings{hendrycks2016_ood_msp,
    title={{A Baseline for Detecting Misclassified and Out-of-Distribution Examples in Neural Networks}},
    author={Dan Hendrycks and Kevin Gimpel},
    booktitle=ICLR,
    year={2017},
}

@inproceedings{sun2021_ood_react,
    title = {{ReAct: Out-of-distribution Detection With Rectified Activations}},
    author = {Sun, Yiyou and Guo, Chuan and Li, Yixuan},  
    booktitle = NIPS,
    year = {2021},
}

@inproceedings{hendrycks2019_ood_mls,
  title={{Scaling Out-of-Distribution Detection for Real-World Settings}},
  author={Dan Hendrycks and Steven Basart and Mantas Mazeika and Andy Zou and Joseph Kwon and Mohammadreza Mostajabi and Jacob Steinhardt and Dawn Song},
  booktitle=ICML,
  year={2022},
}

@inproceedings{liu2020_ood_ebo,
    title = {{E}nergy-based {O}ut-of-distribution {D}etection},
    author = {Liu, Weitang and Wang, Xiaoyun and Owens, John D. and Li, Yixuan},
    booktitle = NIPS,
    year = {2020},  
}

@inproceedings{sun2022dice,
  title={{DICE: Leveraging Sparsification for Out-of-Distribution Detection}},
  author={Sun, Yiyou and Li, Yixuan},
  booktitle=ECCV,
  year={2022}
}

@inproceedings{djurisic2023ash,
title={{Extremely Simple Activation Shaping for Out-of-Distribution Detection}},
author={Andrija Djurisic and Nebojsa Bozanic and Arjun Ashok and Rosanne Liu},
booktitle=ICLR,
year={2023},
}

@article{krizhevsky09cifar,
  title={{Learning Multiple Layers of Features from Tiny Images}},
  author={Krizhevsky, Alex and Hinton, Geoffrey E.},
  journal={Master's thesis, Department of Computer Science, University of Toronto},
  year={2009},
  publisher={Citeseer}
}

@article{zhou2017places,
  title={{Places: A 10 million Image Database for Scene Recognition}},
  author={Zhou, Bolei and Lapedriza, Agata and Khosla, Aditya and Oliva, Aude and Torralba, Antonio},
  journal=PAMI,
  year={2017},
}

@inproceedings{deng2009_imagenet,
  title={{ImageNet: A large-scale hierarchical image database}}, 
  author={Deng, Jia and Dong, Wei and Socher, Richard and Li, Li-Jia and Kai Li and Li Fei-Fei},
  booktitle=CVPR, 
  year={2009},
}

@inproceedings{bitterwolf2023_ninco,
title={{In or Out? Fixing ImageNet Out-of-Distribution Detection Evaluation}},
author={Julian Bitterwolf and Maximilian Mueller and Matthias Hein},
booktitle=ICLRW,
year={2023},
}

@InProceedings{cimpoi14dtd,
  Author    = {Mircea Cimpoi and Subhransu Maji and Iasonas Kokkinos and Sammy Mohamed and Andrea Vedaldi},
  Title     = {{Describing Textures in the Wild}},
  Booktitle = CVPR,
  Year      = {2014}
}

@article{lecun-mnisthandwrittendigit-2010,
  author={Lecun, Yann and Bottou, L{\'e'}on and Bengio, Yoshua and Haffner, Patrick},
  journal=IEEE, 
  title={{Gradient-based Learning Applied to Document Recognition}}, 
  year={1998},
  volume={86},
  number={11},
  pages={2278-2324},
}

@article{Netzer11_svhn,
title	= {{Reading Digits in Natural Images with Unsupervised Feature Learning}},
author	= {Yuval Netzer and Tao Wang and Adam Coates and Alessandro Bissacco and Bo Wu and Andrew Y. Ng},
year	= {2011},
journal	= NIPSW
}

@InProceedings{He2015_resnet,
	author = {Kaiming He and Xiangyu Zhang and Shaoqing Ren and Jian Sun},
	title = {{Deep Residual Learning for Image Recognition}},
	  Booktitle = CVPR,
	year = {2016}
}

@InProceedings{Horn_2018_inat,
author = {Van Horn, Grant and Mac Aodha, Oisin and Song, Yang and Cui, Yin and Sun, Chen and Shepard, Alex and Adam, Hartwig and Perona, Pietro and Belongie, Serge},
title = {{The INaturalist Species Classification and Detection Dataset}},
booktitle = CVPR,
year = {2018}
}

@article{xiao2017_fmnist,
  author       = {Han Xiao and Kashif Rasul and Roland Vollgraf},
  title        = {{Fashion-MNIST: a Novel Image Dataset for Benchmarking Machine Learning Algorithms}},
  journal = {arXiv preprint arXiv:1708.07747},
  year         = {2017},
}

@inproceedings{haroush2022_statframework,
    title={{A Statistical Framework for Efficient Out of Distribution Detection in Deep Neural Networks}},
    author={Matan Haroush and Tzviel Frostig and Ruth Heller and Daniel Soudry},
    booktitle=ICLR,
    year={2022},
}

@InProceedings{Dong_2022_CVPR_nmd,
    author    = {Dong, Xin and Guo, Junfeng and Li, Ang and Ting, Wei-Te and Liu, Cong and Kung, H.T.},
    title     = {{Neural Mean Discrepancy for Efficient Out-of-Distribution Detection}},
    booktitle = CVPR,
    year      = {2022},
}

@InProceedings{sastry20a_gram,
  title = 	 {{Detecting Out-of-Distribution Examples with {G}ram Matrices}},
  author =       {Sastry, Chandramouli Shama and Oore, Sageev},
  booktitle = ICML,
  year = 	 {2020},
}

@InProceedings{lee2018_mahala,
      title={{A Simple Unified Framework for Detecting Out-of-Distribution Samples and Adversarial Attacks}}, 
      author={Kimin Lee and Kibok Lee and Honglak Lee and Jinwoo Shin},
      booktitle = NIPS,
      year={2018},
}

@inproceedings{sun2022knnood,
  title={{Out-of-distribution Detection with Deep Nearest Neighbors}},
  author={Sun, Yiyou and Ming, Yifei and Zhu, Xiaojin and Li, Yixuan},
  booktitle = ICML,
  year={2022}
}

@inproceedings{yang2022openood,
    title={{O}pen{OOD}: {B}enchmarking {G}eneralized {O}ut-of-{D}istribution {D}etection},
    author={Jingkang Yang and Pengyun Wang and Dejian Zou and Zitang Zhou and Kunyuan Ding and WENXUAN PENG and Haoqi Wang and Guangyao Chen and Bo Li and Yiyou Sun and Xuefeng Du and Kaiyang Zhou and Wayne Zhang and Dan Hendrycks and Yixuan Li and Ziwei Liu},
    booktitle=NIPSDB,
    year={2022},
}

@article{zhang2023openood,
  title={OpenOOD v1.5: Enhanced Benchmark for Out-of-Distribution Detection},
  author={Zhang, Jingyang and Yang, Jingkang and Wang, Pengyun and Wang, Haoqi and Lin, Yueqian and Zhang, Haoran and Sun, Yiyou and Du, Xuefeng and Li, Yixuan and Liu, Ziwei and Chen, Yiran and Li, Hai},
  journal={arXiv preprint arXiv:2306.09301},
  year={2023}
}

@article{zhang2024openood,
title={Open{OOD} v1.5: Enhanced Benchmark for Out-of-Distribution Detection},
author={Jingyang Zhang and Jingkang Yang and Pengyun Wang and Haoqi Wang and Yueqian Lin and Haoran Zhang and Yiyou Sun and Xuefeng Du and Yixuan Li and Ziwei Liu and Yiran Chen and Hai Li},
journal={J. of DMLR},
year={2024},
}

@article{hendrycks2021_ml_safety,
  title = {{Unsolved Problems in ML Safety}},
  author = {Hendrycks, Dan and Carlini, Nicholas and Schulman, John and Steinhardt, Jacob},
  journal = {arXiv preprint arXiv:2109.13916},
  year = {2021}, 
}

@article{hendrycks2022_ml_safety,
    title = {{X-Risk Analysis for AI Research}},
    author = {Hendrycks, Dan and Mazeika, Mantas},
    journal = {arXiv preprint arXiv:2206.05862},
    year = {2022},  
}

@article{mohseni2021_ml_safety,
    title = {{Practical Machine Learning Safety: A Survey and Primer}},
    author = {Mohseni, Sina and Wang, Haotao and Yu, Zhiding and Xiao, Chaowei and Wang, Zhangyang and Yadawa, Jay},
    journal = {arXiv preprint arXiv:2106.04823},
    year = {2021},
}

@inproceedings{wang2022_ood_vim,
    title = {{ViM: Out-Of-Distribution with Virtual-logit Matching}},
    author = {Wang, Haoqi and Li, Zhizhong and Feng, Litong and Zhang, Wayne},
    booktitle = CVPR,
    year = {2022},
}

@article{Torralba2008_tinyimagenet,
  title={{80 Million Tiny Images: A Large Data Set for Nonparametric Object and Scene Recognition}}, 
  author={Torralba, Antonio and Fergus, Rob and Freeman, William T.},
  journal = PAMI,
  year={2008},
}

@InProceedings{Krumpl2024_ats,
  title = 	 {{ATS: Adaptive Temperature Scaling for Enhancing Out-of-Distribution Detection Methods}},
  author =       {Gerhard Krumpl and Henning Avenhaus and Horst Possegger and Horst Bischof},
  booktitle = 	 WACV,
  year = 	 {2024},
}

@inproceedings{xu2024scaling,
    title={{Scaling for Training Time and Post-hoc Out-of-distribution Detection Enhancement}},
    author={Kai Xu and Rongyu Chen and Gianni Franchi and Angela Yao},
    booktitle= ICLR,
    year={2024},
}

@article{Ren2021_rmds,
  title={{A Simple Fix to Mahalanobis Distance for Improving Near-OOD Detection}},
  author={Jie Jessie Ren and Stanislav Fort and Jeremiah Zhe Liu and Abhijit Guha Roy and Shreyas Padhy and Balaji Lakshminarayanan},
  journal = {arXiv preprint arXiv:2106.09022},
  year={2021},
}

@inproceedings{zhang2023_she,
    title = {{Out-of-Distribution Detection based on In-Distribution Data Patterns Memorization with Modern Hopfield Energy}},
    author = {Zhang, Jinsong and Fu, Qiang and Chen, Xu and Du, Lun and Li, Zelin and Wang, Gang and Liu, Xiaoguang and Han, Shi and Zhang, Dongmei},
    booktitle = ICLR,
    year = {2023},
}

@INPROCEEDINGS{Bergmann_mvtecad_1,
author = {Bergmann, Paul and Batzner, Kilian and Fauser, Michael and Sattlegger, David and Steger, Carsten},
title = {{The MVTec Anomaly Detection Dataset: A Comprehensive Real-World Dataset for Unsupervised Anomaly Detection}},
year = {2021},
booktitle = IJCV
}

@article{hassani202_cct,
    title        = {{Escaping the Big Data Paradigm with Compact Transformers}},
    author       = {Ali Hassani and Steven Walton and Nikhil Shah and Abulikemu Abuduweili and Jiachen Li and Humphrey Shi},
    year         = 2022,
    journal = {arXiv preprint arXiv:2104.05704},
}

@InProceedings{vaze2022openset,
   title={{Open-Set Recognition: a Good Closed-Set Classifier is All You Need?}},
   author={Sagar Vaze and Kai Han and Andrea Vedaldi and Andrew Zisserman},
   booktitle=ICLR,
   year={2022}
}

@inproceedings{Liu2023_GEN,
title = {GEN: Pushing the Limits of Softmax-Based Out-of-Distribution Detection},
author = {Liu, Xixi and Lochman, Yaroslava and Christopher, Zach},
booktitle = CVPR,
year = {2023}
}

@InProceedings{Liu_2022_CVPR,
    author    = {Liu, Zhuang and Mao, Hanzi and Wu, Chao-Yuan and Feichtenhofer, Christoph and Darrell, Trevor and Xie, Saining},
    title     = {{A ConvNet for the 2020s}},
    booktitle = CVPR,
    year      = {2022},
}

@inproceedings{dosovitskiy2021vit,
title={{An Image is Worth 16x16 Words: Transformers for Image Recognition at Scale}},
author={Alexey Dosovitskiy and Lucas Beyer and Alexander Kolesnikov and Dirk Weissenborn and Xiaohua Zhai and Thomas Unterthiner and Mostafa Dehghani and Matthias Minderer and Georg Heigold and Sylvain Gelly and Jakob Uszkoreit and Neil Houlsby},
booktitle=ICLR,
year={2021},
}

@misc{rw2019timm,
  author = {Ross Wightman},
  title = {PyTorch Image Models},
  year = {2019},
  publisher = {GitHub},
  journal = {GitHub repository},
  doi = {10.5281/zenodo.4414861},
  howpublished = {\url{https://github.com/huggingface/pytorch-image-models}},
  note = {{Accessed 26 June 2026}}
}

@inproceedings{Paszke19_pytorch,
 author = {Paszke, Adam and Gross, Sam and Massa, Francisco and Lerer, Adam and Bradbury, James and Chanan, Gregory and Killeen, Trevor and Lin, Zeming and Gimelshein, Natalia and Antiga, Luca and Desmaison, Alban and Kopf, Andreas and Yang, Edward and DeVito, Zachary and Raison, Martin and Tejani, Alykhan and Chilamkurthy, Sasank and Steiner, Benoit and Fang, Lu and Bai, Junjie and Chintala, Soumith},
 booktitle = NIPS,
 title = {{PyTorch: An Imperative Style, High-Performance Deep Learning Library}},
 year = {2019}
}

@inproceedings{liu2024_fdbd,
  title={Fast Decision Boundary based Out-of-Distribution Detector},
  author={Liu, Litian and Qin, Yao},
  booktitle={ICML},
  year={2024}
}

@InProceedings{He_2022_MAE,
    author    = {He, Kaiming and Chen, Xinlei and Xie, Saining and Li, Yanghao and Doll\'ar, Piotr and Girshick, Ross},
    title     = {{Masked Autoencoders Are Scalable Vision Learners}},
    booktitle = CVPR,
    year      = {2022},
}

@InProceedings{Yu2023_featurenorm,
    author    = {Yu, Yeonguk and Shin, Sungho and Lee, Seongju and Jun, Changhyun and Lee, Kyoobin},
    title     = {{Block Selection Method for Using Feature Norm in Out-of-Distribution Detection}},
    booktitle = CVPR,
    year      = {2023},
}

@inproceedings{mueller2025_mahalanobispp,
    title={{Mahalanobis++: Improving OOD Detection via Feature Normalization}},
    author={Maximilian Mueller and Matthias Hein},
    booktitle=ICML,
    year={2025},
}

@inproceedings{liu2025_nci,
  title={{Detecting Out-of-Distribution through the Lens of Neural Collapse}},
  author={Liu, Litian and Qin, Yao},
  booktitle=CVPR,
  year={2025}
}

@InProceedings{ramesh21a_zero_shot_text_to_image,
  title = 	 {{Zero-Shot Text-to-Image Generation}},
  author =       {Ramesh, Aditya and Pavlov, Mikhail and Goh, Gabriel and Gray, Scott and Voss, Chelsea and Radford, Alec and Chen, Mark and Sutskever, Ilya},
  booktitle = ICML,
  year = 	 {2021},
}

@article{jumper2021_highaccurate,
author = {Jumper, John and Evans, Richard and Pritzel, Alexander and Green, Tim and Figurnov, Michael and Ronneberger, Olaf and Tunyasuvunakool, Kathryn and Bates, Russ and Žídek, Augustin and Potapenko, Anna and Bridgland, Alex and Meyer, Clemens and Kohl, Simon and Ballard, Andrew and Cowie, Andrew and Romera-Paredes, Bernardino and Nikolov, Stanislav and Jain, Rishub and Adler, Jonas and Hassabis, Demis},
year = {2021},
title = {Highly accurate protein structure prediction with AlphaFold},
journal = {Nature}
}

@article{Suhendar2022_fruit_quality_classification,
author = {Suhendar, Haris and Efelina, Vita and Ziveria, M},
year = {2022},
title = {{Fruit Quality Classification using Convolutional Neural Network}},
journal = {JPCS},
}

@InProceedings{gutbrod2025_openmibood,
  author    = {Gutbrod, Max and Rauber, David and Nunes, Danilo Weber and Palm, Christoph},
  title     = {{OpenMIBOOD: Open Medical Imaging Benchmarks for Out-Of-Distribution Detection}},
  booktitle = CVPR,
  year      = {2025},
}

@inproceedings{krumpl2026_iconic444,
    title     = {{ICONIC-444: A 3.1-Million-Image Dataset for OOD Detection Research}},
    author    = {Krumpl, Gerhard and Avenhaus, Henning and Possegger, Horst},
    booktitle = WACV,
    year      = {2026}
}

@inproceedings{krumpl2026_onemodel,
    title     = {{One Model, Many Behaviors: Training-Induced Effects on Out-of-Distribution Detection}},
    author    = {Krumpl, Gerhard and Avenhaus, Henning and Possegger, Horst},
    booktitle = WACV,
    year      = {2026}
}

@InProceedings{Tang2024_cores,
    author    = {Tang, Keke and Hou, Chao and Peng, Weilong and Chen, Runnan and Zhu, Peican and Wang, Wenping and Tian, Zhihong},
    title     = {{CORES: Convolutional Response-based Score for Out-of-distribution Detection}},
    booktitle = CVPR,
    year      = {2024},
}

@inproceedings{Olber2023_binaryneuronactivationpattern,
    author = {B. Olber and K. Radlak and A. Popowicz and M.Szczepankiewicz and K. Chachula },
    booktitle = CVPR,
    title = {{Detection of out-of-distribution samples using binary neuron activation patterns}},
    year = {2023}
}

@inproceedings{Anthony2023_maha,
	year = 2023,
	author = {Harry Anthony and Konstantinos Kamnitsas},
	title = {{On the Use of Mahalanobis Distance for Out-of-distribution Detection with Neural Networks for Medical Imaging}},
	booktitle = {Proc. UNSURE}
}

@inproceedings{fang2024_kernel,
title={{Kernel {PCA} for Out-of-Distribution Detection}},
author={Kun Fang and Qinghua Tao and Kexin Lv and Mingzhen He and Xiaolin Huang and JIE YANG},
booktitle=NIPS,
year={2024},
}

@InProceedings{Zaeemzadeh2021_unionofsubspaces,
    author    = {Zaeemzadeh, Alireza and Bisagno, Niccolo and Sambugaro, Zeno and Conci, Nicola and Rahnavard, Nazanin and Shah, Mubarak},
    title     = {{Out-of-Distribution Detection Using Union of 1-Dimensional Subspaces}},
    booktitle = CVPR,
    year      = {2021},
}

@article{RUECKERT2026_phakir,
title = {{Comparative validation of surgical phase recognition, instrument keypoint estimation, and instrument instance segmentation in endoscopy: Results of the PhaKIR 2024 challenge}},
journal = MedIA,
year = {2026},
author = {Tobias Rueckert and David Rauber and Raphaela Maerkl and Leonard Klausmann and Suemeyye R. Yildiran and Max Gutbrod and Danilo Weber Nunes and Alvaro Fernandez Moreno and Imanol Luengo and Danail Stoyanov and Nicolas Toussaint and Enki Cho and Hyeon Bae Kim and Oh Sung Choo and Ka Young Kim and Seong Tae Kim and Gonçalo Arantes and Kehan Song and Jianjun Zhu and Junchen Xiong and Tingyi Lin and Shunsuke Kikuchi and Hiroki Matsuzaki and Atsushi Kouno and João Renato Ribeiro Manesco and João Paulo Papa and Tae-Min Choi and Tae Kyeong Jeong and Juyoun Park and Oluwatosin Alabi and Meng Wei and Tom Vercauteren and Runzhi Wu and Mengya Xu and An Wang and Long Bai and Hongliang Ren and Amine Yamlahi and Jakob Hennighausen and Lena Maier-Hein and Satoshi Kondo and Satoshi Kasai and Kousuke Hirasawa and Shu Yang and Yihui Wang and Hao Chen and Santiago Rodríguez and Nicolás Aparicio and Leonardo Manrique and Juan Camilo Lyons and Olivia Hosie and Nicolás Ayobi and Pablo Arbeláez and Yiping Li and Yasmina {Al Khalil} and Sahar Nasirihaghighi and Stefanie Speidel and Daniel Rueckert and Hubertus Feussner and Dirk Wilhelm and Christoph Palm},
}

@article{twinanda2016endonet_cholec80,
 author = {Andru Putra Twinanda and
Sherif Shehata and
Didier Mutter and
Jacques Marescaux and
Michel de Mathelin and
Nicolas Padoy},
journal = {{IEEE} Trans. Medical Imaging},
title = {EndoNet: {A} {Deep} {Architecture} for {Recognition} {Tasks} on {Laparoscopic} {Videos}},
year = {2017}
}

@article{bodenstedt2018_endoseg2015,
 author = {Bodenstedt, Sebastian and Allan, Max and Agustinos, Anthony and Du, Xiaofei and Garcia-Peraza-Herrera, Luis and Kenngott, Hannes and Kurmann, Thomas and M{\"u}ller-Stich, Beat and Ourselin, Sebastien and Pakhomov, Daniil and others},
 journal = {arXiv preprint arXiv:1805.02475},
 title = {{Comparative} {Evaluation} of {Instrument} {Segmentation} and {Tracking} {Methods} in {Minimally} {Invasive} {Surgery}},
 year = {2018}
}

@article{allan2020_endoseg2018,
 author = {Allan, Max and Kondo, Satoshi and Bodenstedt, Sebastian and Leger, Stefan and Kadkhodamohammadi, Rahim and Luengo, Imanol and Fuentes, Felix and Flouty, Evangello and Mohammed, Ahmed and Pedersen, Marius and others},
 journal = {arXiv preprint arXiv:2001.11190},
 title = {2018 {Robotic} {Scene} {Segmentation} {Challenge}},
 year = {2020}
}

@inproceedings{jha2020_kvasir,
 author = {Jha, Debesh and Smedsrud, Pia H and Riegler, Michael A and Halvorsen, P{\aa}l and De Lange, Thomas and Johansen, Dag and Johansen, H{\aa}vard D},
 booktitle = {MMM},
 title = {Kvasir-seg: {A} segmented polyp dataset},
 year = {2020}
}

@article{ALHAJJ201924_cataracts,
title = {{CATARACTS: Challenge on automatic tool annotation for cataRACT surgery}},
journal = MedIA,
year = {2019},
author = {Hassan {Al Hajj} and Mathieu Lamard and Pierre-Henri Conze and Soumali Roychowdhury and Xiaowei Hu and Gabija Maršalkaitė and Odysseas Zisimopoulos and Muneer Ahmad Dedmari and Fenqiang Zhao and Jonas Prellberg and Manish Sahu and Adrian Galdran and Teresa Araújo and Duc My Vo and Chandan Panda and Navdeep Dahiya and Satoshi Kondo and Zhengbing Bian and Arash Vahdat and Jonas Bialopetravičius and Evangello Flouty and Chenhui Qiu and Sabrina Dill and Anirban Mukhopadhyay and Pedro Costa and Guilherme Aresta and Senthil Ramamurthy and Sang-Woong Lee and Aurélio Campilho and Stefan Zachow and Shunren Xia and Sailesh Conjeti and Danail Stoyanov and Jogundas Armaitis and Pheng-Ann Heng and William G. Macready and Béatrice Cochener and Gwenolé Quellec},
}

@article{aubreville2023_midog,
 author = {Aubreville, Marc and Wilm, Frauke and Stathonikos, Nikolas and Breininger, Katharina and Donovan, Taryn A and Jabari, Samir and Veta, Mitko and Ganz, Jonathan and Ammeling, Jonas and van Diest, Paul J and others},
 journal = {Scientific data},
 title = {{A comprehensive multi-domain dataset for mitotic figure detection}},
 year = {2023}
}

@inproceedings{amorim2020_ccagt_1,
 author = {Amorim, Jo{\~a}o Gustavo Atkinson and Macarini, Luiz Antonio Buschetto and Matias, Andr{\'e} Vict{\'o}ria and Cerentini, Allan and Onofre, Fabiana Botelho De Miranda and Onofre, Alexandre Sherlley Casimiro and Von Wangenheim, Aldo},
 booktitle =  CBMS,
 title = {{A novel approach on segmentation of agnor-stained cytology images using deep learning}},
 year = {2020}
}

@misc{atkinson_amorim_ccagt_2,
 author = {Atkinson Amorim, João Gustavo and Matias, André and Bottamedi, Tainee and Sanches, Vinícius and Costa, Ane Francyne and Onofre, Fabiana and Onofre, Alexandre and Wangenheim, Aldo},
 title = {{CCAgT}: {Images} of {Cervical} {Cells} with {AgNOR} {Stain} {Technique}},
 year = {2022}
}

@article{saikia2019_fnac,
 author = {Saikia, Amartya Ranjan and Bora, Kangkana and Mahanta, Lipi B and Das, Anup Kumar},
 journal = {Tissue and Cell},
 title = {Comparative assessment of {CNN} architectures for classification of breast {FNAC} images},
 year = {2019}
}

@article{lamontagne2019_oasis,
 author = {LaMontagne, Pamela J and Benzinger, Tammie LS and Morris, John C and Keefe, Sarah and Hornbeck, Russ and Xiong, Chengjie and Grant, Elizabeth and Hassenstab, Jason and Moulder, Krista and Vlassenko, Andrei G and others},
 journal = {medrxiv},
 title = {{OASIS-3: longitudinal neuroimaging, clinical, and cognitive dataset for normal aging and {Alzheimer} disease}},
 year = {2019}
}

@article{isensee2019_hdbet,
 author = {Isensee, Fabian and Schell, Marianne and Pflueger, Irada and Brugnara, Gianluca and Bonekamp, David and Neuberger, Ulf and Wick, Antje and Schlemmer, Heinz-Peter and Heiland, Sabine and Wick, Wolfgang and others},
 journal = {Human Brain Mapping},
 title = {Automated brain extraction of multisequence {MRI} using artificial neural networks},
 year = {2019}
}

@article{liew2022_atlas,
 author = {Liew, Sook-Lei and Lo, Bethany P and Donnelly, Miranda R and Zavaliangos-Petropulu, Artemis and Jeong, Jessica N and Barisano, Giuseppe and Hutton, Alexandre and Simon, Julia P and Juliano, Julia M and Suri, Anisha and others},
 journal = {Scientific data},
 title = {A large, curated, open-source stroke neuroimaging dataset to improve lesion segmentation algorithms},
 year = {2022}
}

@article{baid2021_brats1,
 author = {Baid, Ujjwal and Ghodasara, Satyam and Mohan, Suyash and Bilello, Michel and Calabrese, Evan and Colak, Errol and Farahani, Keyvan and Kalpathy-Cramer, Jayashree and Kitamura, Felipe C and Pati, Sarthak and others},
 journal = {arXiv preprint arXiv:2107.02314},
 title = {The {RSNA-ASNR-MICCAI BraTS} 2021 {Benchmark} on {Brain} {Tumor} {Segmentation} and {Radiogenomic} {Classification}},
 year = {2021}
}

@article{bakas2017_brats2,
 author = {Bakas, Spyridon and Akbari, Hamed and Sotiras, Aristeidis and Bilello, Michel and Rozycki, Martin and Kirby, Justin S and Freymann, John B and Farahani, Keyvan and Davatzikos, Christos},
 journal = {Scientific data},
 title = {Advancing the {Cancer} {Genome} {Atlas} {Glioma} {MRI} {Collections} with {Expert} {Segmentation} {Labels} and {Radiomic} {Features}},
 year = {2017}
}

@article{menze2014_brats3,
 author = {Bjoern H. Menze and
Andr{\'{a}}s Jakab and
Stefan Bauer and
Jayashree Kalpathy{-}Cramer and
Keyvan Farahani and
Justin S. Kirby and
Yuliya Burren and
Nicole Porz and
Johannes Slotboom and
Roland Wiest and
Levente Lanczi and
Elizabeth R. Gerstner and
Marc{-}Andr{\'{e}} Weber and
Tal Arbel and
Brian B. Avants and
Nicholas Ayache and
Patricia Buendia and
D. Louis Collins and
Nicolas Cordier and
Jason J. Corso and
Antonio Criminisi and
Tilak Das and
Herve Delingette and
{\c{C}}agatay Demiralp and
Christopher R. Durst and
Michel Dojat and
Senan Doyle and
Joana Festa and
Florence Forbes and
Ezequiel Geremia and
Ben Glocker and
Polina Golland and
Xiaotao Guo and
Andac Hamamci and
Khan M. Iftekharuddin and
Raj Jena and
Nigel M. John and
Ender Konukoglu and
Danial Lashkari and
Jos{\'{e}} Antonio Mariz and
Raphael Meier and
S{\'{e}}rgio Pereira and
Doina Precup and
Stephen J. Price and
Tammy Riklin Raviv and
Syed M. S. Reza and
Michael T. Ryan and
Duygu Sarikaya and
Lawrence H. Schwartz and
Hoo{-}Chang Shin and
Jamie Shotton and
Carlos A. Silva and
Nuno J. Sousa and
Nagesh K. Subbanna and
G{\'{a}}bor Sz{\'{e}}kely and
Thomas J. Taylor and
Owen M. Thomas and
Nicholas J. Tustison and
G{\"{o}}zde B. {\"{U}}nal and
Flor Vasseur and
Max Wintermark and
Dong Hye Ye and
Liang Zhao and
Binsheng Zhao and
Darko Zikic and
Marcel Prastawa and
Mauricio Reyes and
Koen Van Leemput},
 journal = TMI,
 title = {The {Multimodal} {Brain} {Tumor} {Image} {Segmentation} {Benchmark} {(BRATS)}},
 year = {2015}
}

@article{antonelli2022_mdsh1,
 author = {Antonelli, Michela and Reinke, Annika and Bakas, Spyridon and Farahani, Keyvan and Kopp-Schneider, Annette and Landman, Bennett A and Litjens, Geert and Menze, Bjoern and Ronneberger, Olaf and Summers, Ronald M and others},
 journal = {Nature communications},
 title = {The medical segmentation decathlon},
 year = {2022}
}

@article{simpson2019_mdsh2,
 author = {Simpson, Amber L and Antonelli, Michela and Bakas, Spyridon and Bilello, Michel and Farahani, Keyvan and Van Ginneken, Bram and Kopp-Schneider, Annette and Landman, Bennett A and Litjens, Geert and Menze, Bjoern and others},
 journal = {arXiv preprint arXiv:1902.09063},
 title = {A large annotated medical image dataset for the development and evaluation of segmentation algorithms},
 year = {2019}
}

@article{tobon2015_mdsh3,
 author = {Catalina Tobon{-}Gomez and
Arjan J. Geers and
Jochen Peters and
J{\"{u}}rgen Weese and
Karen Pinto and
Rashed Karim and
Mohammed Ammar and
Abdelaziz Daoudi and
J{\'{a}}n Margeta and
Zulma L. Sandoval and
Birgit Stender and
Yefeng Zheng and
Maria A. Zuluaga and
Juli{\'{a}}n Betancur and
Nicholas Ayache and
Mohammed Amine Chikh and
Jean{-}Louis Dillenseger and
B. Michael Kelm and
Sa{\"{\i}}d Mahmoudi and
S{\'{e}}bastien Ourselin and
Alexander Schlaefer and
Tobias Schaeffter and
Reza Razavi and
Kawal S. Rhode},
 journal = TMI,
 title = {Benchmark for {Algorithms} {Segmenting} the {Left} {Atrium} {From} {3D} {CT} and
{MRI} {Datasets}},
 year = {2015}
}

@misc{CHAOSdata2019,
 author = {Ali Emre Kavur and M. Alper Selver and Oğuz Dicle and Mustafa Barış and  N. Sinem Gezer},
 publisher = {Zenodo},
 title = {{CHAOS - {Combined} {(CT-MR)} {Healthy} {Abdominal} {Organ} {Segmentation} {Challenge} {Data}}},
 version = {v1.03},
 year = {2019}
}

@article{KAVUR2021_chaos2,
title = {CHAOS Challenge - combined (CT-MR) healthy abdominal organ segmentation},
journal = MedIA,
year = {2021},
author = {A. Emre Kavur and N. Sinem Gezer and Mustafa Barış and Sinem Aslan and Pierre-Henri Conze and Vladimir Groza and Duc Duy Pham and Soumick Chatterjee and Philipp Ernst and Savaş Özkan and Bora Baydar and Dmitry Lachinov and Shuo Han and Josef Pauli and Fabian Isensee and Matthias Perkonigg and Rachana Sathish and Ronnie Rajan and Debdoot Sheet and Gurbandurdy Dovletov and Oliver Speck and Andreas Nürnberger and Klaus H. Maier-Hein and Gözde {Bozdağı Akar} and Gözde Ünal and Oğuz Dicle and M. Alper Selver},
}

@inproceedings{liu2021Swin,
  title={{Swin Transformer: Hierarchical Vision Transformer using Shifted Windows}},
  author={Liu, Ze and Lin, Yutong and Cao, Yue and Hu, Han and Wei, Yixuan and Zhang, Zheng and Lin, Stephen and Guo, Baining},
  booktitle=ICCV,
  year={2021}
}

@INPROCEEDINGS {du2018_r21d,
author = { Tran, Du and Wang, Heng and Torresani, Lorenzo and Ray, Jamie and LeCun, Yann and Paluri, Manohar },
booktitle = CVPR,
title = {{ A Closer Look at Spatiotemporal Convolutions for Action Recognition }},
year = {2018}
}
